\documentclass[journal]{IEEEtran}
\usepackage[utf8]{inputenc} 
\usepackage[T1]{fontenc}    
\usepackage{hyperref}       
\usepackage{url}            
\usepackage{booktabs}       
\usepackage{amsfonts}       
\usepackage{nicefrac}       
\usepackage{microtype}
\usepackage{graphicx}
\usepackage{subfigure}
\usepackage{booktabs} 
\usepackage[section]{placeins}
\usepackage{amsmath}
\usepackage{enumerate}
\usepackage{multirow}
\usepackage{makecell}
\usepackage{threeparttable}
\usepackage{subfigure}
\usepackage{color}
\usepackage{amssymb}
\usepackage{algorithmic}
\usepackage{setspace}
\usepackage{colortbl}
\usepackage{paralist}
\usepackage{bm}
\usepackage[english]{babel}
\usepackage[normalem]{ulem}

\usepackage[ruled]{algorithm2e} 

\newcommand{\D}{{\cal D}}
\newcommand{\C}{{\cal C}}
\newcommand{\bx}{\bm{x}}
\newcommand{\bbeta}{\bm{\beta}}
\newcommand{\balpha}{\bm{\alpha}}
\newcommand{\bV}{\bm{V}}
\newcommand{\bz}{\bm{z}}
\newcommand{\EV}{\mathbb{E}}

\newcommand{\round}{\mathrm{round}}
\newcommand{\diag}{\mathrm{diag}}

\newcommand{\comments}[1]{}

\usepackage{array}
\newcolumntype{L}[1]{>{\vspace{0.2em}\begin{minipage}{#1}\raggedright\let\newline\\
\arraybackslash\hspace{0pt}}m{#1}<{\end{minipage}\vspace{0.2em}}}
\newcolumntype{R}[1]{>{\vspace{0.2em}\begin{minipage}{#1}\raggedleft\let\newline\\
\arraybackslash\hspace{0pt}}m{#1}<{\end{minipage}\vspace{0.2em}}}
\newcolumntype{C}[1]{>{\vspace{0.11em}\begin{minipage}{#1}\centering\let\newline\\
\arraybackslash\hspace{0pt}}m{#1}<{\end{minipage}\vspace{0.11em}}}

\ifCLASSINFOpdf
\else
\fi

\begin{document}
%
\title{Multiple-criteria Based Active Learning with Fixed-size Determinantal Point Processes}



\author{\IEEEauthorblockN{Xueying Zhan\IEEEauthorrefmark{1}, Qing Li\IEEEauthorrefmark{2} and Antoni B.~Chan\IEEEauthorrefmark{3}}\\
\IEEEauthorblockA{\IEEEauthorrefmark{1}Department of Computer Science, 
City University of Hong Kong, 
Email: xyzhan2-c@my.cityu.edu.hk}\\
\IEEEauthorblockA{\IEEEauthorrefmark{2}Department of Computing, 
The Hong Kong Polytechnic University, 
Email: csqli@comp.polyu.edu.hk}\\
\IEEEauthorblockA{\IEEEauthorrefmark{3}Department of Computer Science, 
City University of Hong Kong, 
Email: abchan@cityu.edu.hk}
}
%
%
%


%



\IEEEtitleabstractindextext{%
\begin{abstract}
Active learning aims to achieve greater accuracy with less training data by selecting the most useful data samples from which it learns. Single-criterion based methods (i.e., informativeness and representativeness based methods) are simple and efficient; however, they lack adaptability to different real-world scenarios. In this paper, we introduce a multiple-criteria based active learning algorithm, which incorporates three complementary criteria, i.e., informativeness, representativeness and diversity, to make appropriate selections in the active learning rounds under different data types. We consider the selection process as a Determinantal Point Process, which good balance among these criteria. We refine the query selection strategy by both selecting the hardest unlabeled data sample and biasing towards the classifiers that are more suitable for the current data distribution. In addition, we also consider the dependencies and relationships between these data points in data selection by means of centroid-based clustering approaches. Through evaluations on synthetic and real-world datasets, we show that our method performs significantly better and is more stable than other multiple-criteria based AL algorithms.
\end{abstract}

}

\maketitle

\IEEEdisplaynontitleabstractindextext

%
\IEEEpeerreviewmaketitle

\section{Introduction}

\label{intro}
Active learning (AL) aims to select the most crucial instances and query their labels through the interaction with oracles \cite{settles2009active, fang2017learning, wang2019active}. In real world applications, it is relatively easy to obtain a large amount of unlabeled data, but difficult to get their labels, since the labeling process is quite expensive and time-consuming \cite{shen2004multi}. Therefore, AL is designed to solve the problem through selecting the most informative or representative subsets for labeling. A general process starts from a small or even empty initial labeled set, during which the most crucial instances will be selected from the unlabeled data pool for labeling, and added to training set for updating the classifiers \cite{konyushkova2017learning}. The process  proceeds in rounds until the basic classifier achieves the expected performance or the pre-defined labeling budget is exhausted.

There are two main criteria that are widely adopted in constructing AL querying strategies: informativeness and representativeness. Informativeness refers to the ability to reduce the uncertainty of instances in statistical models. Classical methods include uncertainty sampling \cite{lewis1994heterogeneous}, expected model change \cite{vezhnevets2012weakly}, minimizing misclassification error \cite{long2008active}, variance reduction \cite{cohn1994neural}, etc. These approaches are simple and efficient, but they might also limit the performance of AL, since the informative data samples might not exploit the true structure information of the unlabeled data pool, and the outliers could be included according to these approaches \cite{wang2019active}. The representativeness criterion fills in the above shortcomings: by comparing the similarity/distance among data samples, it measures whether an instance well represents the overall pattern of the unlabeled data pool. For example, Liu et al. propose a representative-based AL algorithm with max-min distance \cite{liu2016representative}. Higher representative scores mean that the data points are less likely to be outliers. However, this criterion requires querying a large number of instances before reaching the optimal decision boundary, and hence it is not as efficient as the informativeness criterion. Huang et al. proposed an AL querying strategy by querying both informative and representative data samples based on the min-max view of AL \cite{huang2010active}.

Besides the above two criteria, diversity, which only appears in batch-mode AL, is also a common criterion that aims to maximize the training utility within a batch. In the batch setting, for instance, if only the uncertainty of the model is considered, then the data points within a batch will all be identical, which is undesirable. Hence, it is advantageous to select samples to maximize batch diversity, but this might include points that provide little new information to the model \cite{ash2019deep}. Combining different selection criterion help AL approaches adapt to various application scenarios. Shen et al. proposed a multi-criteria AL method for an NLP task, which incorporated informativeness, representativeness and diversity with two different selection strategies \cite{shen2004multi}. Wang et al. introduced an AL model based on multi-criteria decision making (MCMD) \cite{wang2014active}. Although multiple-criteria based AL strategies achieve good results, questions remain over the best method to achieve good balance among informativeness, representativeness and diversity.

In this paper, we build a multiple-criteria based AL model, which integrates the informativeness, representativeness and diversity measure by leveraging the Determinantal Point Process (DPP) \cite{kulesza2012determinantal, kulesza2011k}. DPP is a repulsive point process that generates both high quality and diverse subsets from a ground set, where the description of quality varies across different tasks. In this paper, we define the quality as the informativeness score of each instance. For the informativeness criterion, since a single classifier might not be suitable for various data distributions (and using the wrong classifier might even hinder the AL process), we form our informativeness criterion by a committee of classifiers. We weigh these classifiers that best match the current data distribution, while the data points are ranked by their difficulty of being classified correctly, according to disagreements of the committee. A data point that has less confidence to be classified correctly receives higher informativeness score since it is more controversial. As for the representativeness criterion, we consider the spatial arrangement among the data points, such that the selected data points are: 1) far from the previously selected data points, and 2) representative of the unlabeled data points. This is similar in spirit to the clustering problems with various objectives (e.g., k-center, k-median and k-means). We employ the k-center algorithm to establish the network of affiliations between the labeled and unlabeled data pools, and in between the unlabeled data pool. Finally, the diversity criterion is integrated into DPP, whose advantage lies in that it efficiently scores the entire set of unlabeled data samples, rather than scoring each of them individually.

Besides the criteria adopted by the model, another important factor that may influence multi-criteria AL performance is how to control the trade-off between the various criteria. The most widely adopted strategy is to form a weighted combination of the criteria, with fixed hyperparameters that weigh their contributions.
%
%
However, in real-world AL scenarios, the weighting hyperparameters need to be set carefully on each dataset, in order to achieve optimal efficiency.
Keeping the hyperparameters fixed may lead to suboptimal selection, therefore influencing the model performance \cite{ash2019deep}.
%
Such an issue makes DPP \cite{kulesza2012determinantal, kulesza2011k} a natural fit due to its ability to determine the trade-off between multiple criteria with no hyperparameter.
%
 %
Specifically, we treat each AL iteration as a DPP, which models the sample correlations by considering both data quality (higher informativeness score) and diversity (higher representativeness score and larger dissimilarity within one selected batch).


In summary, the contributions of this paper are summarized as follows:
\begin{compactitem}
\item We propose a multiple-criteria based Active Learning algorithm that integrates the informativeness, representativeness and diversity criteria via the Determinantial Point Process. We obtain the optimal and most stable performance under various data distributions by considering multiple criteria, while using DPP to automatically determine the trade-off between sampling high quality and diverse points, and does not contain any hyperparameter.
\item We consider the matching degree between the committee of classifiers and target tasks when calculating the informativeness score, by assigning higher weights to the classifiers that are more suitable to the target data distribution.
\item We propose the Area under the Budget Curve (AUBC) metric to evaluate the performance of various AL methods under varying budgets.
\end{compactitem}

The remainder of this paper is organized as follows. In Section~\ref{relate}, we review related work on pool-based AL and DPP. Section~\ref{model} describes our proposed algorithm in detail. Section~\ref{exper} shows the performance of the proposed AL algorithm on both synthetic and real-world datasets. Finally Section~\ref{summary} concludes the paper and lists the future work.

\section{Related Work}
\label{relate}
In this section, we review related work on pool-based AL and DPP.

\subsection{Pool-based Active Learning}
Most AL algorithms rely on fixed heuristic selection strategies, e.g., uniform sampling (Uniform), uncertainty sampling (US), and query-by-committee (QBC) \cite{settles2009active}. Querying the instances whose predicted labels have the least certainty, US \cite{lewis1994heterogeneous} has become one of the most frequently used AL heuristics since it is both simple and computationally efficient. However, US only considers the uncertainty of samples and ignores their category distribution, thereby restricting the quality of sampling \cite{ye2016practice}. QBC \cite{seung1992query} uses a committee of models $\C = \{\theta^{(1)},..., \theta^{(C)}\}$ (constructed by ensemble methods or various models), which are trained on the currently labeled data \cite{settles2009active} to predict the labels of the unlabeled data samples, among which the ones with the largest disagreement will be selected for labeling by an oracle. However, with a majority voting strategy, QBC treats each classifier and data point with the same importance, which fails to take into account the difference in the classification ability of its classifiers on various datasets. Based on these considerations, we propose a selection strategy according to the difficulty of each unlabeled data sample and the capability of each classifier (to be detailed in Section~\ref{informativeness}).

Some AL strategies serve for specific classifiers. Kapoor et al. proposed an algorithm which balances exploration and exploitation by incorporating mean and variance estimation of the Gaussian Process classifier\cite{kapoor2007active} . Kremer et al. proposed a SVM-based AL strategy by minimizing the distances between data points and classification hyperplane \cite{kremer2014active}. These model-driven active learning strategies aim to estimate how strongly learning from a data point influences the current model.

In addition to these strategies that aim to find the instances with uncertainty or make the largest impact of current models, other works also select instances with the following properties: 1) the selected instance should be dissimilar with other instances in the unlabeled data pool;  and 2) the selected instance should be dissimilar with other instances in one batch. Properties 1 and 2 are considered as the representativeness and diversity criteria in the active learning task. Ebert et al. introduced a reinforced active learning formulation, which enables a time varying trade-off between exploration and exploitation \cite{ebert2012ralf}.

Shen et al. incorporated informativeness, representativeness and diversity with two different combination strategies \cite{shen2004multi}:
Strategy 1 first pre-selects the most informative data samples by US, then clusters them into $k$ classes, whose centroids are the selected data samples in one batch. However, informative instances usually do not exploit the structure information of unlabeled data. Moreover, when the number of samples for clustering is small, the cluster centroid may also be an outlier.
Strategy 2 first selects a subset of the data via ranking with a weighted sum of the informativeness and representativeness scores, and then selects samples with the largest pairwise dissimilarity in the subset. Both strategies are sub-optimal, since they fail to consider global correlations among the entire unlabeled data pool.

The biggest advantage of AL is that it reduces the cost of labeling data, whereas Deep Learning (DL) could learn well in the context of high-dimensional data processing and automatic feature extraction. Combining the two approaches (referred as Deep Active Learning, DAL) would enhance each other. Gal et al. developed an AL framework by taking advances in bayesian DL for high-dimensional data \cite{Gal2017DeepBA}. An extension of \cite{Gal2017DeepBA} is \cite{kirsch2019batchbald} (batch-mode extension), which jointly scoring points by estimating the mutual information between a joint of multiple data points and the model parameters. Ash et al. proposed a deep batch AL model which creates the diverse batches of instances that the current model is uncertain by diverse gradient embedding \cite{ash2019deep}. Ducoffe et al. proposed an uncertainty-based deep active learning model based on DeepFool Adversarial Attack, which queries examples lying close to the decision boundary \cite{ducoffe2018adversarial} .

\subsection{Determinantal Point Process}

The Determinantal Point Process (DPP) is an effective probabilistic model of mimicking particles with repulsive interactions in theoretical quantum physics \cite{macchi1975coincidence}. It prohibits co-occurrences of highly correlated quantum states. In machine learning, DPPs \cite{kulesza2012determinantal} have been used for various tasks, such as recommender systems \cite{chen2018fast}, video summarization \cite{li18how} and pedestrian detection \cite{lee2016individualness}. Generally speaking, this model is suitable for scenarios where a diversified subset needs to be selected. Zhang et al. proposed a mini-batch diversification scheme based on DPPs for stochastic gradient descent algorithms \cite{zhang2017determinantal}. Elfeki et al. propose an unsupervised penalty loss that alleviates mode collapse while producing higher quality samples of generative models inspired by DPPs \cite{elfeki2019gdpp}. In recent years, 
DPP variants have been derived for different task modes. Kulesza et al. proposed conditional fixed-size DPP ($k$DPPs) which models only sets of cardinality $k$, and 
improves the suitability of the DPP for various tasks \cite{kulesza2011k}. Mariet et al. introduced a DPP model that uses Kronecker (tensor) product kernels to enable efficient sampling and learning for DPPs \cite{mariet2016kronecker}. Affandi et al. construct a Markov DPP for modeling diverse sequences of subsets \cite{affandi2012markov}. Biyek et al. proposed a batch AL with DPP, which integrates the uncertainty measure with DPP (diversity measure); they focus on approximating the $k$DPPs and adding a hyperparameter for weighting the informativeness (uncertainty) vector \cite{biyik2019batch}. In contrast, we focus on finding the informativeness and representativeness criterion that are generally suitable, and adopt $k$DPP as a good framework for incorporating them together.

\section{Methodology}
\label{model}
In this section we describe our multiple-criteria based AL algorithm. In an unlabeled data pool consisting of $N$ instances ($e.g.$, pictures or paragraphs), each of which belongs to one of several possible categories of interest, we will sequentially select instances from the unlabeled data pool based on multiple-criteria and query their labels from oracle or human annotators. These criteria include informativeness, representativeness and diversity to maximize the overall quality and dissimilarity within a batch. These criteria are combined together through DPP via modeling the diversity and quality of unlabeled data simultaneously.

\subsection{Problem Definition}
\label{definition}
We consider a general AL process for classification problems, in which an oracle provides a fixed number of labels or the active learners have fixed budget. Let $B$ be the number of labels that the oracle can provide, i.e., the budget. Our goal is to select the most critical $B$ instances so as to obtain the highest classification accuracy. We have a small initial labeled dataset $\D_{l}=\{(\bx_1,y_1),...,(\bx_M,y_M)\}$ and a large unlabeled data pool $\D_{u}=\{\bx_1,...,\bx_N\}$, where each instance $\bx_i\in \mathbb{R}^d$ is a $d$-dimensional feature vector and $y \in \{0,1\}$ is the class label of $\bx_i$ for binary classification task\footnote{We extend this to the multi-class case at the end of Section~\ref{informativeness}.}. In each iteration, the active learner selects a batch with size $S$ from $\D_u$, and queries their labels from the oracle/human annotator, based on which we update $\D_l$ and $\D_u$. The process terminates when the budget $B$ is exhausted. In the selection process, we have the following three requirements: (1) the selected instance should be hard to be correctly classified by classifiers; (2) the selected instances should vary from other data points, including both labeled and unlabeled data; (3) the selected instances within one batch should be as diverse as possible.

\subsection{Multi-criteria based Active Learning}
In this sub-section we propose metrics for the three AL criterion, i.e., informativeness, representativeness, and diversity, along with the DPP for combining these criterion for sample selection.

\subsubsection{Informativeness}
\label{informativeness}
A typical strategy for measuring informativeness is QBC. Comparing with certainty-based measure (e.g., US), such a committee-based measure could work for many data sets. This is because at the beginning, we do not know the distribution of the data, and different classifiers have different performance for various distributions of data. However, QBC treats each classifier with the same importance, which may be inappropriate in some cases. For example, for non-linearly separable data, the non-linear classifiers should be assigned higher weights compared to linear classifiers, and vice versa. Similarly, those instances that are more difficult to be classified (e.g., data points that are close to the decision boundary) should also be given more attention. To address these two problems, we propose to use a GLAD framework (Generative model of Labels, Abilities, and Difficulties) \cite{whitehill2009whose}. GLAD was originally designed for crowdsourcing task to aggregate labels by modeling the ground truth labels with observed labels, data difficulties and labeler accuracies. This is very close to our requirements but of slightly difference, since we do not need to estimate the ground truth labels (ground truth would be provided by AL process later). We are more interested in utilizing the byproduct of GLAD (i.e., data difficulties and labeler accuracies estimated by GLAD) to weigh the committee of classifiers.

In each iteration, we first train the committee of classifiers $\C = \{\theta^{(1)},..., \theta^{(C)}\}$ on the current labeled dataset, and the classifier $j$ will make prediction $v_{ij}$ for the unlabeled instance $i$. The voting matrix $\bV={[v_{ij}]}_{i \in \D_u, j\in \C}$ is modeled via probabilistic inference, based on the difficulty of instances (denoted as $\alpha_i\geq0$ for instance $i$) and the fitness between the current data and the classifiers (denoted as $\beta_j\geq0$ for classifier $j$). Specifically, the probability of the predicted label $v_{ij}$ being equal to the ground truth label (denoted as $z_i$), i.e., the probability of correct prediction, is
\begin{equation}
p(v_{ij}=z_{i}|\alpha_{i},\beta_{j})=\frac{1}{1+e^{-\alpha_{i}\beta_{j}}}.
\end{equation}
Note that the range of the  probability is $0.5$ to $1.0$, where $0.5$ corresponds to a random situation (random classifier with random data distribution), and 1.0 corresponds to an easy instance and strong classifier.
In particular, consider the following three extreme cases:

\begin{compactitem}
\item The instance $i$ is extremely easy to be classified ($\alpha_i \to +\infty$), while classifier $j$ is ideally suited to the current data distribution ($\beta_j \to +\infty$). Thus, the probability of correct prediction is $1$.

\item The instance $i$ is extremely difficult to classify ($\alpha_i = 0$), and the probability of a correct prediction is $0.5$.
\item The classifier $j$ is entirely inappropriate to the current data distribution ($\beta_j = 0$), and the probability of correct prediction also converges to $0.5$.
\end{compactitem}

Since the ground truth labels  $\bz=\{z_i\}$ are unobservable, we employ the Expectation Maximization (EM) algorithm to estimate the model parameters $\balpha=\{\alpha_i\}$ and $\bbeta=\{\beta_j\}$.

Given the set of observed predictions $\bV=\{v_{ij}\}$, the maximum a posteriori (MAP) estimation of $(\balpha, \bbeta)$ is
\begin{equation}
\begin{aligned}
\balpha^*, \bbeta^* & = \arg\max\limits_{\balpha, \bbeta} p(\balpha, \bbeta|\bV) \\
                & = \arg\max \sum_{\bz\in{\{0,1\}}^{n}}p(\bz)p(\balpha, \bbeta|\bV,\bz).
\end{aligned}
\end{equation}
In the E-step, we have:
\begin{equation}
\begin{aligned}
p(z_i|\mathbf{v_i},\alpha_i,\bbeta) \propto p(z_i) \prod_j p(v_{ij}|z_i,\alpha_i,\beta_j),
\end{aligned}
\end{equation}
where $p(z_i|\alpha_i,\bbeta) = p(z_i)$, depending on the conditional independence assumption \cite{whitehill2009whose}.

In the M-step, we build the Q-function,
\begin{equation}
\begin{aligned}
Q(\balpha,\bbeta) & = \EV[\ln p(\bV,\bz|\balpha,\bbeta)] \\
& = \sum_i \EV[\ln p(z_i)] + \sum_{ij}\EV[\ln p(v_{ij}|z_i,\alpha_i,\beta_j)],
\end{aligned}
\end{equation}
and maximize it w.r.t. $(\balpha,\bbeta)$ using gradient ascent.

After EM, we obtain the estimated instance difficulty score ${\alpha}_i$, the classifier capability score ${\beta}_j$ and the confidence score $c_{ij}=p(v_{ij}=z_i|{\alpha}_i, {\beta}_j)$. The informativeness score $\mu_i$ is defined as the class entropy,
\begin{equation}
\label{info}
\begin{aligned}
\mu_i = -\sum_{k} P^{(i)}(y_k) \log_k P^{(i)}(y_k),
\end{aligned}
\end{equation}
where for binary classification, $k\in\{0,1\}$ and
\begin{equation}
\label{bin}
\begin{aligned}
P^{(i)}(y_k) = \tfrac{1}{C}\sum_{j \in \C}[{c_{ij}k+(1-c_{ij})(1-k)}],
\end{aligned}
\end{equation}
is the average class probability weighted by the classifier confidence.

Extension to Multi-Class: We next propose a novel extension of GLAD for binary classification to multi-class classification where $y \in \{1,...,K\}$, with $K>2$ classes. The average multi-class class probability is estimated by the multinomial distribution of $K$ categories:

\begin{equation}
\label{mult}
\begin{aligned}
P^{(i)}(y_k) = \tfrac{1}{C}\sum_j \Big[c_{ij}^{\delta(v_{ij},k)}\left(\tfrac{1-c_{ij}}{K-1}\right)^{1-\delta(v_{ij},k)}\Big],
\end{aligned}
\end{equation}
where the Kronecker delta function defined below,
\begin{equation}
\begin{aligned}
\delta(v_{ij},k) = \begin{cases}
                   1, & v_{ij} = k,
                  \\0, & v_{ij} \neq k,
                          \end{cases}
\end{aligned}
\end{equation}
is employed to distinguish the votings of different classifiers.
The probability of incorrect prediction is amortized over the remaining labels.

\subsubsection{Representativeness}

The representativeness criterion aims to choose a subset of the dataset that well summarizes the whole dataset, so that the model trained on the subset could perform as closely as possible to the model trained on the whole dataset.
A common approach is to optimize a centroid-based clustering objective. In our work, we have first obtained the representative information with the help of the k-center algorithm \cite{kleindessner2019fair}. We have a finite set $\D_f = \D_u \cup \D_l$ and define $d : \D_f \times \D_f \to \mathbb{R}$ as the distance metric for $\D_f$. The goal of k-center algorithm is to find a subset $\mathcal{O} \subseteq \D_f$ ($|\mathcal{O}| = \kappa$) such that the maximum distance of the point in $\D_f$ to the closest point in $\mathcal{O}$ is minimized,

\begin{equation}
\begin{aligned}
\min_{\substack{\mathcal{O} \subseteq \D_f \\|\mathcal{O}|=\kappa}} \max_{d_f \in \D_f} \min_{o \in \D_l \cup \mathcal{O}} d(d_f,o),
\end{aligned}
\end{equation}
where $\kappa$ is the number of centers (not included the given centers $D_l$) and $\mathcal{O} = \{o_1,...,o_{\kappa}\}$ is the set of new centers. The set of centers creates spheres with a minimum radius that cover the entire data $D_f$. Note that: 1) the labels of $\D_l$ do not participate in the process of formulating the center set since we only consider the pairwise distance between each data point; 2) $\D_l$ works as the given center set, which might help decrease the radius; and 3) the final representativeness considers both the distance among the unlabeled data pool and the relationship between unlabeled and labeled data \cite{sener2018active}.

After obtaining set $\mathcal{O}$, we construct representative vectors (denoted as $\mathbf{r}_i$) for the data points in $\D_u$. We consider this as a clustering problem, where the relationship between the data points and centers is represented by intra-cluster similarity 
and inter-cluster similarity, 
with a good clustering result to have high intra-cluster similarity and low inter-cluster similarity. The similarities 
are calculated by the cosine similarity between $\mathbf{\D_u}$ and $\mathcal{O}$.
%
Specifically, if $\bx_i\in \D_u$ is assigned to center $o_j$, then
\begin{equation}
\label{rep}
\begin{aligned}
 {r_i}_j = \cos(\bx_i,\mathbf{o_j}).
\end{aligned}
\end{equation}
 We consider $\mathbf{r}_i = [r_{ij}]_{j=1}^{\kappa}$ as the new ``feature representation'' for the representativeness information. For the data pair $(\bx_i,\bx_j)$ in the same cluster, the similarity between the corresponding $\mathbf{r}_i$ and $\mathbf{r}_j$ should be high, since they both have high intra-cluster similarity to the same cluster. If the pair are assigned to different clusters, the similarity will be low. Note that the number of centers $\kappa$ is a non-fixed parameter during AL, and $\kappa$ should decrease as the unlabeled data pool decreases in size. In this work, we set $\kappa = \round(\sqrt{|D_u|/2})$. If $\kappa = |\D_u|$, then $\mathbf{r}_i$ is the pairwise similarity vector between $\bx_i$ and all other points.

\subsubsection{Diversity}

The diversity criterion only appears in batch-mode AL and it aims to maximize the training utility of a batch. To maximize its effectiveness, we prefer a batch in which the examples have high variance among each other. Since the diversity criterion is incorporated into the DPP, we will introduce it along with our introduction of $k$DPPs below.

\subsubsection{$k$DPPs}
\label{kdpps}
We consider the AL process with a Determinantal Point Process (DPP) formulation. In this paper, we use the quality and diversity factors particularly to define the DPP for AL with a unary informativeness score (quality) and pairwise similarity (diversity) of unlabeled data points.

Formally, let $N$ be the total number of instances in the unlabeled data pool. Each instance $i$ has its own quality score $q_i$ and similarity score $s_{ij}$ with another instance $j$. Let $\D_t \subset \D_u$ be the index set of a selected instance batch. DPP aims to simultaneously select high-quality instances while avoiding similar instances. We set the quality score as the informativeness score $\mu_i$, and calculate the similarity score using the representativeness vector $\mathbf{r}_i$. The similarity matrix can be constructed using various kernels, e.g., Laplacian kernel, Gaussian kernel, sigmoid kernel, polynomial kernel, heat kernel \cite{guo2017active}:

\begin{align}
\label{Laplacian}
& s^{laplacian}_{ij}=\exp(-\gamma||\mathbf{r}_i-\mathbf{r}_j||_1),
\\
& s^{gaussian}_{ij}=\exp(-\frac{||\mathbf{r}_i-\mathbf{r}_j||^2}{2\sigma^2}),
\\
& s^{sigmoid}_{ij}=\tanh(\gamma \mathbf{r}_i^T\mathbf{r}_j+c_0),
\\
& s^{poly}_{ij}=(\gamma\mathbf{r}_i^T\mathbf{r}_j+c_0)^{d_0},
\\
& s^{heat}_{ij}=\exp(-\frac{||\mathbf{r}_i-\mathbf{r}_j||^2}{\sigma^2}),
\end{align}
%
%
%
%
%
%
where $\sigma$ is set to the mean pairwise Euclidean distance between representativeness vectors in the unlabeled data pool, $c_0$ is a coefficient, $d_0$ is the degree of polynomial kernel. Note that $\gamma$ is set as $\frac{1}{\kappa}$. Both Gaussian and heat kernels belong to RBF family of kernels, but use different $\gamma$ settings. The performance of  different kernel functions will be discussed in Section~\ref{kernels}.

\textbf{DPP for AL.}
Using the informativeness, representativeness and diversity factors, we construct the positive semi-definite DPP kernel matrix. The DPP can be represented as an $N \times N$ L-ensemble kernel matrix $\mathbf{L}$, denoted as ``L-representation''.
In the ``L-representation'', sampling a subset $\D^t \in \D_u$, we specify the atomic probabilities for every possible instantiation of $\D_t$:
\begin{equation}
\begin{aligned}
P(\D^t \subseteq \D_u) = \det(\mathbf{L}_{\D^t}),
\end{aligned}
\end{equation}
where $\mathbf{L}_{\D^t}$ is a $|\D^t| \times |\D^t|$ kernel matrix (a submatrix of $\mathbf{L}$),  representing the pairwise correlations within $\D^t$.
Then we have
\begin{equation}
\begin{aligned}
P(\D^t) = \frac{\det(\mathbf{L}_{\D^t})}{\det(I + \mathbf{L})},
\end{aligned}
\end{equation}
where 
the normalization constant $\sum_{\D_t \subseteq \D_u}\det(\mathbf{L}_{\D_t}) = \det(I + \mathbf{L})$ is given in closed form.

We denote our DPP with ``L-representation'' as $\mathbf{L}_{\D^t}={[L_{ij}]}_{{i,j} \in \D^t}$, where $L_{ij}=\mu_i \mu_j s_{ij}$.
With the DPP, the probability of selecting instances is computed as the determinant of $\mathbf{L}_{\D^t}$.
%
Kulesza et al. \cite{kulesza2012determinantal} provides an intuitive geometric interpretation of how determinants work for selecting both high quality and diverse subsets. For any positive semi-definite $\mathbf{L}$ matrix, we can always find a  matrix $\mathbf{B}$ such that $\mathbf{L} = \mathbf{B}^T\mathbf{B}$. For example, in our algorithm, the $\mathbf{L}$ matrix can be decomposed as $\mathbf{L}=\diag(\bm{\mu}) \mathbf{S} \mathbf{S}^T \diag(\bm{\mu})$. The determinants are represented as $\det(\mathbf{L}_{\D_t}) = Vol^2(\{B_i\}_{i \in {\D_t}})$, where $B_i$ is the i-th column of $\mathbf{B}$. Intuitively, the columns of $\mathbf{B}$ can be considered as the feature representations describing the elements of $\D_u$, and $\mathbf{L}$ measures the inner product similarity between these feature representations. Thus, the probability of selecting $\D_t$ as the subset via DPP is calculated by the volume spanned by its associated feature representations. Hence, the subsets that have higher inner dissimilarity are more probable because their feature vectors are more orthogonal and thus span larger volumes (i.e., have larger determinants). Instances with similar feature representations are selected together with low probability. Furthermore, the instances with larger magnitude feature representations are more likely to be selected, because they multiply the spanned volumes for sets containing them.

\textbf{kDPP for AL.}
 A standard DPP models both the size and content of the selected subset by the number of initially selected eigenvectors and the span of those vectors, and assigns a probability to every subset of $\D_u$. Therefore, realisation of DPPs has a random number of elements in the sampled subset. However, in our batch-mode AL task, the batch size $S$ is fixed in each iteration. Therefore, we consider a fixed size Determinantal Point Process ($k$DPP) \cite{kulesza2011k}, to address this issue by conditioning a DPP on the cardinality of random $k$-sets ($k \in \{1,...,N\}$) (in our work, $k$ refers to the batch size $S$). The optimization goal is
\begin{equation}
\label{dppgoalnew}
\begin{aligned}
\D_t^*=\mathop{\arg\max}\limits_{\D_t \subset \D_u} \mathcal{P}^S_\mathbf{L}(\D_t),
\end{aligned}
\end{equation}
where
\begin{equation}
\label{dppgoalnew2}
\begin{aligned}
\mathcal{P}^S_{\mathbf{L}}(\D_t) = \left\{ \begin{array}{lcl} \frac{\det(\mathbf{L}_{\D_t})}{\sum_{|\D_t^{'}|}\det(\mathbf{L}_{\D_t^{'}})},  & \mbox{if} & |\D_t| = S, \\ 0, & \mbox{if} & |\D_t| \neq S. \end{array} \right.
\end{aligned}
\end{equation}

A $k$DPP is also a mixture of elementary DPPs (possible subsets), but it only assigns non-zero weights to those with size $k$.
Since this is a sampling process, $k$DPPs and standard DPP only differ in sampling the index sets -- $k$DPPs sample the index set according to their corresponding mixture components \cite{kulesza2011k}.

The $k$DPP sampling process is described in detail in \cite{kulesza2011k}. We give a brief introduction of sampling from $k$DPPs here. The $k$DPPs sampling algorithms proceed in two phases. Firstly, a subset $V$ of the eigenvectors decomposed from $\mathbf{L}$ is randomly selected, then a set of cardinality $|V| = S$ is sampled based on the eigenvectors.
Our optimization goal is defined by Eq.~(\ref{dppgoalnew}), however, estimating it takes exponential time cost. Furthermore, in Eq.~(\ref{dppgoalnew2}), the normalization could be given by the $S$-th elementary symmetric polynomial evaluated at the eigenvalues (denote as $\lambda$). Note that $\lambda$ can be provided by the eigen-decomposition of $\mathbf{L}$.
We have

\begin{equation}
\label{det}
\begin{aligned}
\sum_{|\D_t^{'}|=S}\det(\mathbf{L}_{\D_t^{'}}) & = \det(I + \mathbf{L})\sum_{|\D_t^{'}|=S}P_L(\D_t^{'}) \\
& = \sum_{|J|=S}\prod_{n \in J}\lambda_n.
\end{aligned}
\end{equation}

Defining the $S$-th elementary symmetric polynomial as $e^N_S = \sum_{|J|=S}\prod_{n \in J}\lambda_n$, and substituting Eq.~(\ref{det}) into Eq.~(\ref{dppgoalnew}) yields

\begin{equation}
\label{det2}
P_L^S 
 = \frac{1}{e^N_S}\det(I+\mathbf{L})P_L 
= \frac{1}{e^N_S} \sum_{|J|=S}P^{V_J}\prod_{n \in J}\lambda_n,
\end{equation}
where $P^{V_J}$ refers to an elementary DPP and $V_J$ denotes the set of orthonormal vectors $\{v_n\}_{n \in J}$. An elementary DPP refers to the DPP if every eigenvalue of its marginal kernel is in $\{0, 1\}$. Hence we can sample from a $k$DPP by sampling from index sets (denoted as $J$) according to the mixture components. Here we could observe that the ``L-ensemble'' can be interpreted as a mixture of elementary DPPs, in which their mixture weight is given by the products of eigenvalues ($\lambda_n$) relating to the eigenvectors $\{v_n\}_{n \in J}$ and normalized by the $\det(I+\mathbf{L}) = \prod^{N}_{n=1}(1+\lambda_n)$. Eq.~(\ref{dppgoalnew}) can be solved by Newton's identities.

The marginal probability of index $N$ is

\begin{equation}
\label{det3}
Pr(N \in J) 
 = \frac{\lambda_N}{e^N_S} \sum_{\substack{{J^{'} \subseteq \{1,...,N-1\}}\\ {|J^{'}|=S-1}}} \prod_{n \subseteq J^{'}} \lambda_n
= \lambda_N \frac{e^{N-1}_{S-1}}{e^N_S}.
\end{equation}
%
%
In the $k$DPPs sampling process, we firstly select $J$ with Eq.~(\ref{det3}), then sample data from $P^{V_J}$. The main steps of our proposed algorithm are summarized in Algorithm~\ref{dppalg}.

\renewcommand{\algorithmicrequire}{ \textbf{Input:}} 
\renewcommand{\algorithmicensure}{ \textbf{Output:}} 

\begin{algorithm}[tb]
\caption{AL with $k$DPPs}
\label{dppalg}
\begin{algorithmic}[1]
\REQUIRE
initial labeled dataset $\D_l$, unlabeled data pool $\D_u$, budget $B$, batch size $S$, a committee of classifiers $\C=\{{\theta}^{(1)},...,{\theta}^{(C)}\}$. \\
\ENSURE $\D_l$.
\REPEAT
\STATE Compute informativeness score $\bm{\mu}$ by Eq.~(\ref{info}).
\STATE Compute representativeness vector $\mathbf{r}_i$ of each unlabeled data sample $i$ by Eq.~(\ref{rep}).
\STATE Compute similarity score matrix $\mathbf{S}$ by Eq.~(\ref{Laplacian}) 
.
\STATE Construct L-matrix, $\mathbf{L}=\diag(\bm{\mu}) \mathbf{S} \mathbf{S}^T \diag(\bm{\mu})$.
\STATE Obtain index set ($\D^t$) of $\D^{new}$ from $kDPPs(\mathbf{L},S)$.
\STATE Query labels of via $\D^t$ from \emph{oracle}.
\STATE $\D^{new}_l$ $\gets$ $\D_l \cup \D^t$.
\STATE $\D^{new}_u$ $\gets$ $\D_u \backslash \D^t$.
\STATE Update $\C$ with new $\D^{new}_l$.
\UNTIL{$B$ is exhausted}
\end{algorithmic}
\end{algorithm}

\subsection{Computational Complexity}
The time complexity of each iteration of the informativeness criterion is $O(T(N)+CN)$, where $N$ is the number of unlabeled samples in that iteration, and $T(\cdot)$ refers to the classifiers' total training time. In the GLAD based selection strategy, the time complexity is $O(CN)$. We employ EM algorithm to estimate parameters, the computational complexity of the E-Step is linear in the number of instances and the total number of labels. For the M-Step, the values of the Q function and its gradient are computed iteratively. Computing each function is linear in the number of instances, number of labelers, and the total number of labels. The k-center algorithm can be implemented in $O(\kappa+(M+N)|\D^t_l|)$, if we assume that $d$ can be evaluated in constant time (in this paper, $d$ is evaluated before the active learning process starts). The time complexity of the diversity criterion is decided by the complexity of kDPPs, where sampling from a k-DPP requires $O(NS^2)$ (includes the pre-processing steps) and $S$ is the batch size in each iteration. The overall computational complexity of a completed multi-criteria active learning process is $O(\frac{B}{S}(T(N)+CN+(\kappa+(M+N)|\D^t_l|)+S^2N))$, where $\frac{B}{S}$ is the number of iterations required in the process.

\section{Experiments}
\label{exper}
In this section we evaluate the performance of our proposed AL algorithm on both synthetic and real-world datasets.

\subsection{Experiment Settings}
\subsubsection{Dataset Description} We consider both real-world and synthetic datasets for binary and multi-class classification task. Synthetic datasets help us explore the performance of models on various data distributions, while real-world datasets are employed to validate the effectiveness of our proposed model under real world situations. The real-world datasets are collected from UCI datasets \cite{Dua2019uci}, including \emph{Ionosphere}, \emph{Clean1}, \emph{Wdbc}, \emph{Sonar}, \emph{Tic-tac-toe} for binary classification, and \emph{iris}, \emph{Wine}, \emph{Glass} and \emph{Vehicle} datasets for multi-class classification. The synthetic datasets include \emph{R15} \cite{veenman2002maximum}, \emph{EX8a} (non-linearly separable), \emph{EX8b} (linearly separable) \cite{andrew2008stan}, Two-gaussian-clouds (\emph{GCloud}, including both balanced and unbalanced versions), in which each class follows a Gaussian distribution with a different mean and the same isotropic covariance \cite{konyushkova2017learning}. A summary of these datasets is shown in Table~\ref{d}.

\begin{table}[tb]
\footnotesize
\centering
\caption{Datasets used in the experiments. $(d, n, K)$ are the feature dimension, number of samples, and number of classes. The Imbalance Ratio (IR) is the ratio of the number of samples in the majority class to that of the minority class.}
\label{d}
{\begin{tabular}{@{}llc@{\hspace{0.2cm}}c@{\hspace{0.2cm}}c@{\hspace{0.2cm}}c@{}}
\hline
Dataset & Property & $d$ & $n$ & $K$ & Imbalance Ratio \\
\hline
Ionosphere & Real & $34$ & $351$ & $2$ & $1.8$\\
Clean1 & Real & $168$ & $475$ & $2$ & $1.3$\\
Wdbc & Real & $30$ & $569$ & $2$ & $1.7$\\
Sonar & Real & $60$ & $108$ & $2$ & $1.1$\\
Tic-tac-toe & Real & $18$ & $958$ & $2$ & $6.8$ \\
Iris & Real & $4$ & $150$ & $3$ & $1.0$\\
Wine & Real & $13$ & $178$ & $3$ & $1.5$\\
Glass & Real & $9$ & $214$ & $6$ & $8.4$\\
Vehicle & Real & $18$ & $846$ & $4$ & $1.1$\\
R15 & Synthetic & $2$ & $600$ & $15$ & $1.0$ \\
EX8a & Synthetic & $2$ & $863$ & $2$ & $1.0$\\
EX8b & Synthetic & $2$ & $206$ & $2$ & $1.0$\\
Gcloud Balance & Synthetic  & $2$ & $1,000$ & $2$ & $1.0$ \\
Gcloud Unbalance & Synthetic & $2$ & $1,000$ & $2$ & $1.8$\\
\hline
\end{tabular}}
\end{table}


\subsubsection{Baselines and Protocols} We compare our model against six baselines: (1) Uniform Sampling (\textbf{Uniform}); (2) Reinforced Active Learning Formulation (\textbf{Graph}), a diversity promoting sampling method that uses graph density to determine most representative points \cite{ebert2012ralf}; (3) Hierarchical Clustering (\textbf{Hier}), which exploits cluster structure in data \cite{dasgupta2008hierarchical}; (4) Informative and diverse (\textbf{InfoDiv}) batch sampler, which samples points with small margin while maintaining same distribution over clusters as training set \cite{chattopadhyay2013batch}; (5) K-center Greedy (\textbf{KCenter}), which finds data points that minimize the maximum distance of any point to a center \cite{sener2017geometric}; 
(6) Margin-based AL (\textbf{Margin}), which selects a batch with the smallest margin / highest uncertainty \cite{balcan2007margin}.
These baselines are summarized in Table~\ref{baseline} and are implemented in an AL toolbox\footnote{https://github.com/google/active-learning}.

\begin{table}[tb]
\small
\centering
\caption{Baseline methods and their selection criteria.}
\label{baseline}
{\begin{tabular}{c|ccc}
\hline & Informativeness & Representativeness & Diversity\\
\hline
Uniform & $-$ & $-$ & $-$ \\
Graph & \checkmark& \checkmark& \checkmark \\
Hier & $-$ & \checkmark & \checkmark \\
InfoDiv & \checkmark & $-$ & \checkmark \\
KCenter & $-$ & \checkmark & \checkmark \\
Margin & \checkmark & $-$ & \checkmark \\
Ours & \checkmark & \checkmark & \checkmark \\
\hline
\end{tabular}}
\end{table}

\subsubsection{Implementation Details}

In our experiments, we randomly select $60\%$ instances for training and $40\%$ instances for testing. We select data samples from the training set and evaluate the classification performance on the testing set. We choose logistic regression (LR), Linear Discriminant Analysis (LDA), SVM with linear kernel (SVM-Linear), SVM with RBF kernel (SVM-RBF), SVM with Polynomial kernel (SVM-POLY) and Gaussian Process Classifier (GPC) as the committee in our informativeness strategy. To avoid bias problems, we have avoided any specific dataset tuning or pre-processing. For each dataset, we experimented with ever-increasing budgets. We repeat each experiment $10$ times with randomly split training and testing sets, and report the average testing performance. We adopted SVM-RBF as the classifier to test our method and the baselines. We initialize all the active learning process with $20$ labeled data points that are randomly selected from the training set. For batch setting, we set $S = \{1, 5, 10\}$. We use the Laplacian kernel in Eq.~(\ref{Laplacian}) when constructing the DPP kernel.

\subsubsection{Evaluation Metrics}
\label{evalmetric}
\begin{figure*} [tb]
\centering
\begin{minipage}{4.2cm}
\centering
\includegraphics[width=4.2cm]{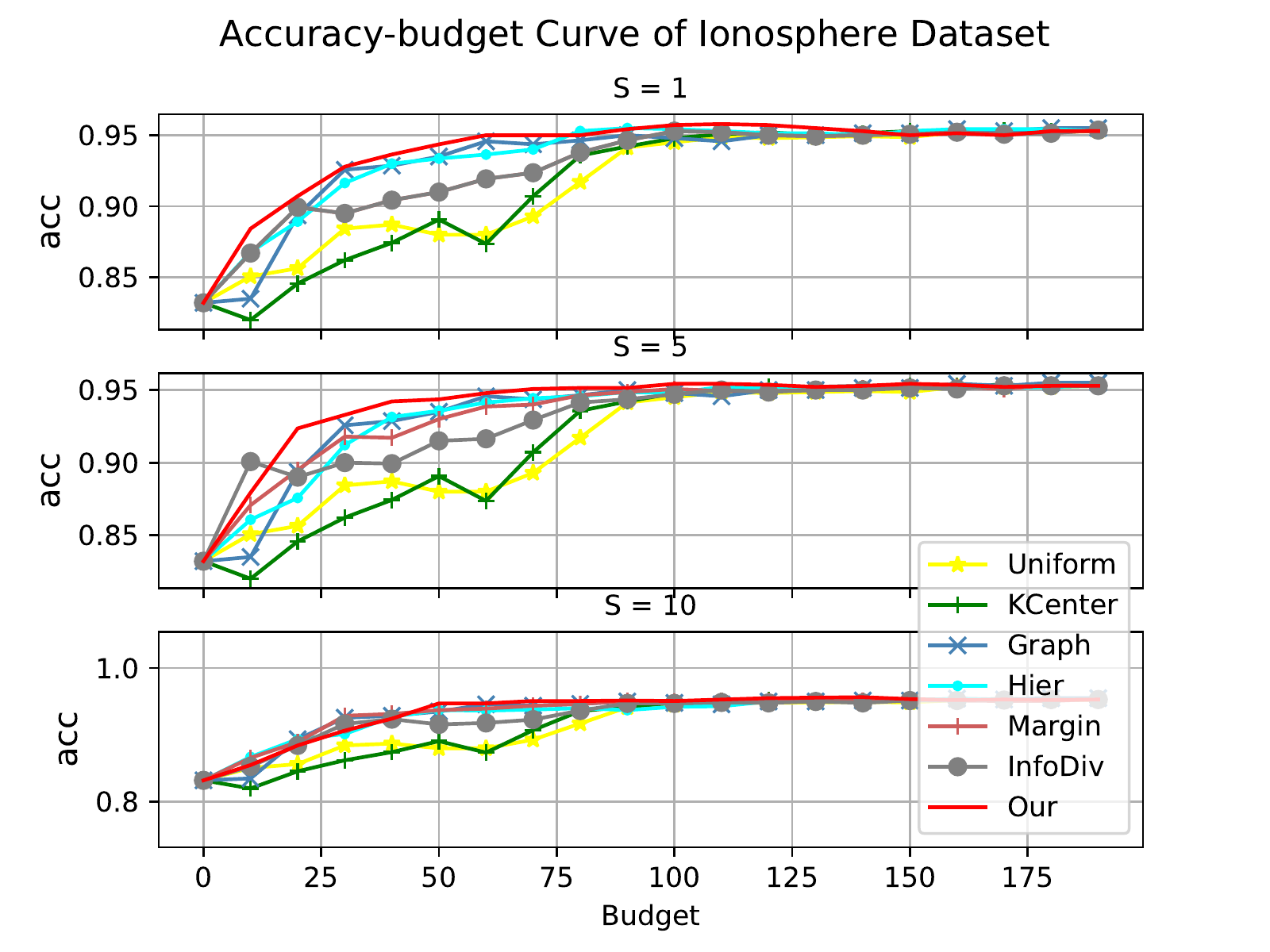}
\footnotesize(a) Ionosphere
\end{minipage}
\begin{minipage}{4.2cm}
\centering
\includegraphics[width=4.2cm]{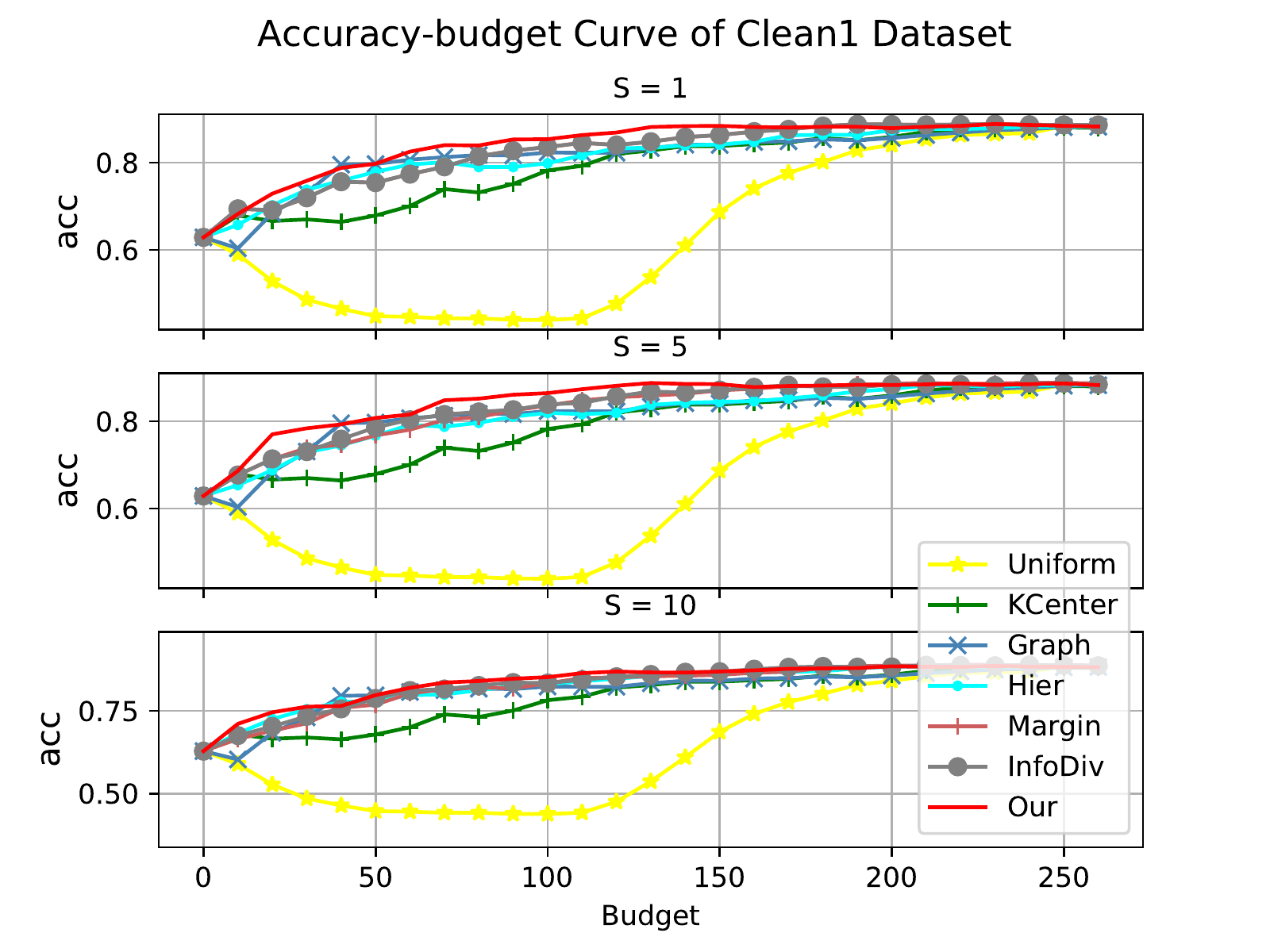}
\footnotesize(b) Clean1
\end{minipage}
\begin{minipage}{4.2cm}
\centering
\includegraphics[width=4.2cm]{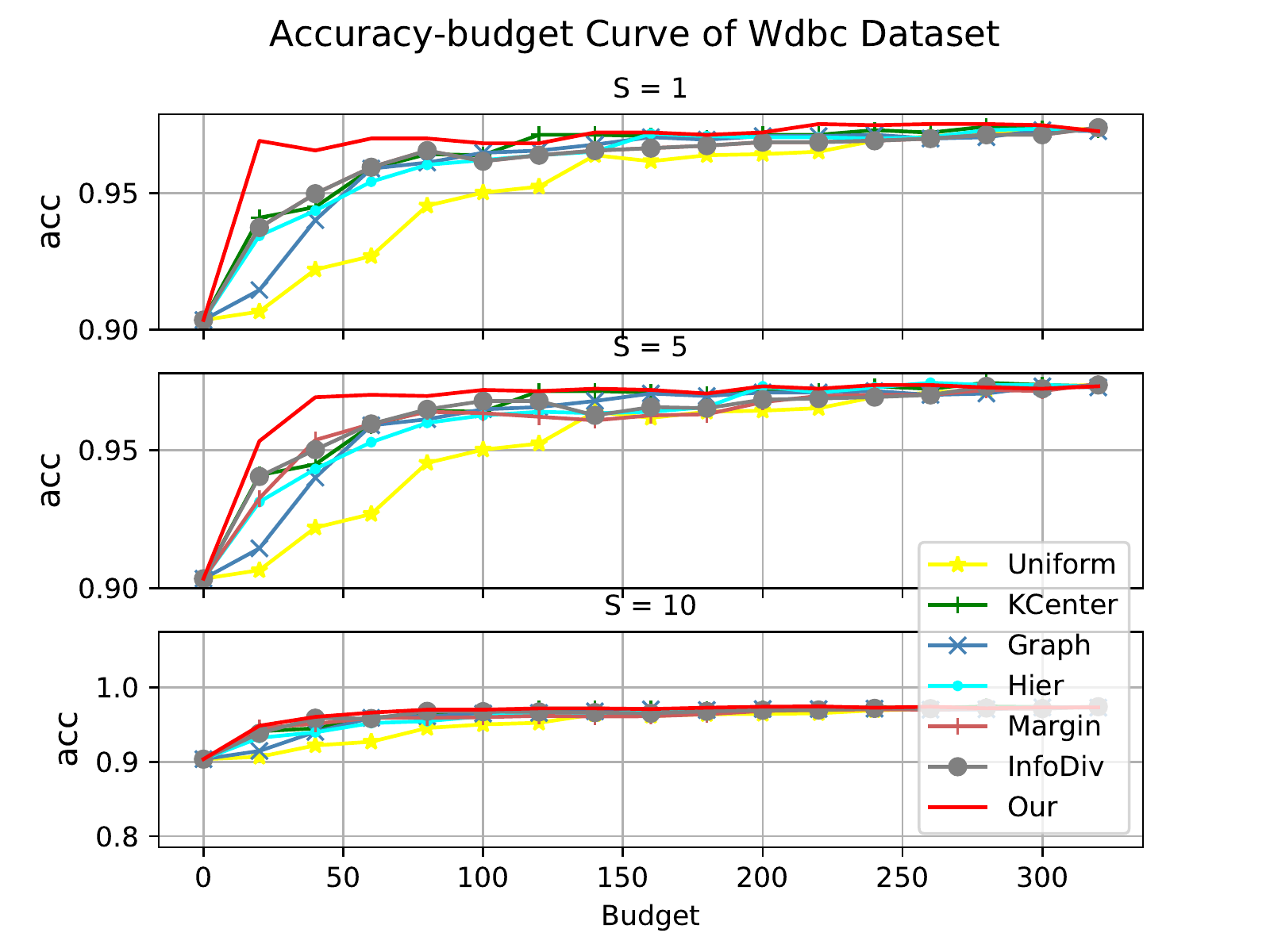}
\footnotesize(c) wdbc
\end{minipage}
\begin{minipage}{4.2cm}
\centering
\includegraphics[width=4.2cm]{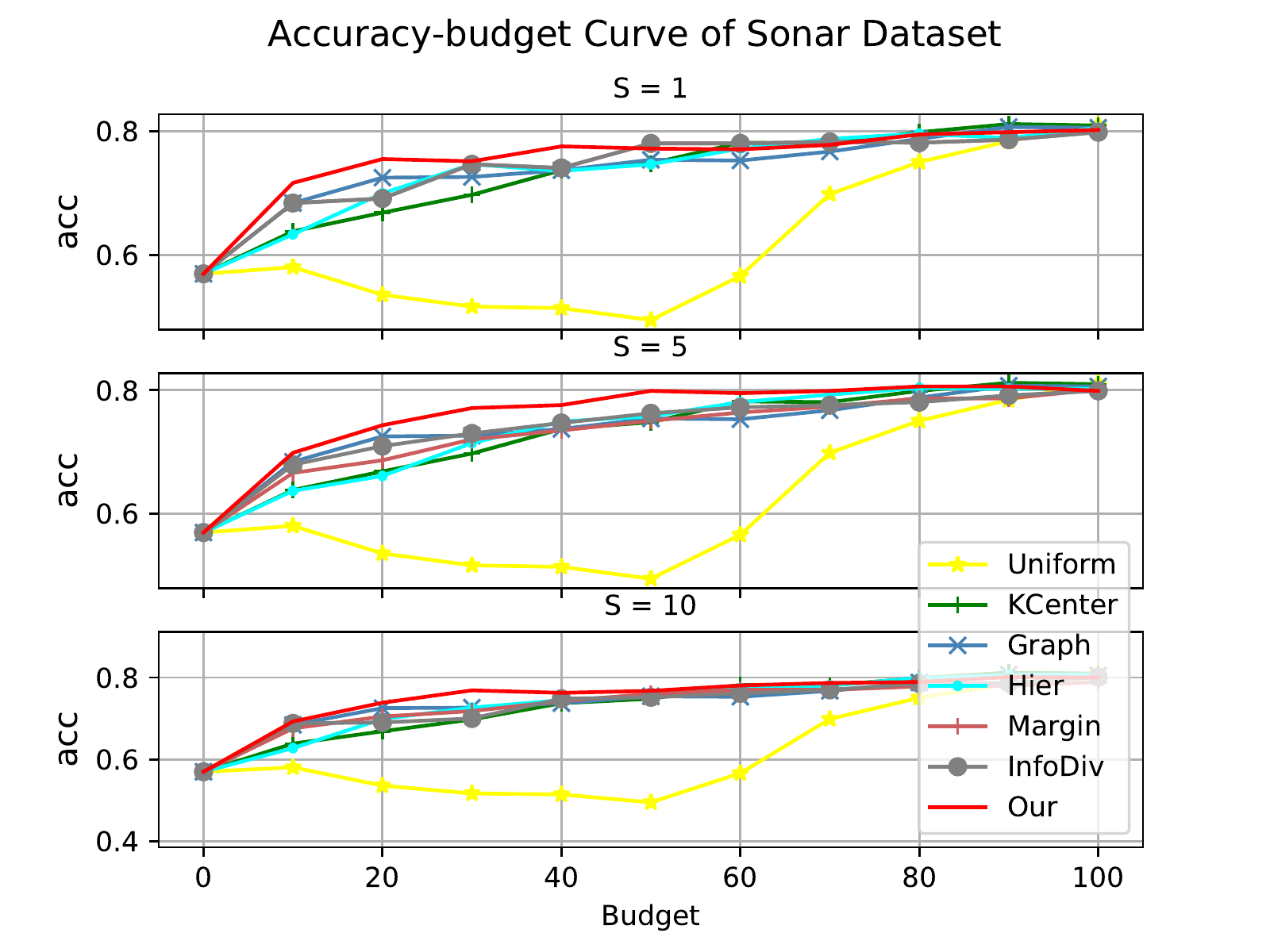}
\footnotesize(d) Sonar
\end{minipage}
\begin{minipage}{4.2cm}
\centering
\includegraphics[width=4.2cm]{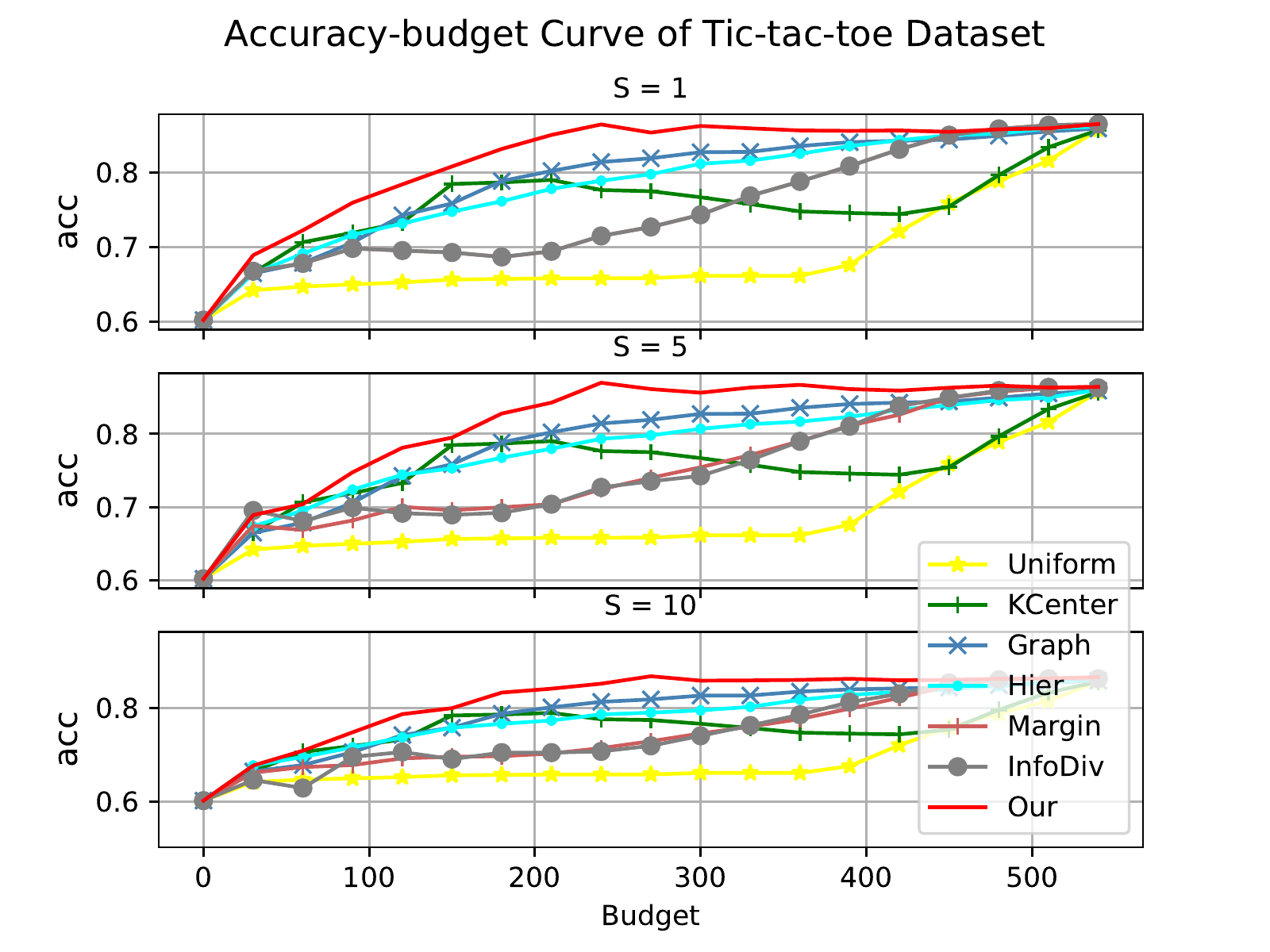}
\footnotesize(e) Tic-tac-toe
\end{minipage}
\begin{minipage}{4.2cm}
\centering
\includegraphics[width=4.2cm]{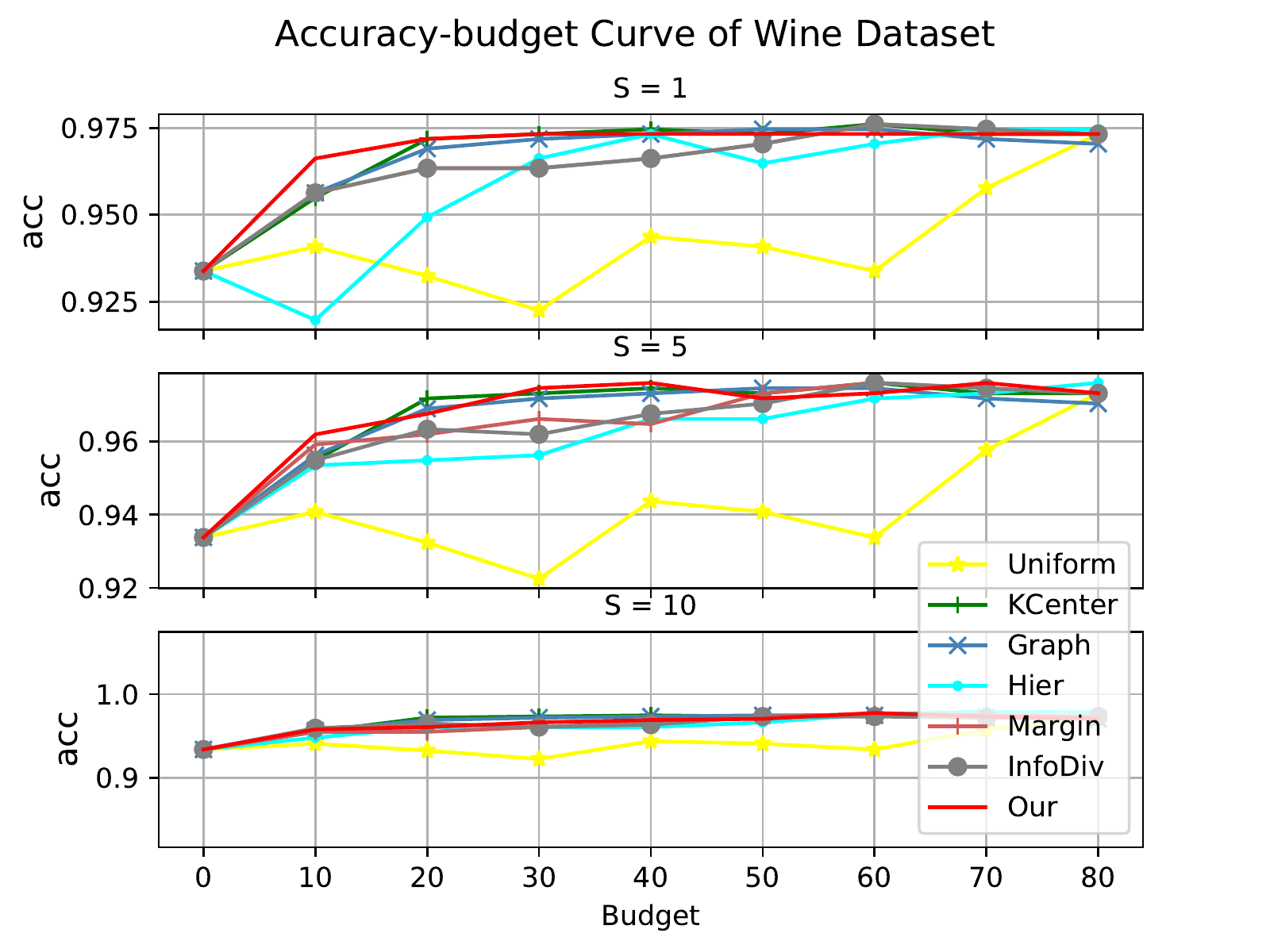}
\footnotesize(f) Wine
\end{minipage}
\begin{minipage}{4.2cm}
\centering
\includegraphics[width=4.2cm]{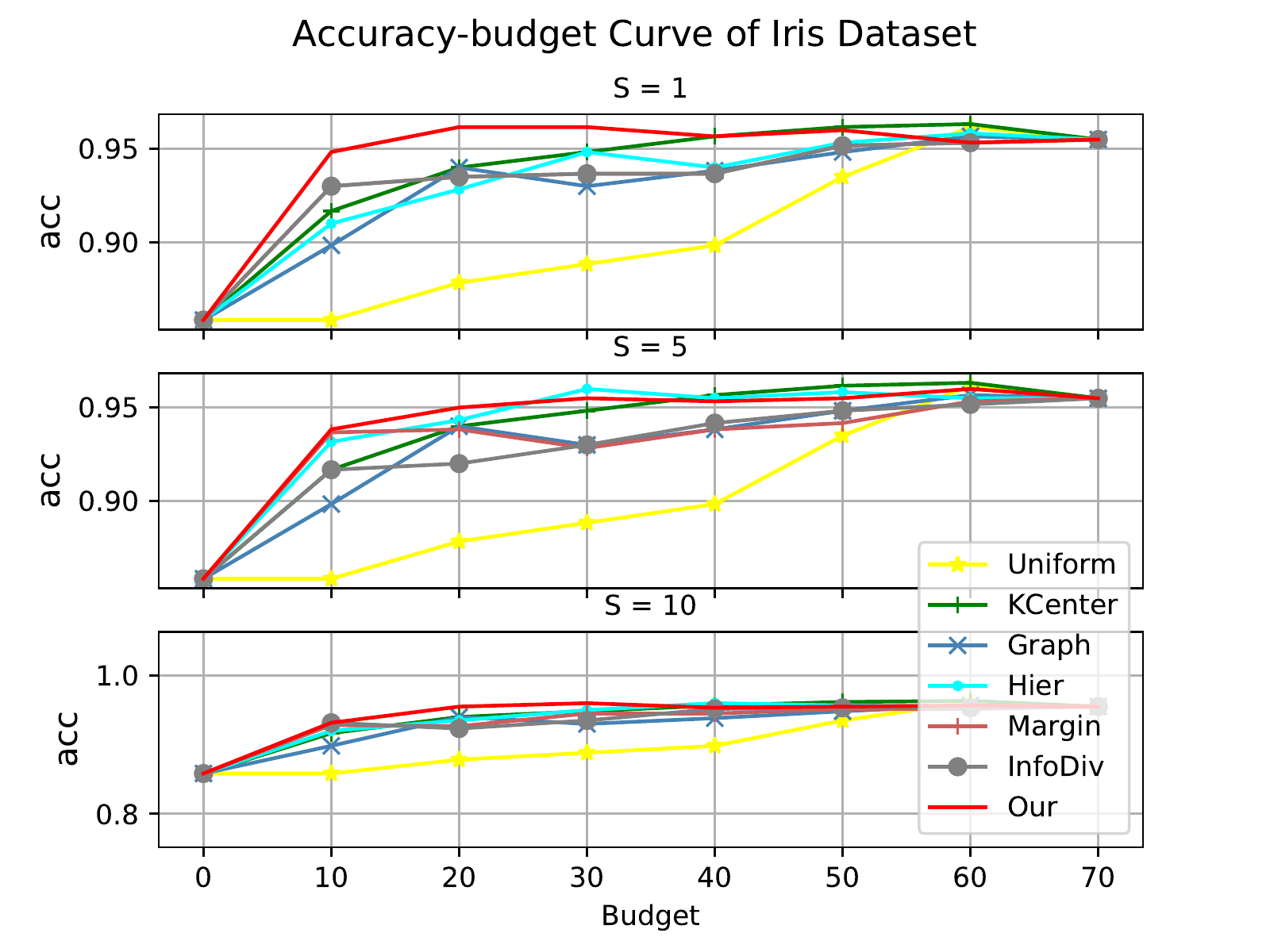}
\footnotesize(g) Iris
\end{minipage}
\begin{minipage}{4.2cm}
\centering
\includegraphics[width=4.2cm]{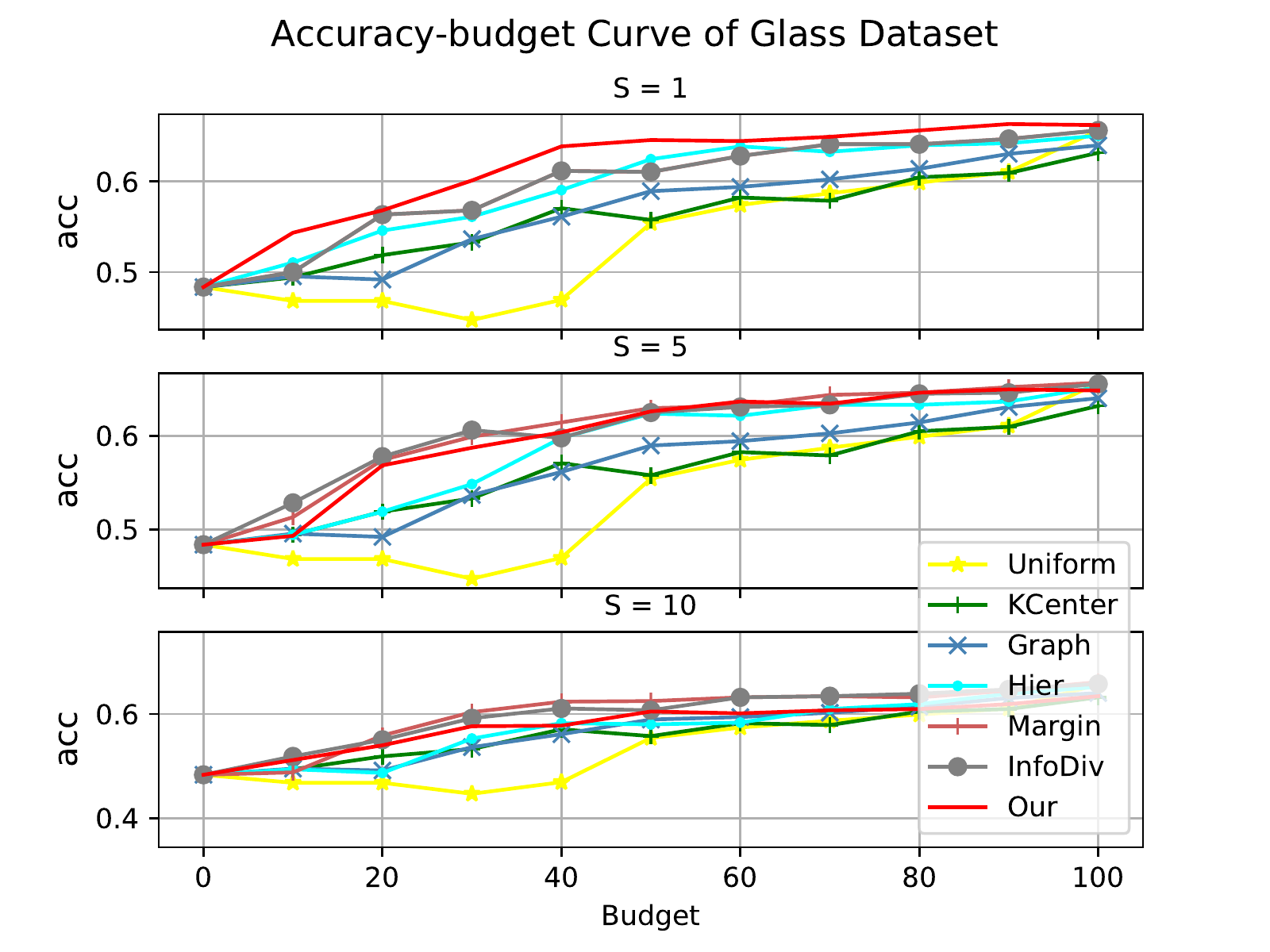}
\footnotesize(h) Glass
\end{minipage}
\begin{minipage}{4.2cm}
\centering
\includegraphics[width=4.2cm]{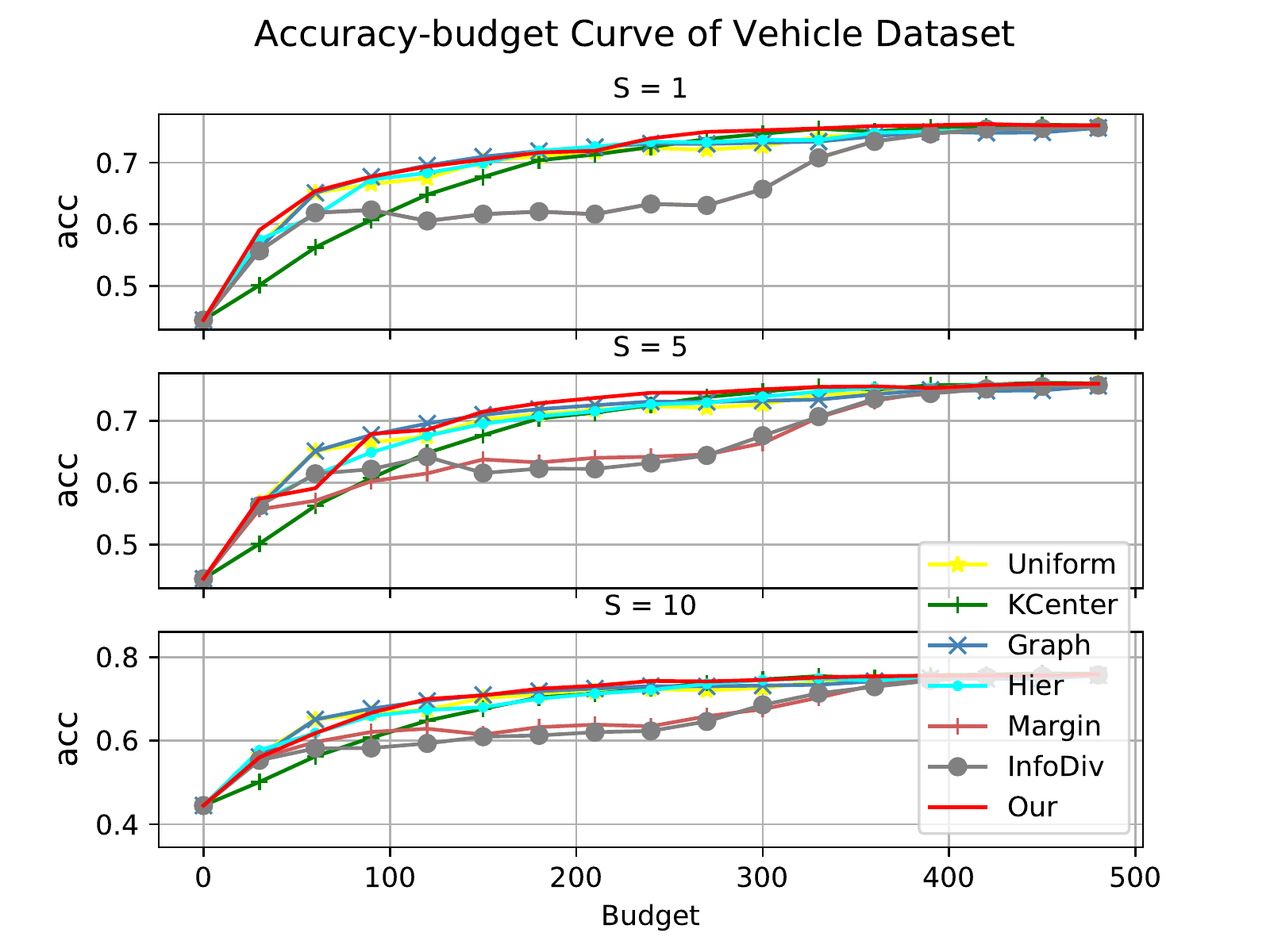}
\footnotesize(i) Vehicle
\end{minipage}
\begin{minipage}{4.2cm}
\centering
\includegraphics[width=4.2cm]{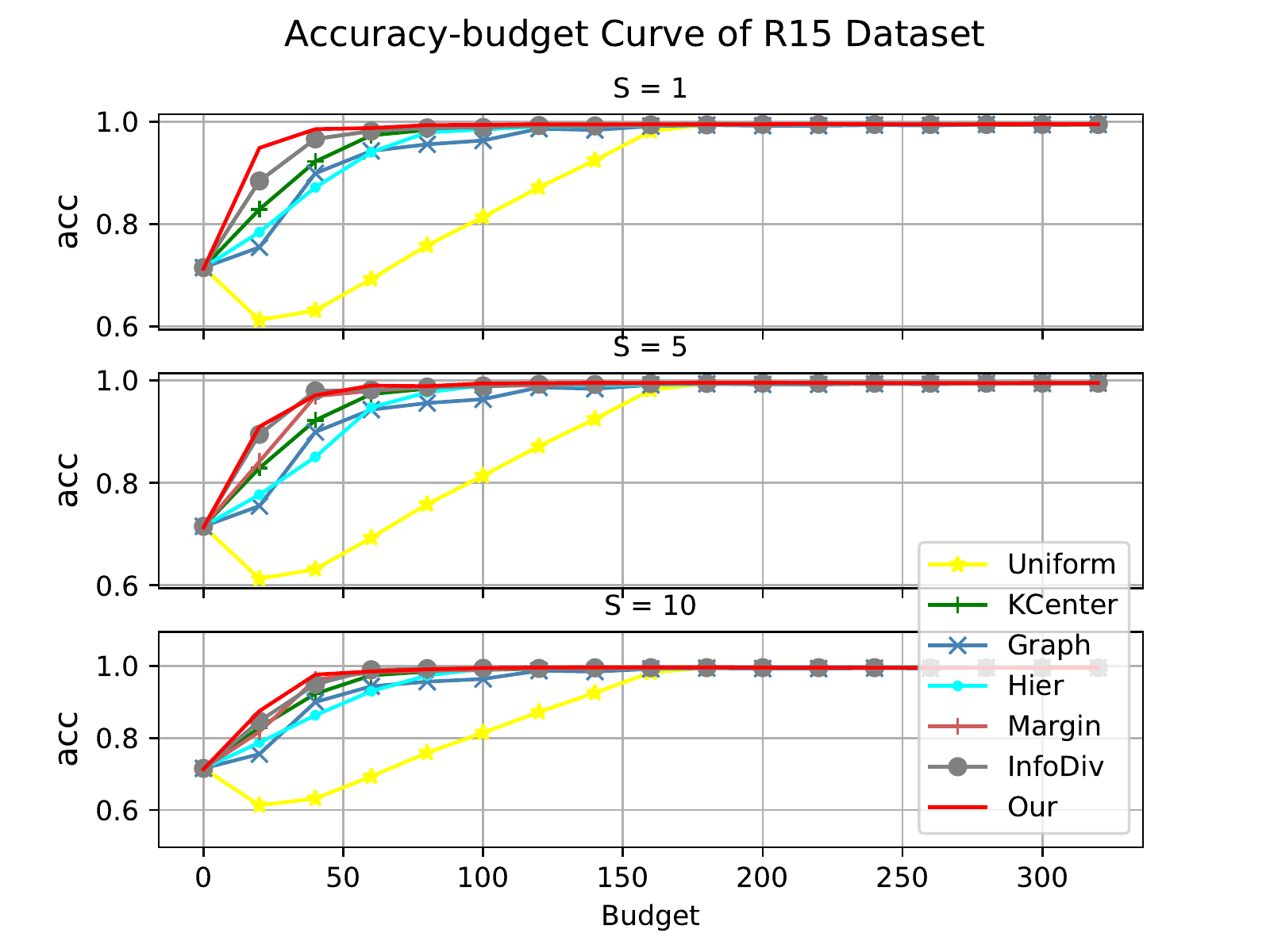}
\footnotesize(j) R15
\end{minipage}
\begin{minipage}{4.2cm}
\centering
\includegraphics[width=4.2cm]{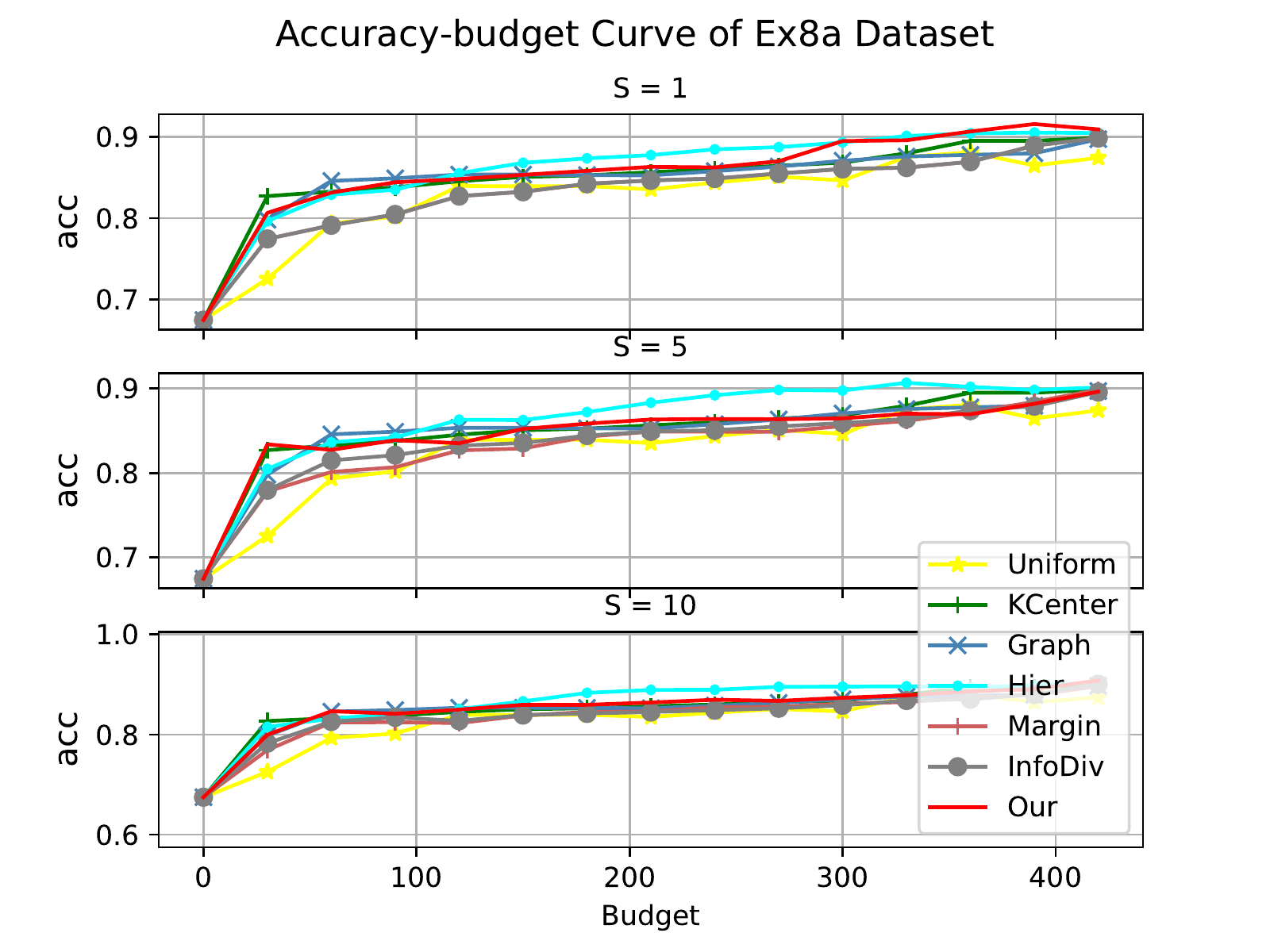}
\footnotesize(k) EX8a
\end{minipage}
\begin{minipage}{4.2cm}
\centering
\includegraphics[width=4.2cm]{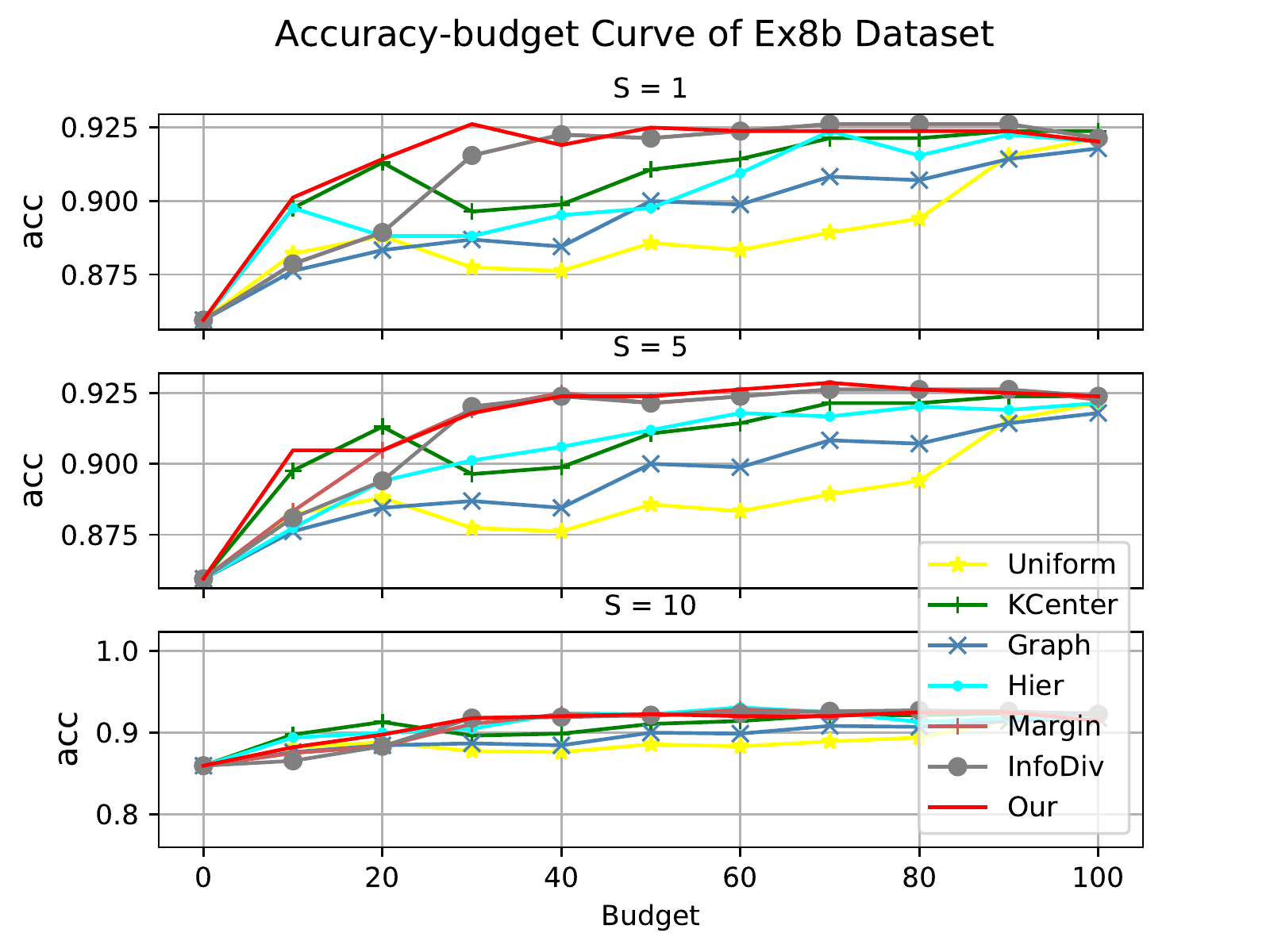}
\footnotesize(l) EX8b
\end{minipage}
\begin{minipage}{4.2cm}
\centering
\includegraphics[width=4.2cm]{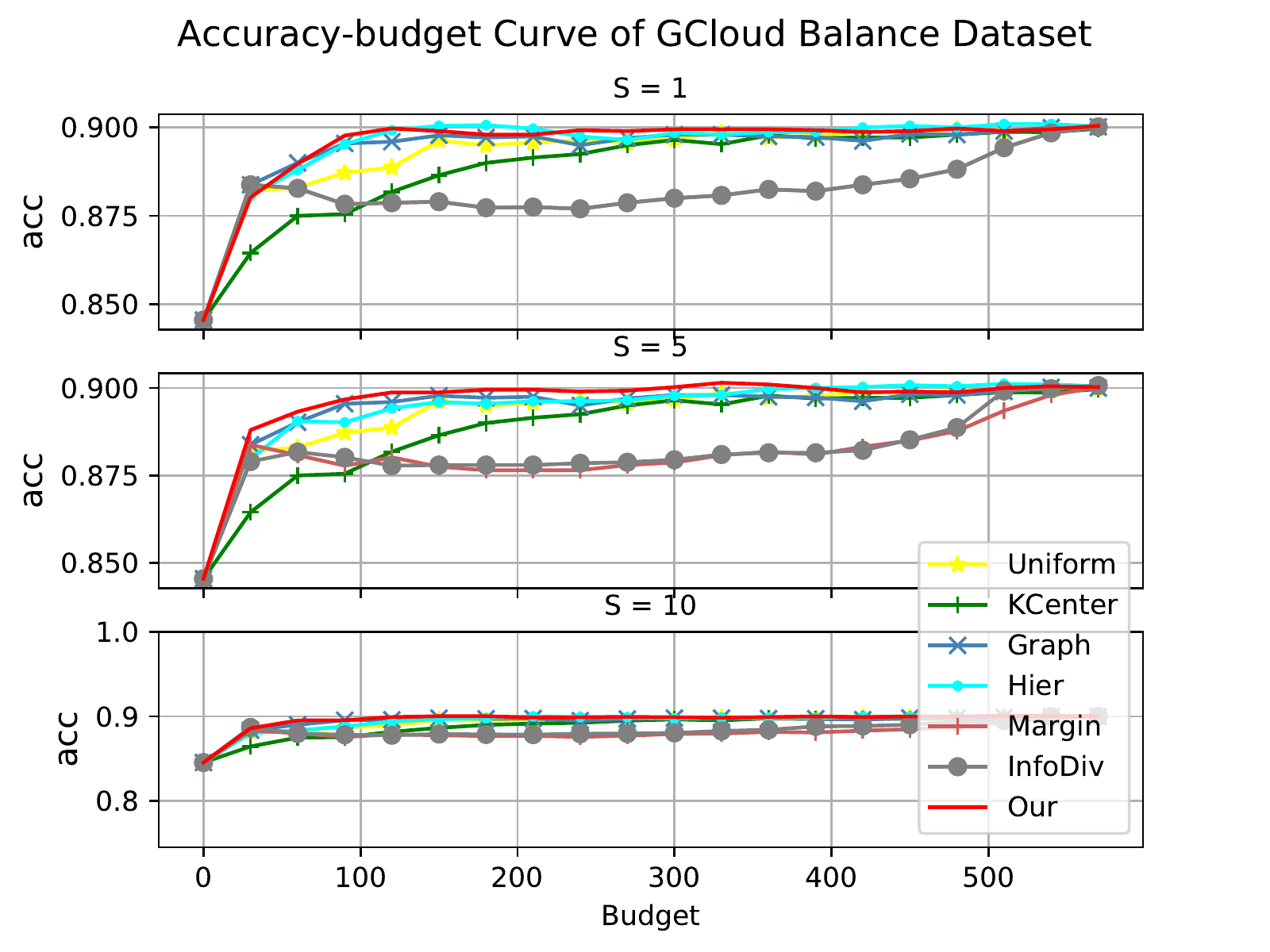}
\footnotesize(m) GCloud Balance
\end{minipage}
\begin{minipage}{4.2cm}
\centering
\includegraphics[width=4.2cm]{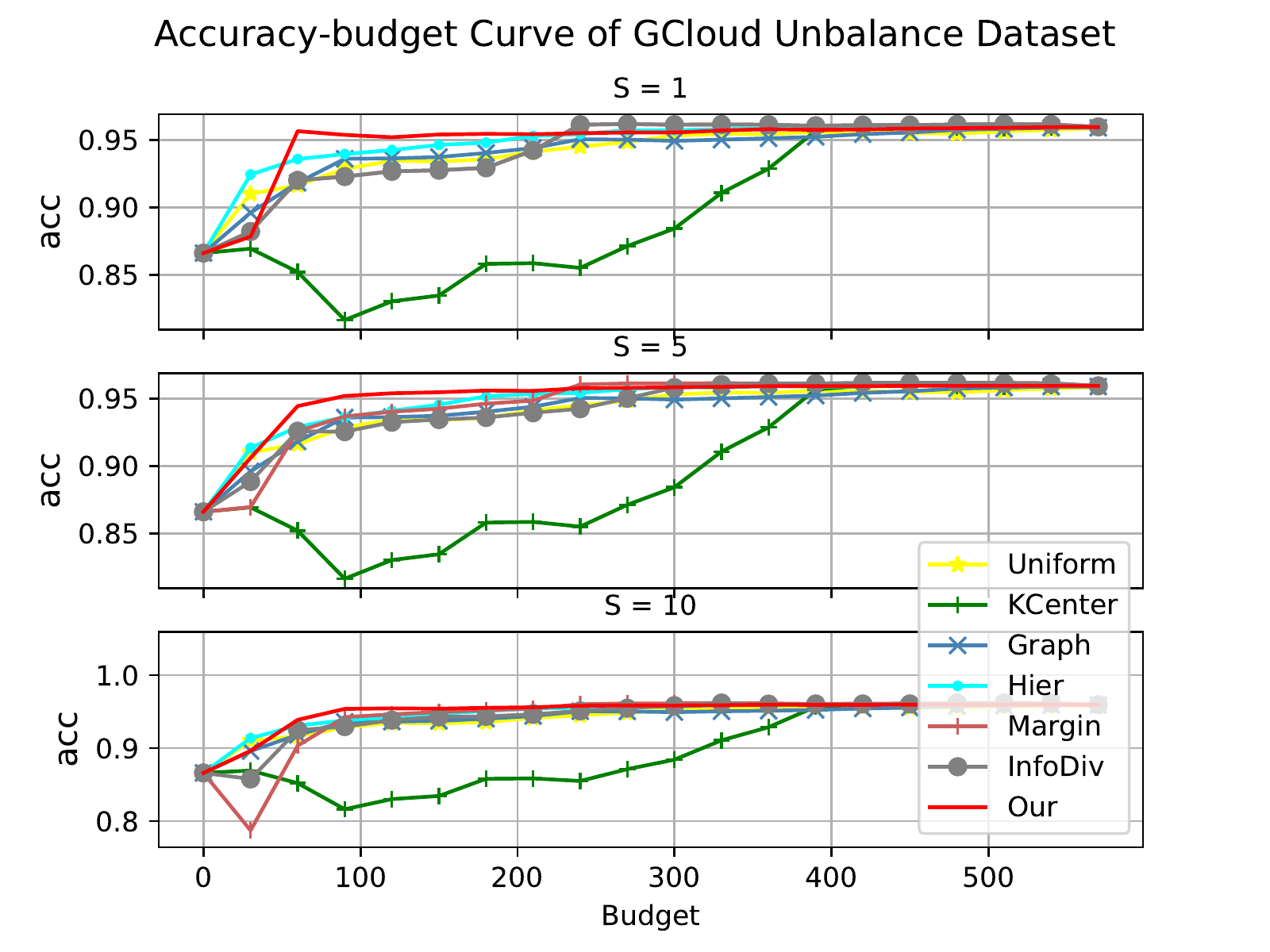}
\footnotesize(n) GCloud Unbalance
\end{minipage}
\caption{Accuracy vs.~budget curves on $14$ real-world and synthetic datasets ($S=\{1,5,10\}$).}
\label{acc-figure}
\end{figure*}

\begin{figure*} [tb]
\centering
\begin{minipage}{4.2cm}
\centering
\includegraphics[width=4.2cm]{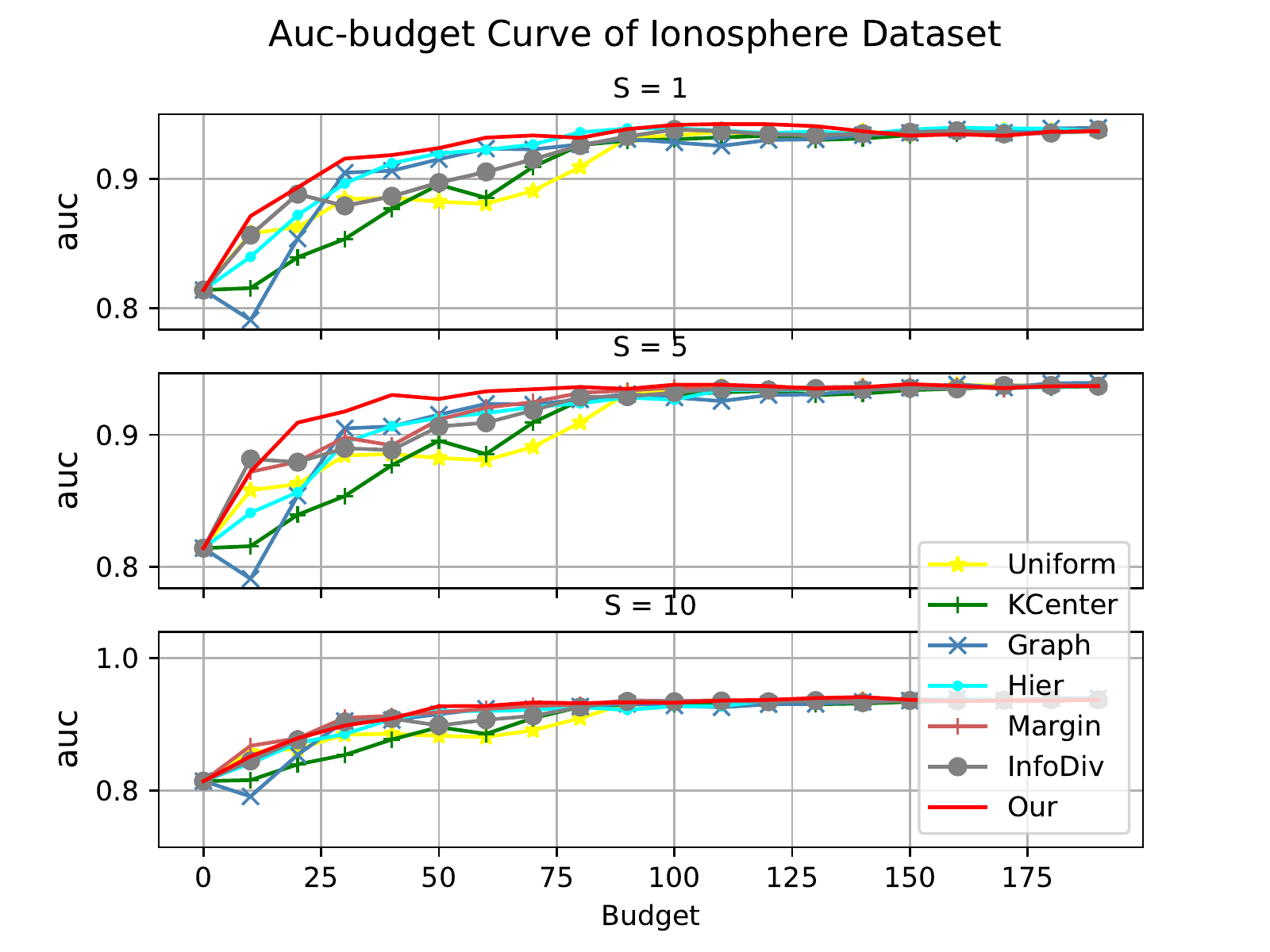}
\footnotesize(a) Ionosphere
\end{minipage}
\begin{minipage}{4.2cm}
\centering
\includegraphics[width=4.2cm]{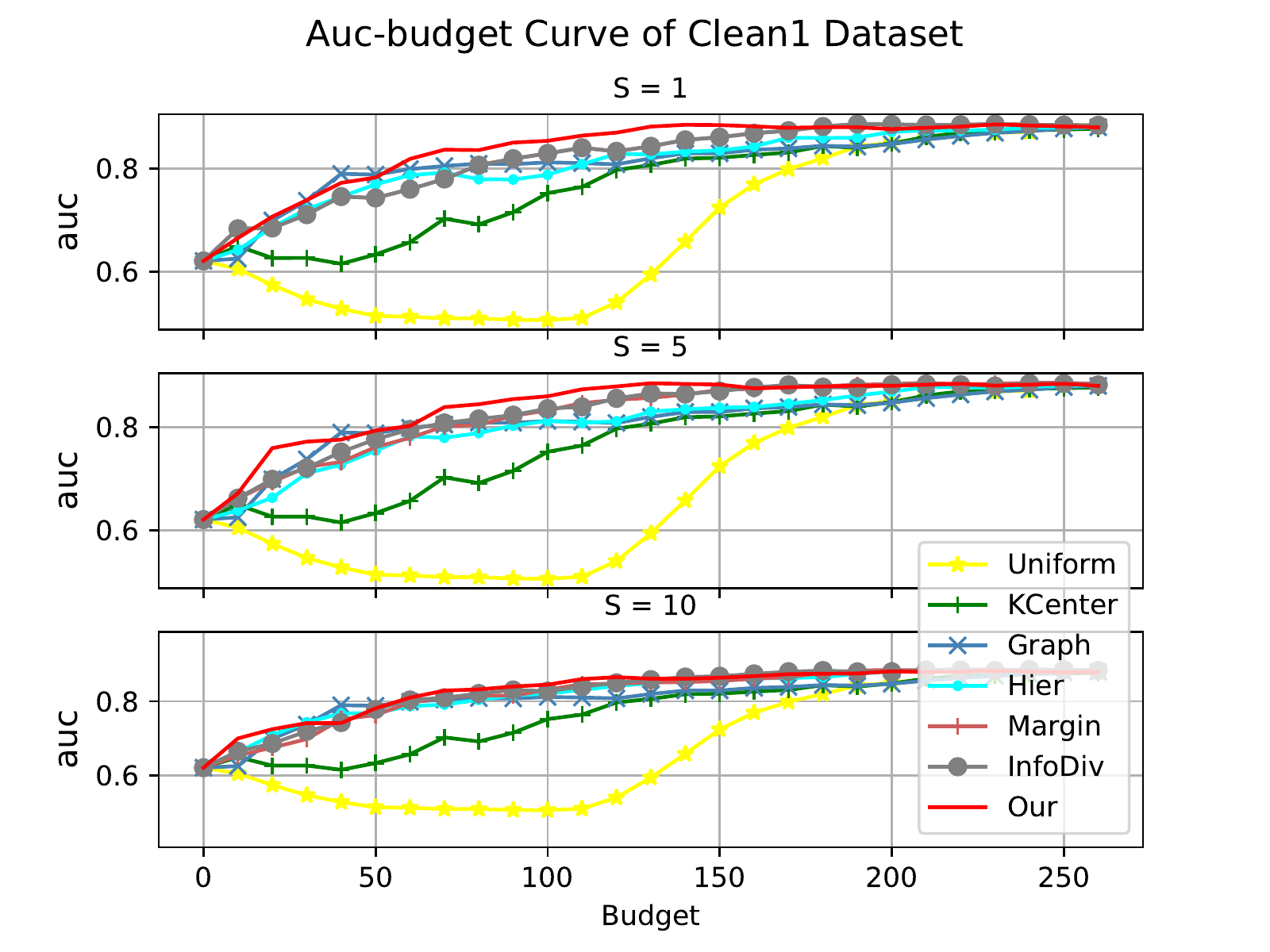}
\footnotesize(b) Clean1
\end{minipage}
\begin{minipage}{4.2cm}
\centering
\includegraphics[width=4.2cm]{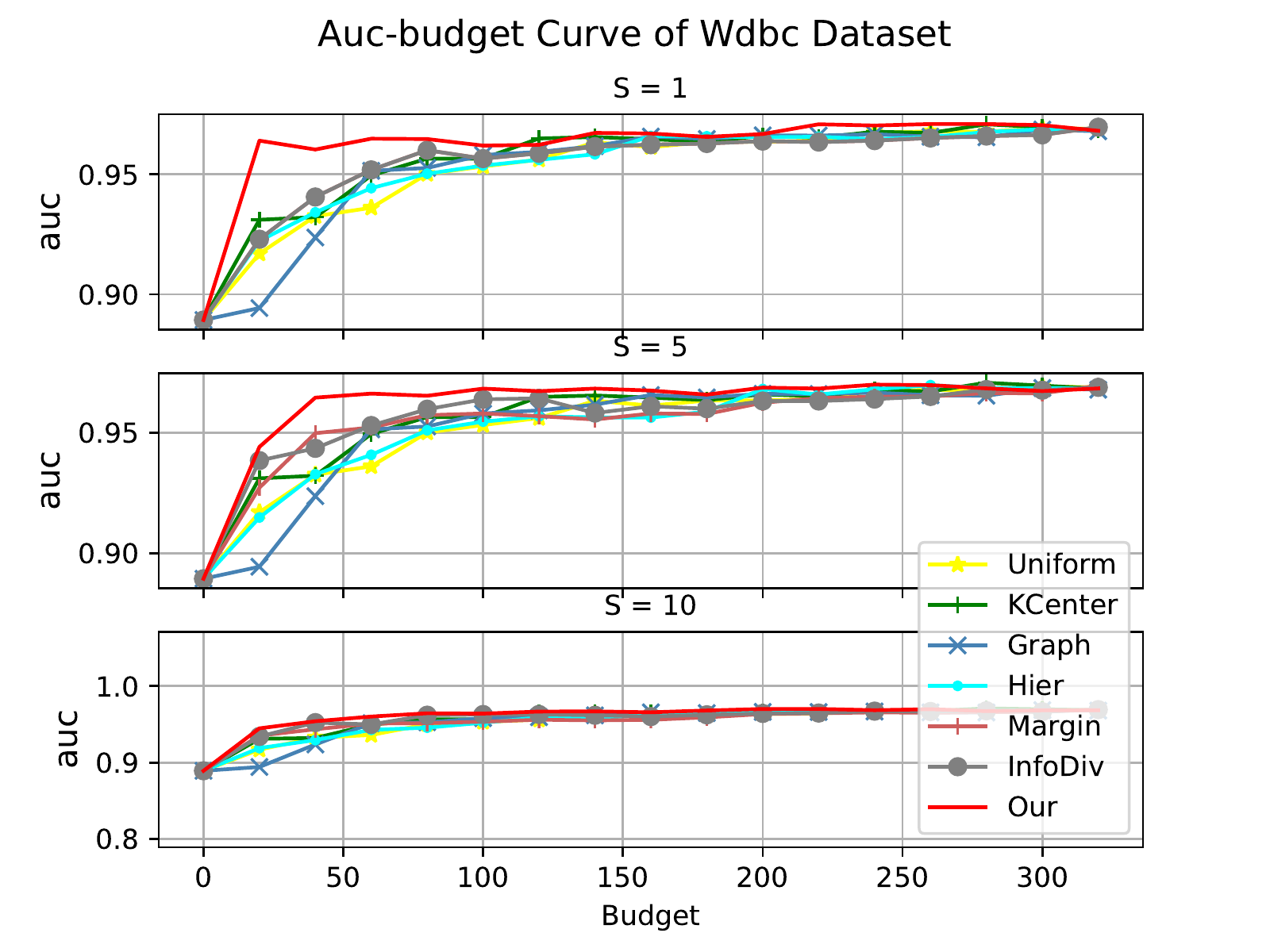}
\footnotesize(c) wdbc
\end{minipage}
\begin{minipage}{4.2cm}
\centering
\includegraphics[width=4.2cm]{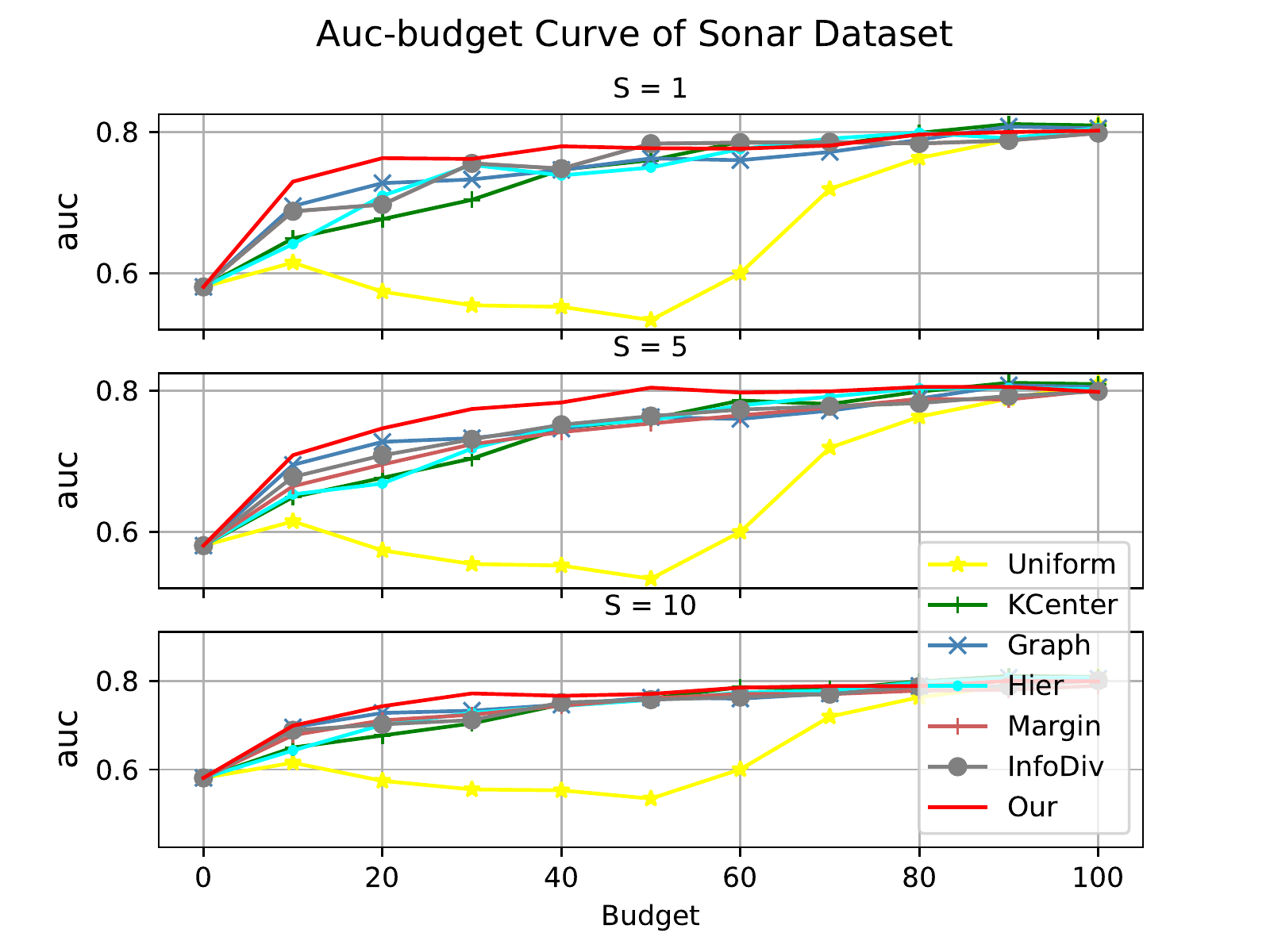}
\footnotesize(d) Sonar
\end{minipage}
\begin{minipage}{4.2cm}
\centering
\includegraphics[width=4.2cm]{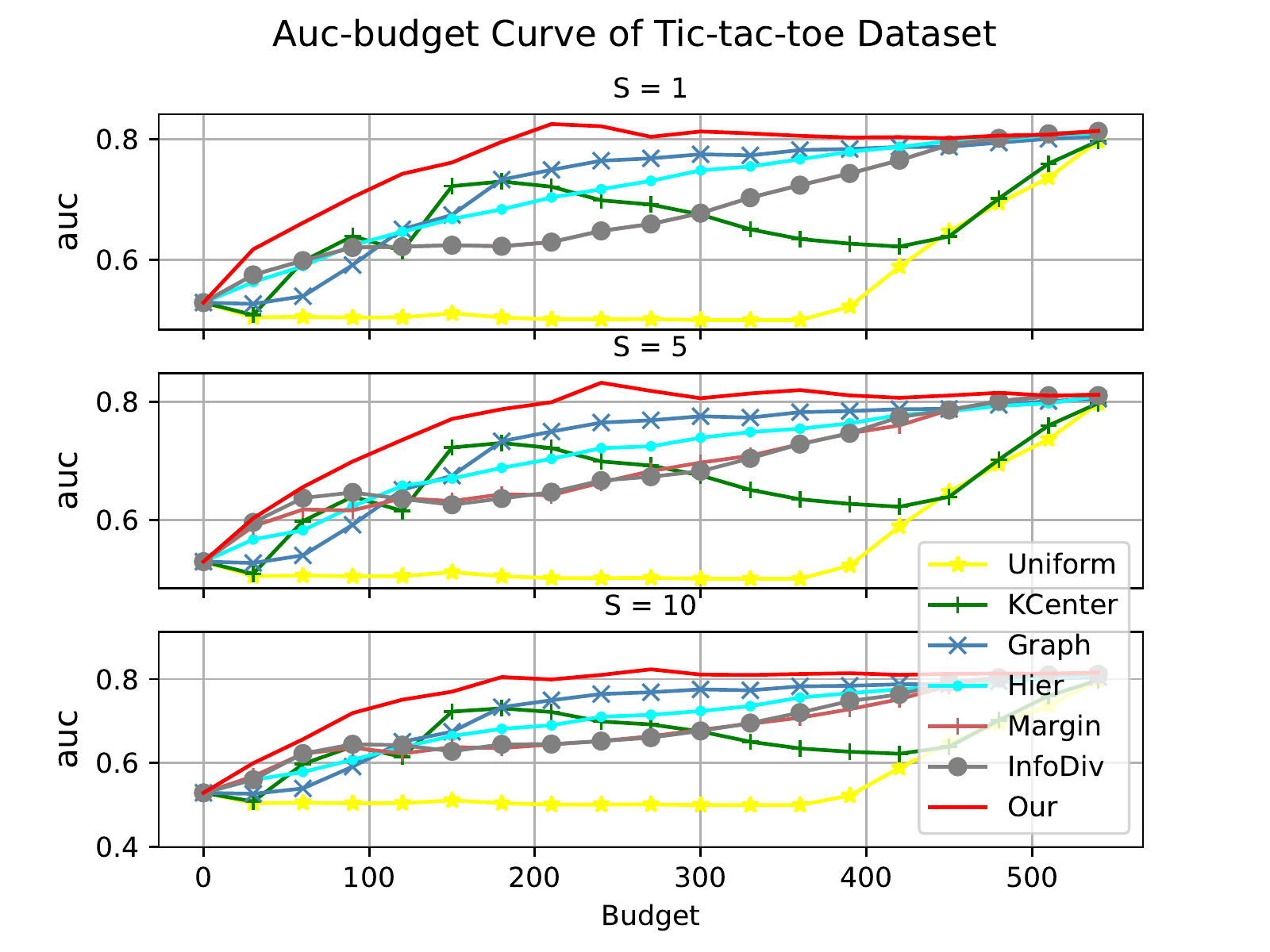}
\footnotesize(e) Tic-tac-toe
\end{minipage}
\begin{minipage}{4.2cm}
\centering
\includegraphics[width=4.2cm]{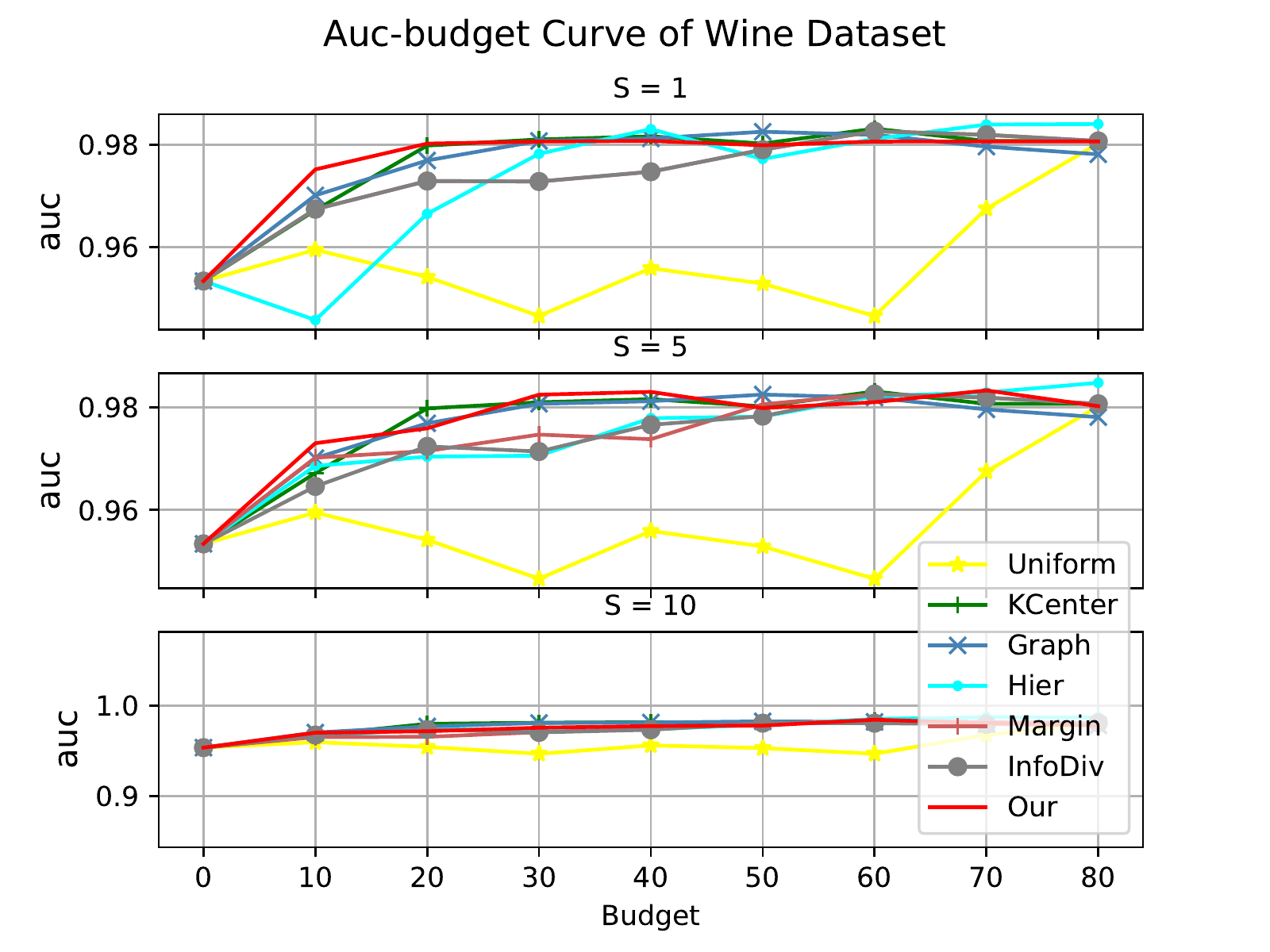}
\footnotesize(f) Wine
\end{minipage}
\begin{minipage}{4.2cm}
\centering
\includegraphics[width=4.2cm]{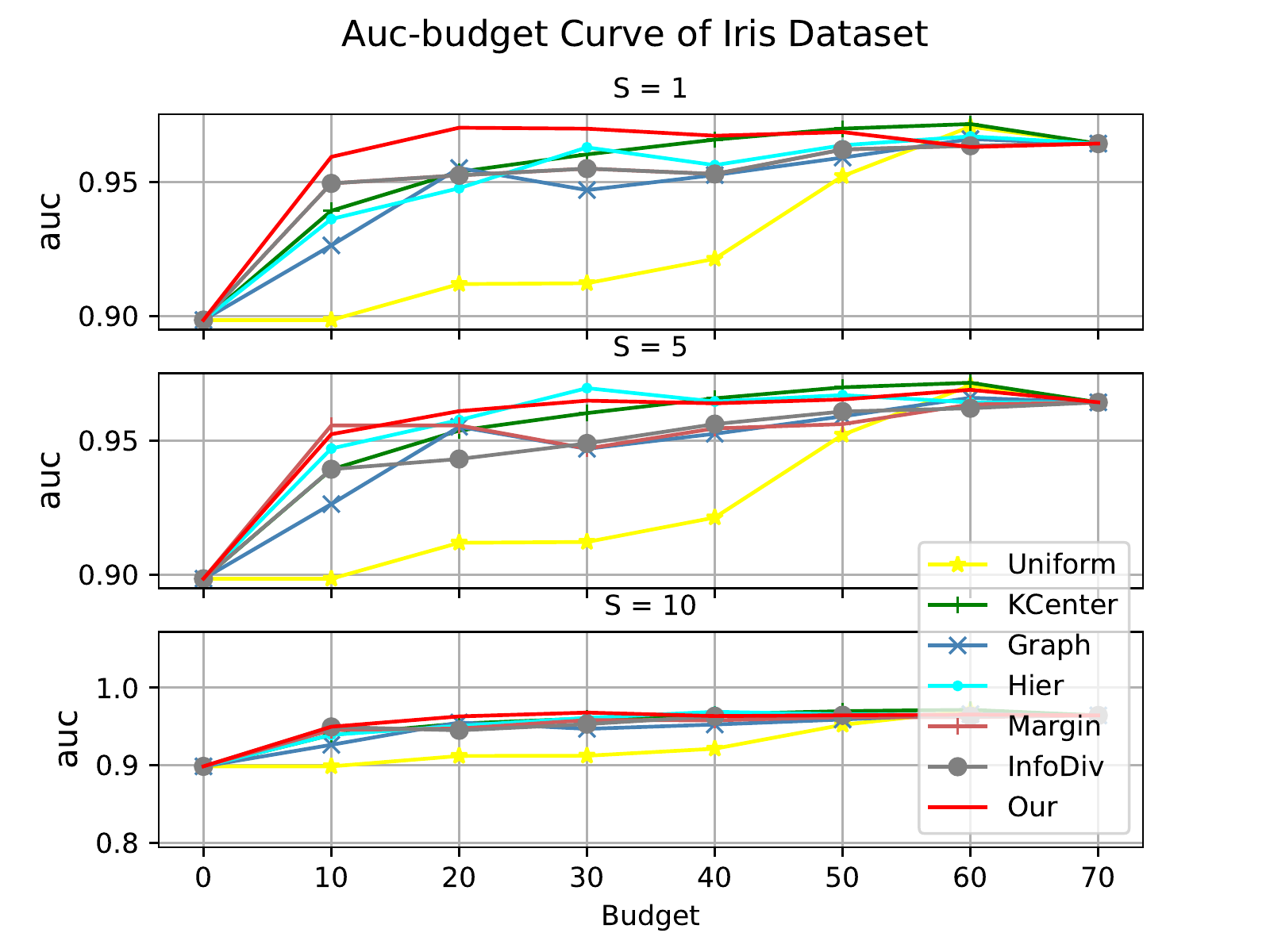}
\footnotesize(g) Iris
\end{minipage}
\begin{minipage}{4.2cm}
\centering
\includegraphics[width=4.2cm]{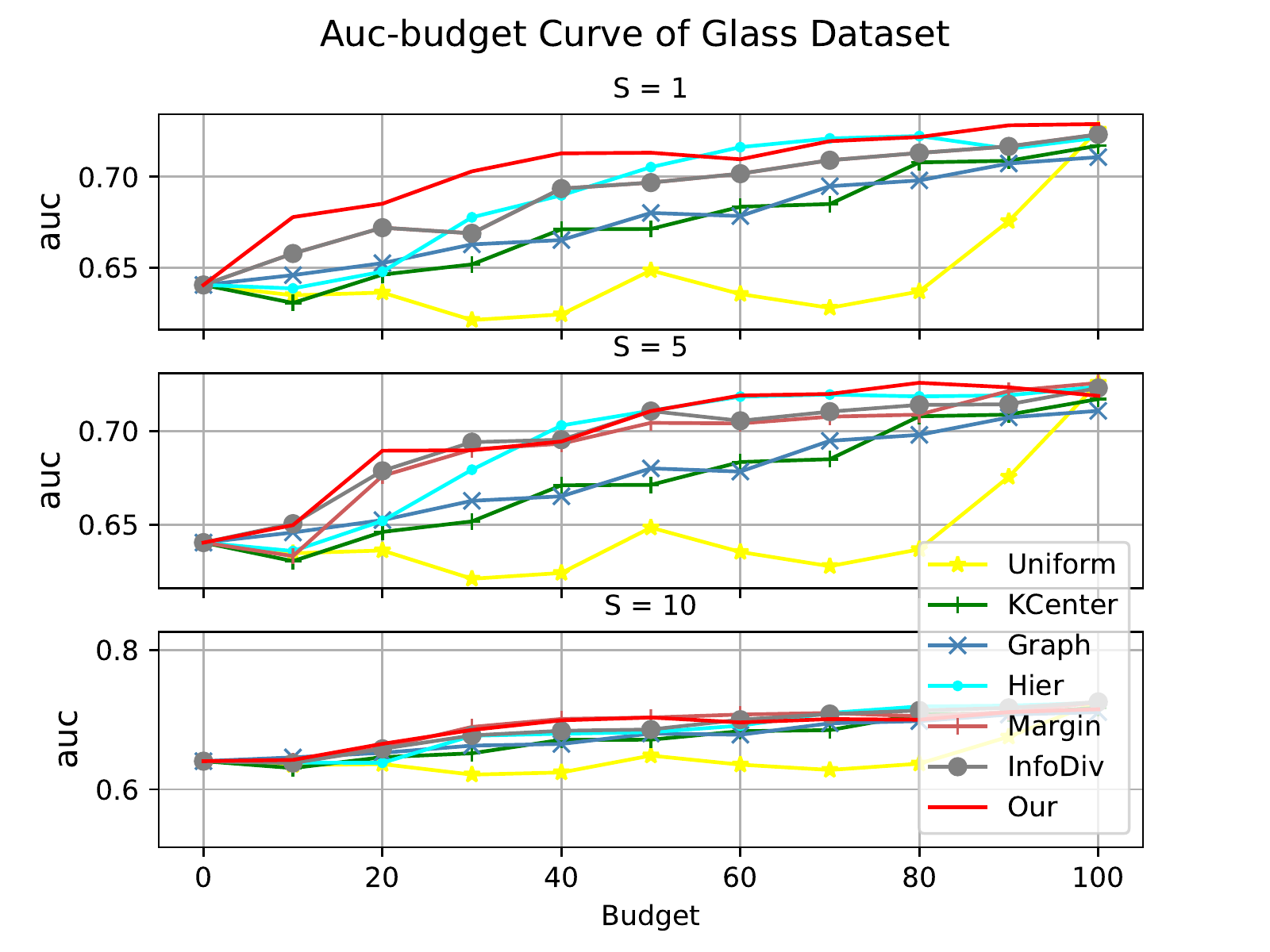}
\footnotesize(h) Glass
\end{minipage}
\begin{minipage}{4.2cm}
\centering
\includegraphics[width=4.2cm]{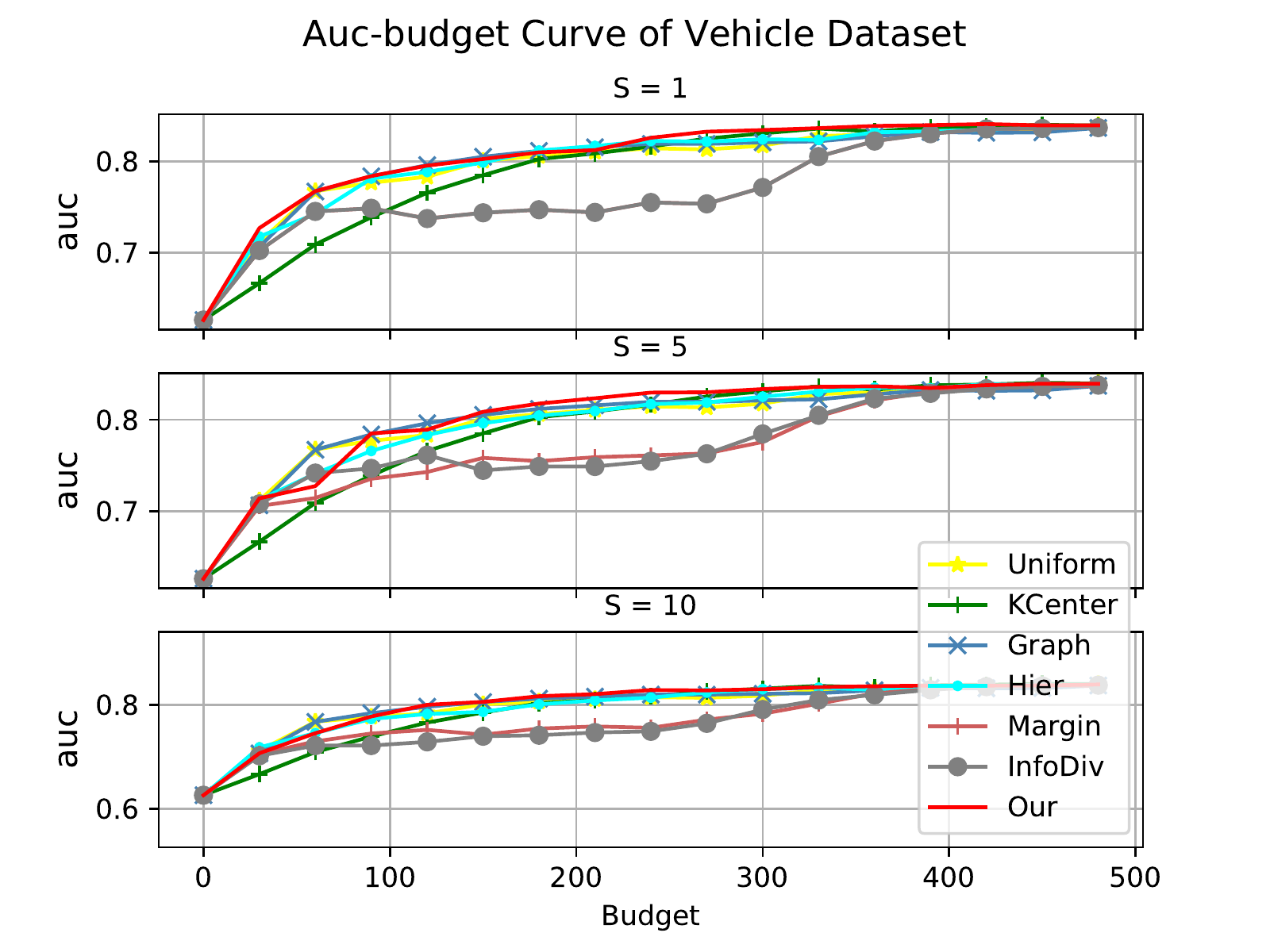}
\footnotesize(i) Vehicle
\end{minipage}
\begin{minipage}{4.2cm}
\centering
\includegraphics[width=4.2cm]{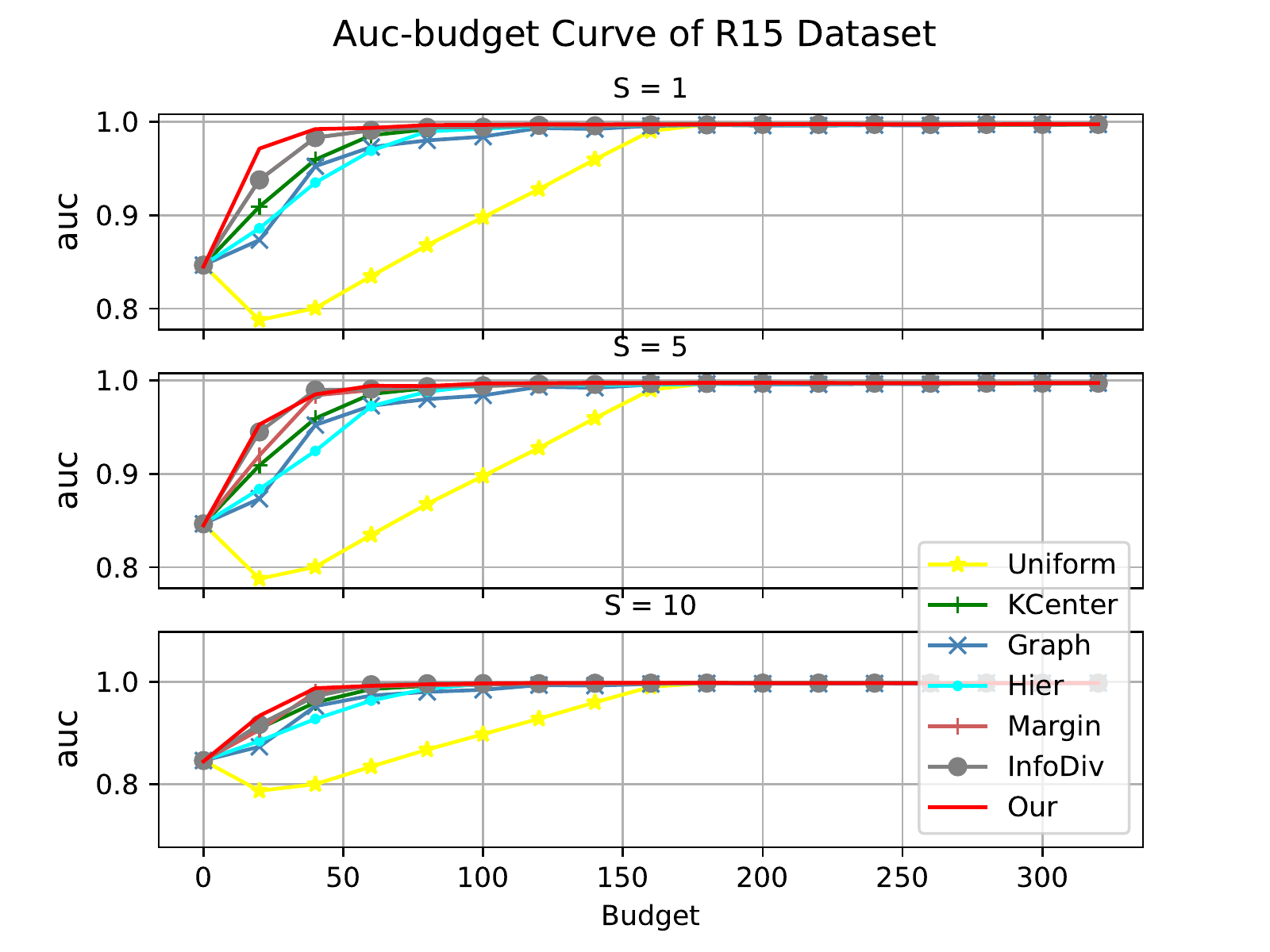}
\footnotesize(j) R15
\end{minipage}
\begin{minipage}{4.2cm}
\centering
\includegraphics[width=4.2cm]{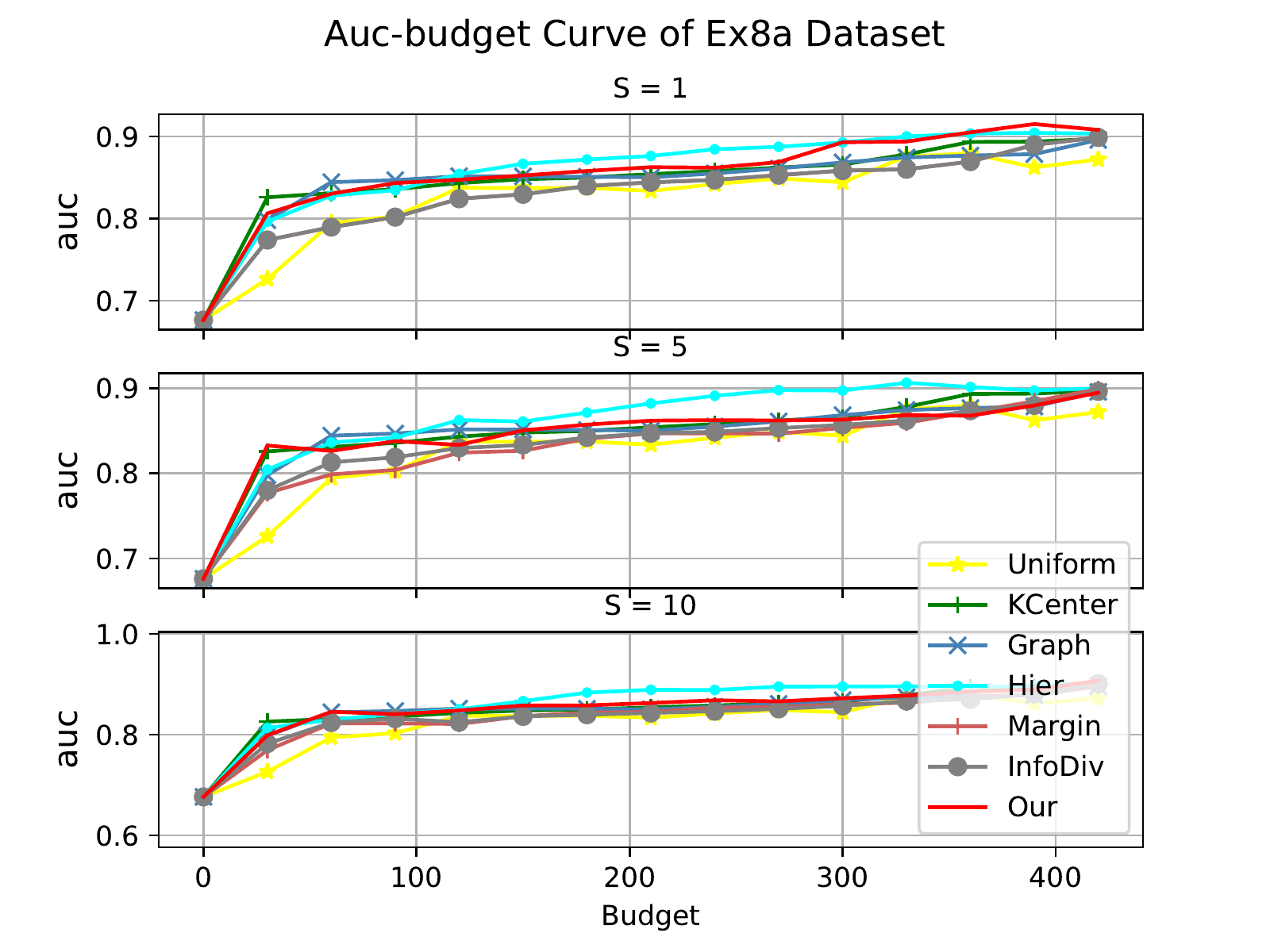}
\footnotesize(k) EX8a
\end{minipage}
\begin{minipage}{4.2cm}
\centering
\includegraphics[width=4.2cm]{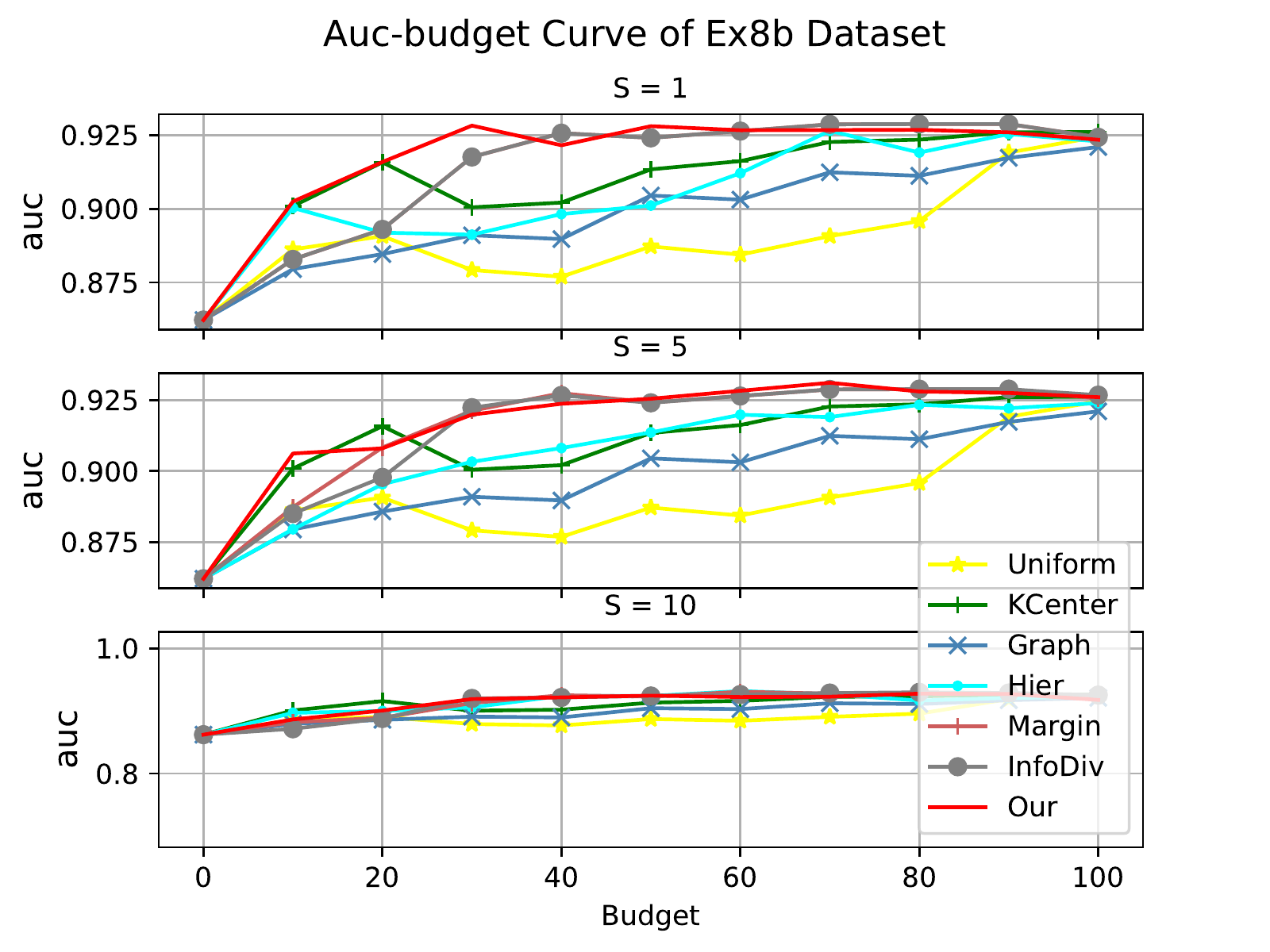}
\footnotesize(l) EX8b
\end{minipage}
\begin{minipage}{4.2cm}
\centering
\includegraphics[width=4.2cm]{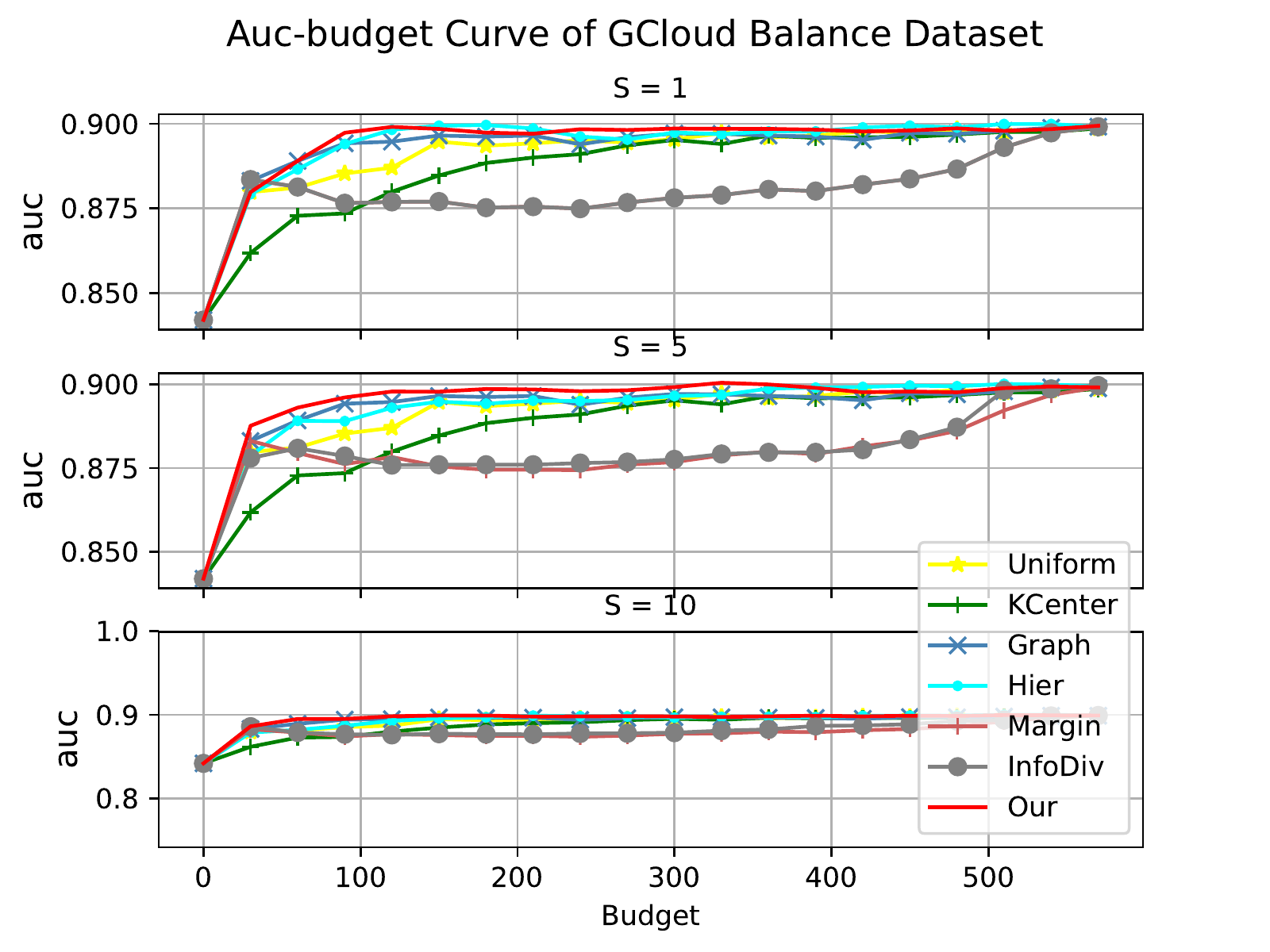}
\footnotesize(m) GCloud Balance
\end{minipage}
\begin{minipage}{4.2cm}
\centering
\includegraphics[width=4.2cm]{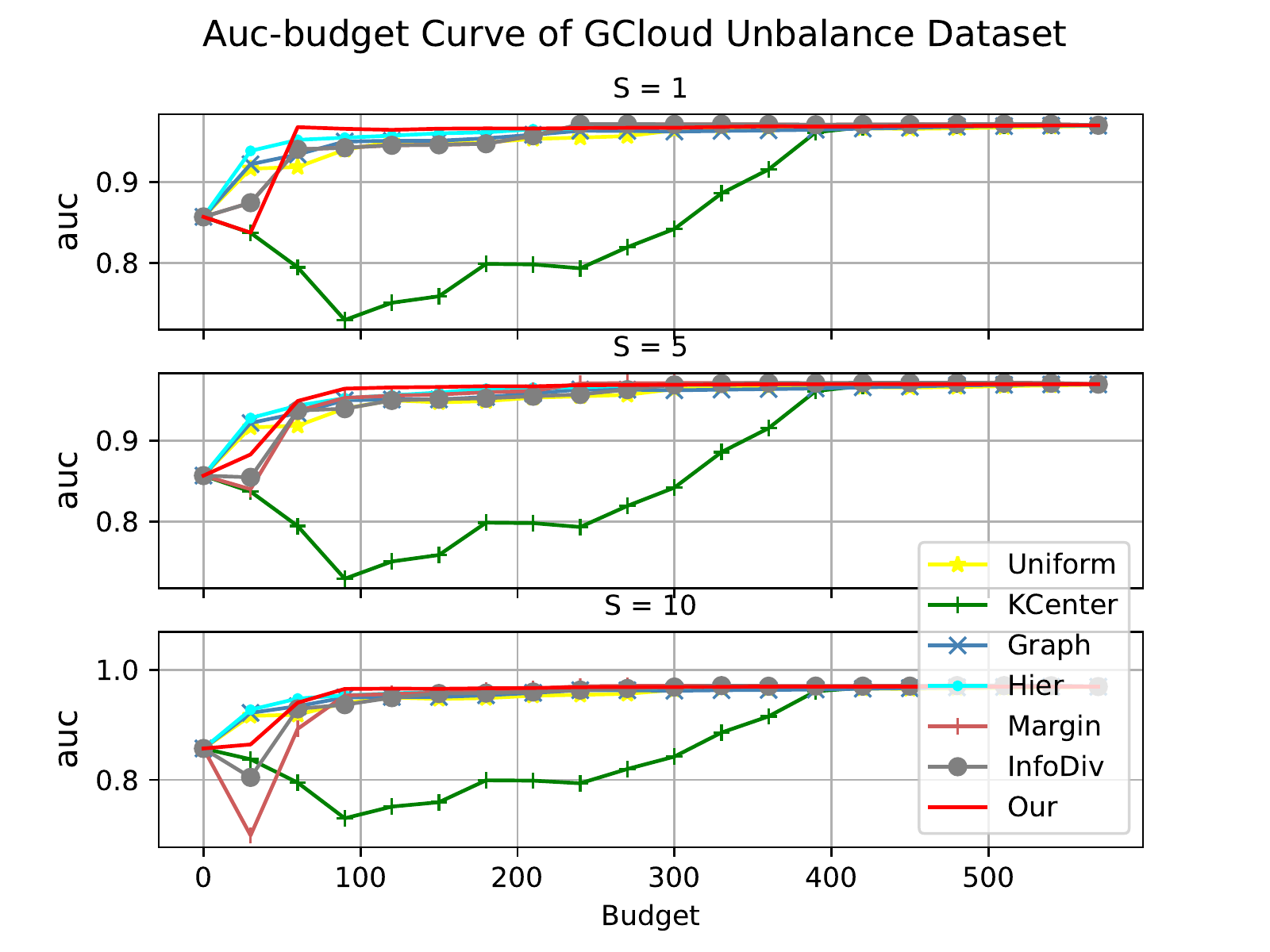}
\footnotesize(n) GCloud Unbalance
\end{minipage}
\caption{Auc vs.~budget curves on $14$ real-world and synthetic datasets ($S=\{1,5,10\}$).}
\label{auc-figure}
\end{figure*}

\begin{figure*} [tb]
\centering
\begin{minipage}{4.2cm}
\centering
\includegraphics[width=4.2cm]{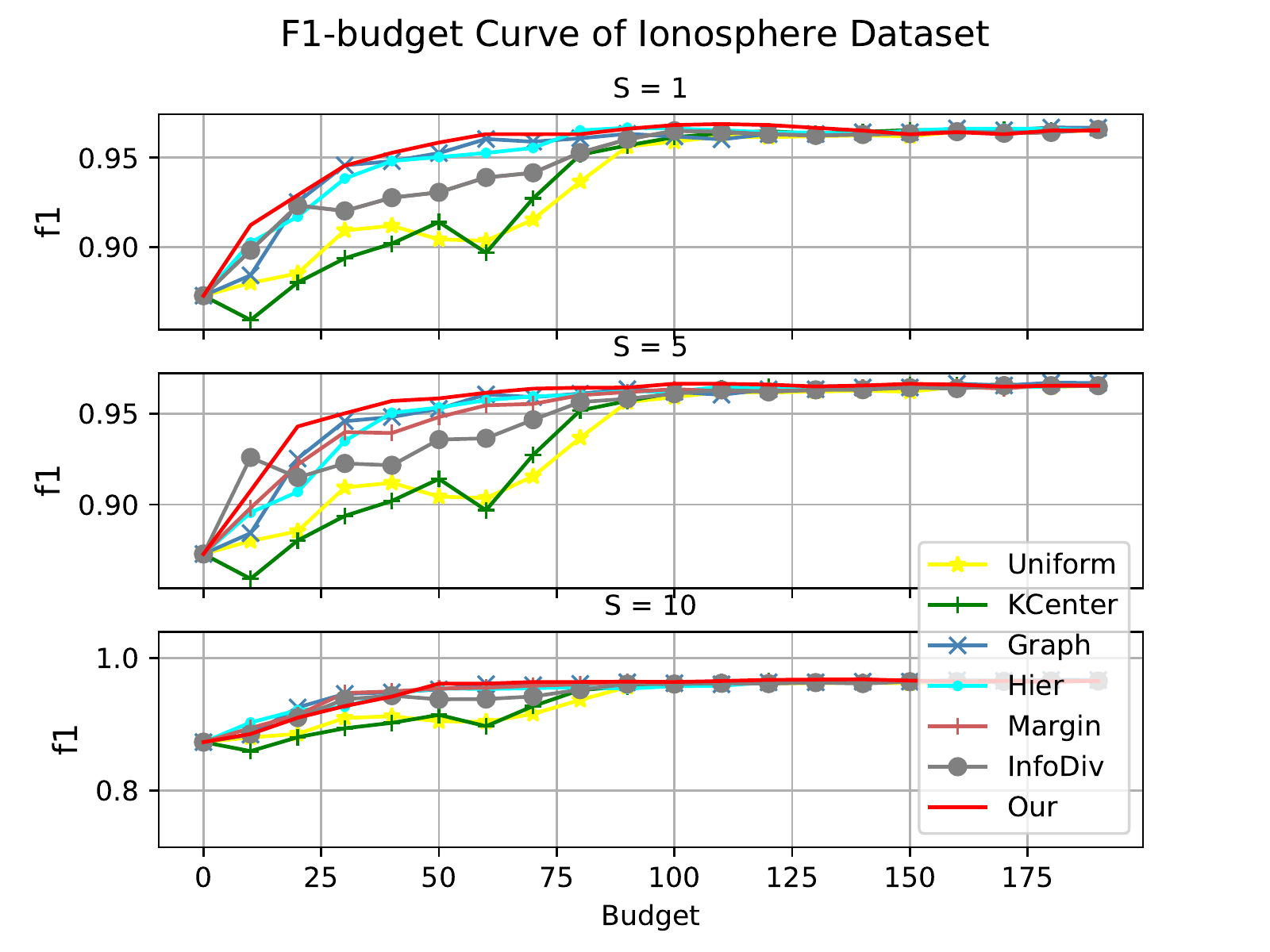}
\footnotesize(a) Ionosphere
\end{minipage}
\begin{minipage}{4.2cm}
\centering
\includegraphics[width=4.2cm]{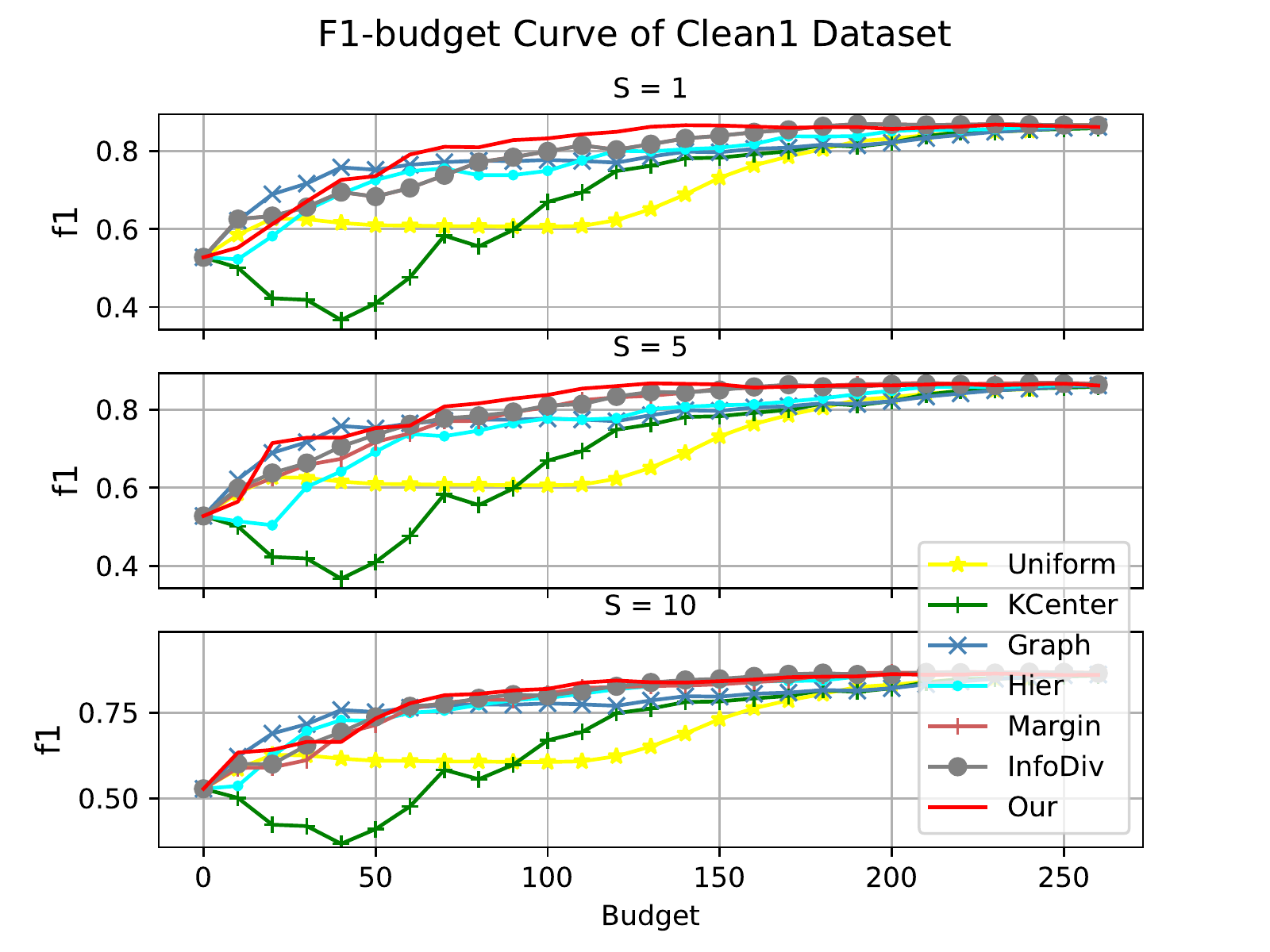}
\footnotesize(b) Clean1
\end{minipage}
\begin{minipage}{4.2cm}
\centering
\includegraphics[width=4.2cm]{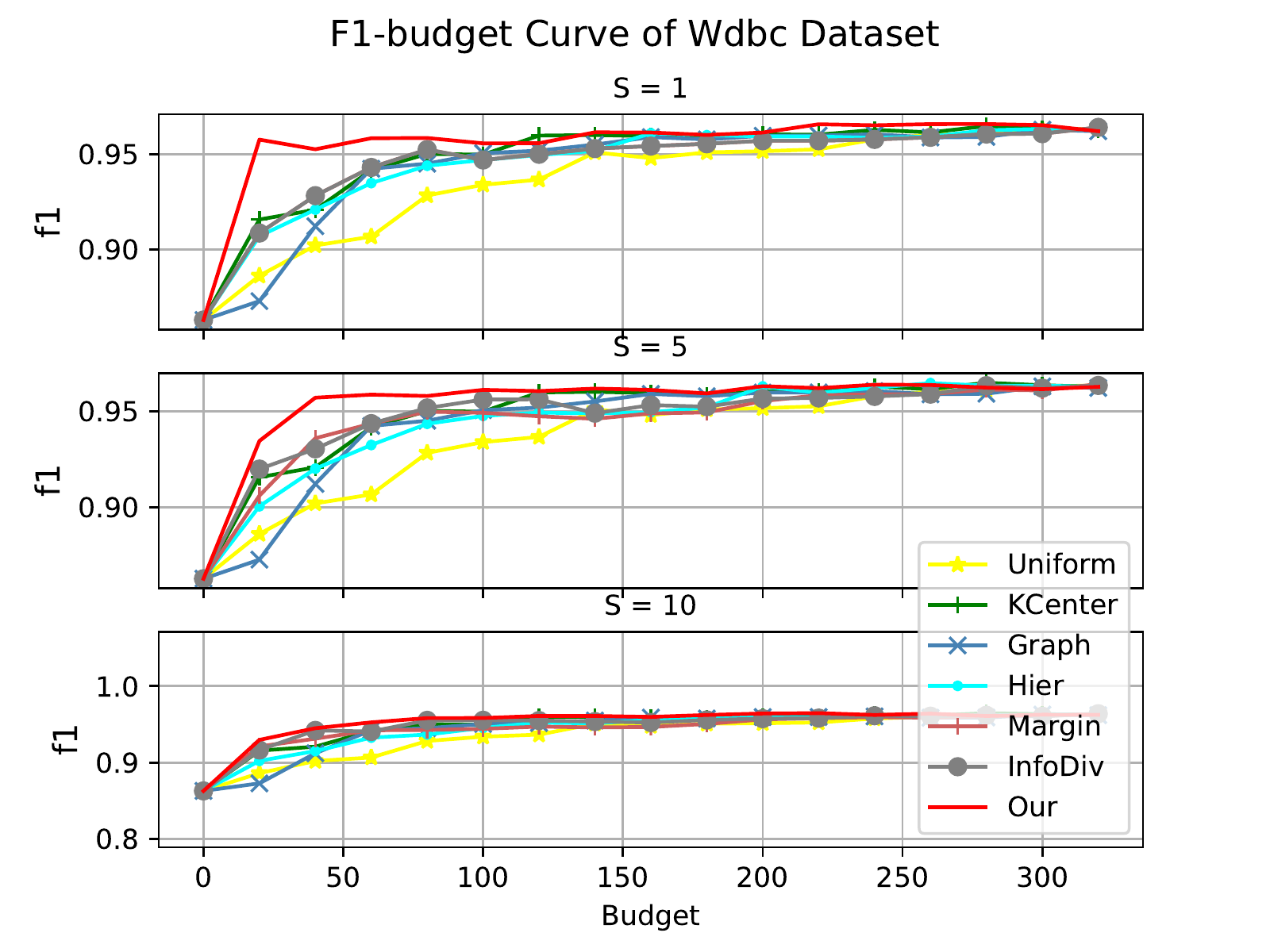}
\footnotesize(c) wdbc
\end{minipage}
\begin{minipage}{4.2cm}
\centering
\includegraphics[width=4.2cm]{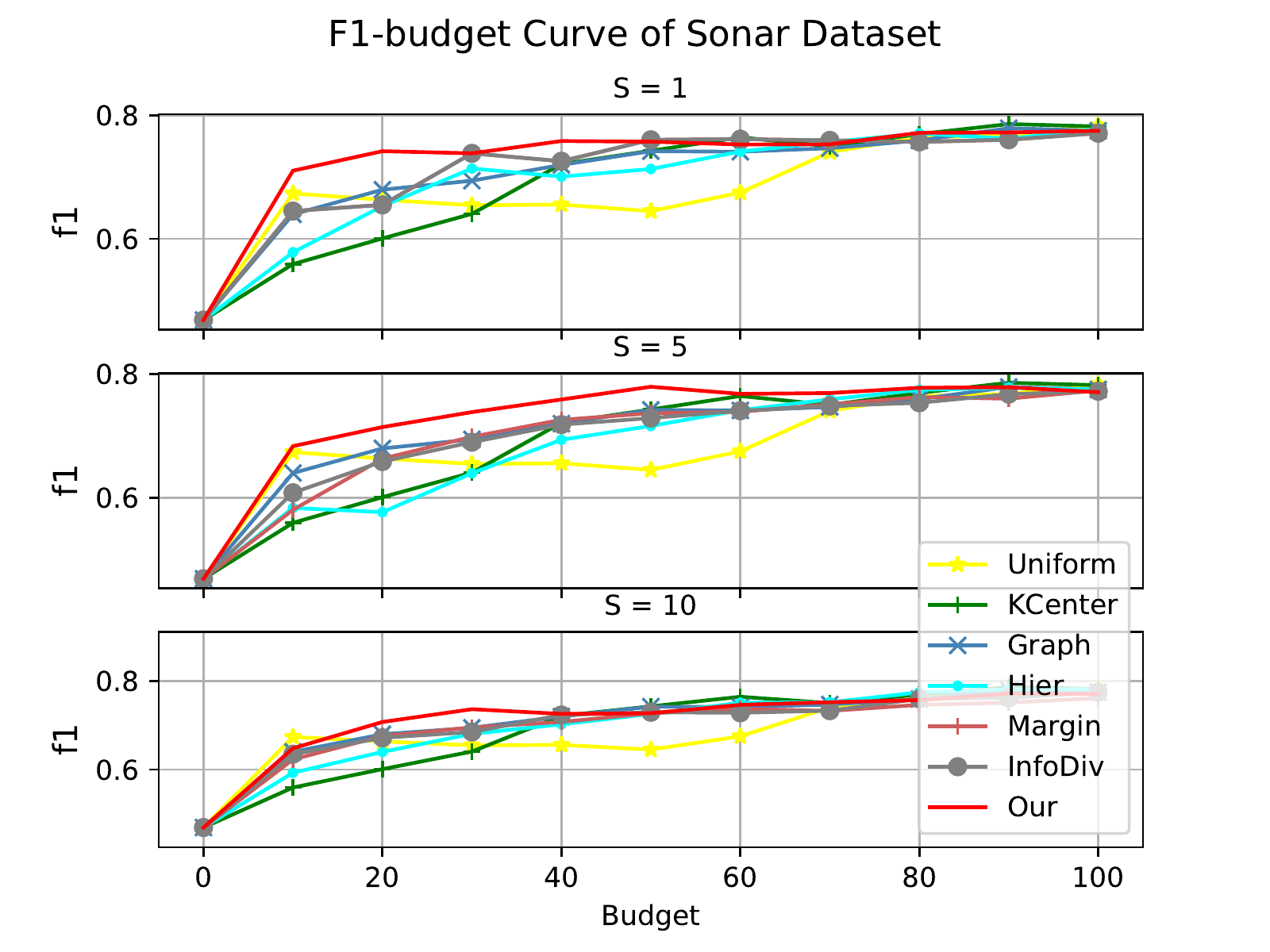}
\footnotesize(d) Sonar
\end{minipage}
\begin{minipage}{4.2cm}
\centering
\includegraphics[width=4.2cm]{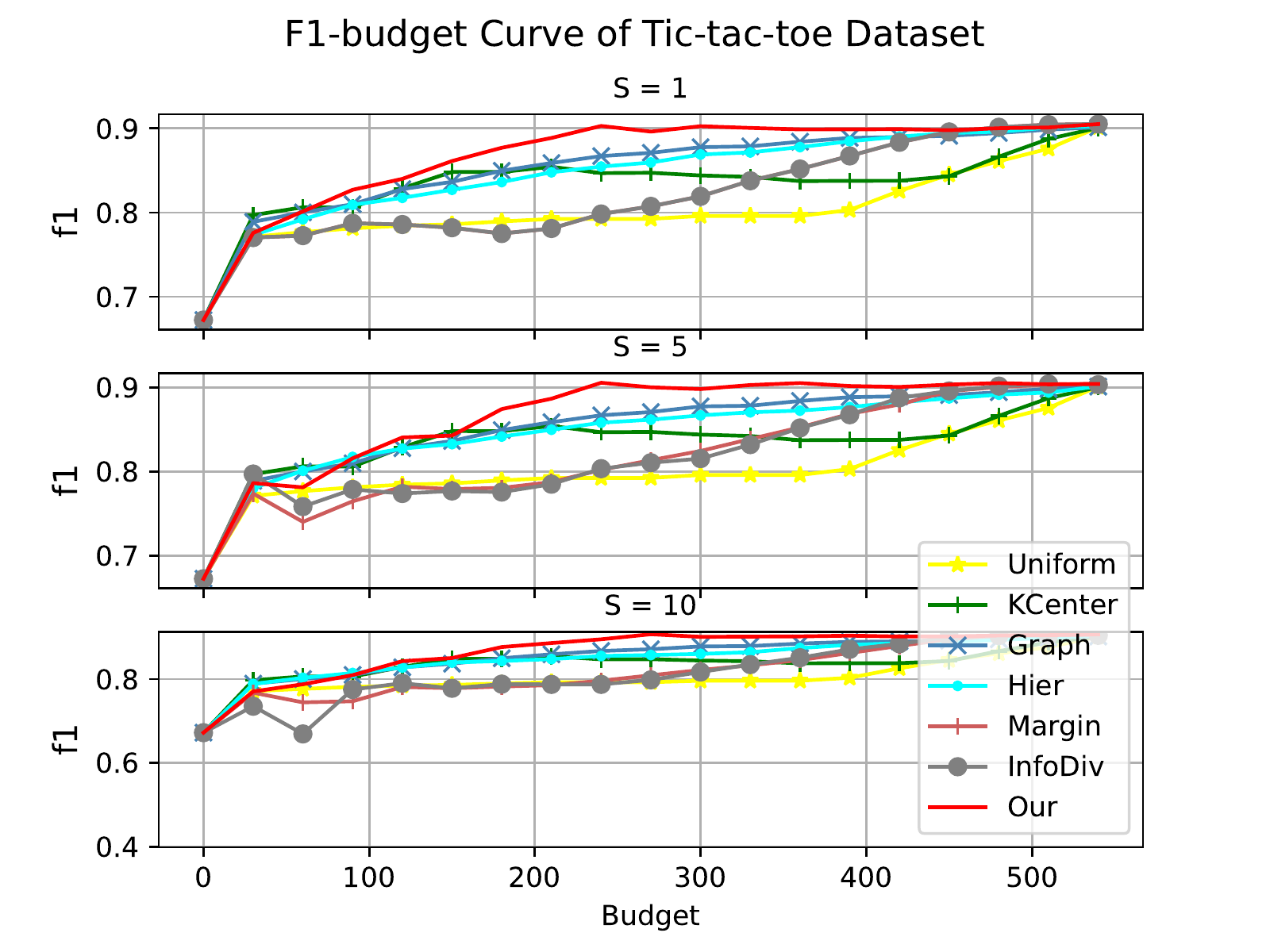}
\footnotesize(e) Tic-tac-toe
\end{minipage}
\begin{minipage}{4.2cm}
\centering
\includegraphics[width=4.2cm]{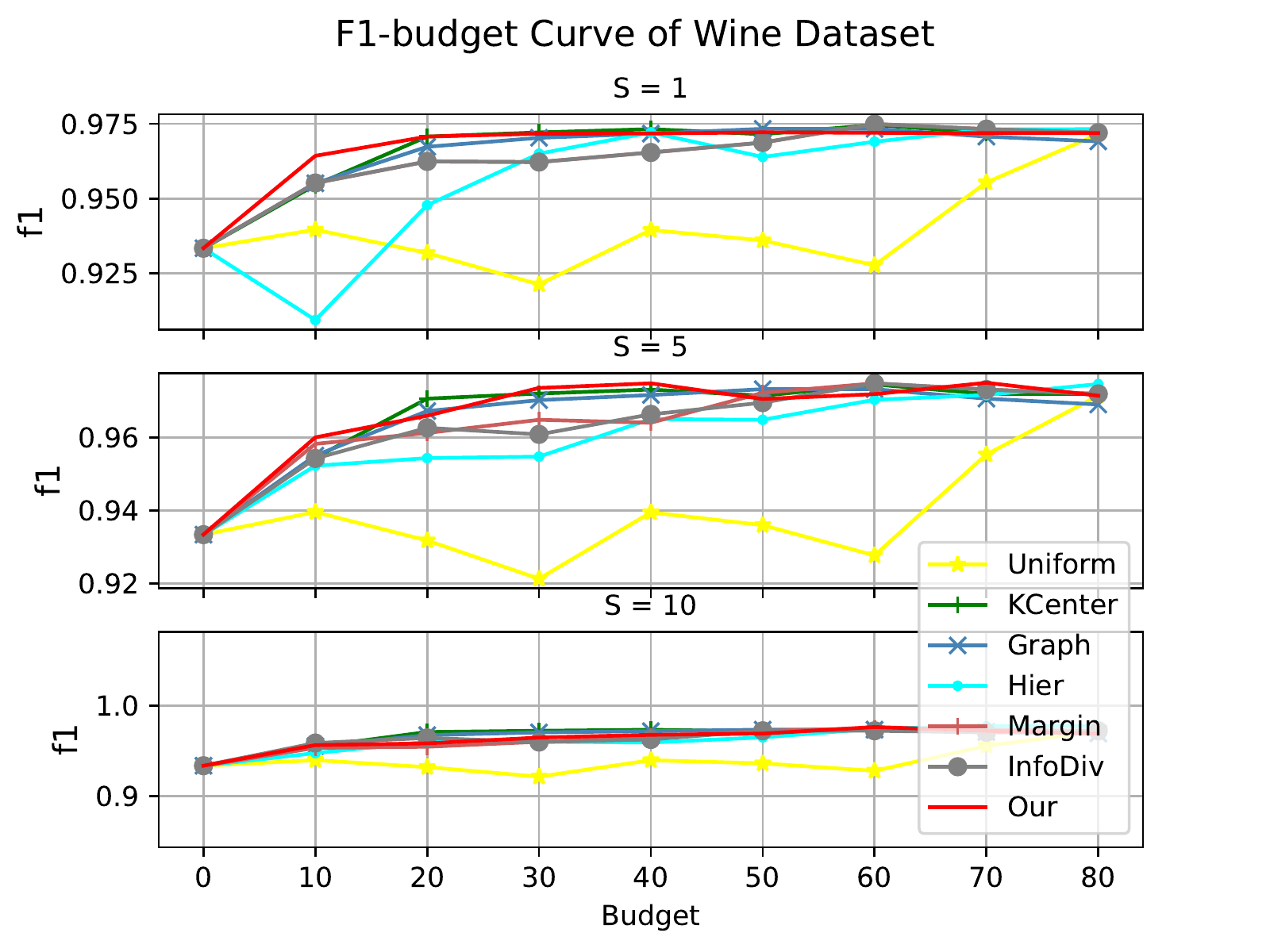}
\footnotesize(f) Wine
\end{minipage}
\begin{minipage}{4.2cm}
\centering
\includegraphics[width=4.2cm]{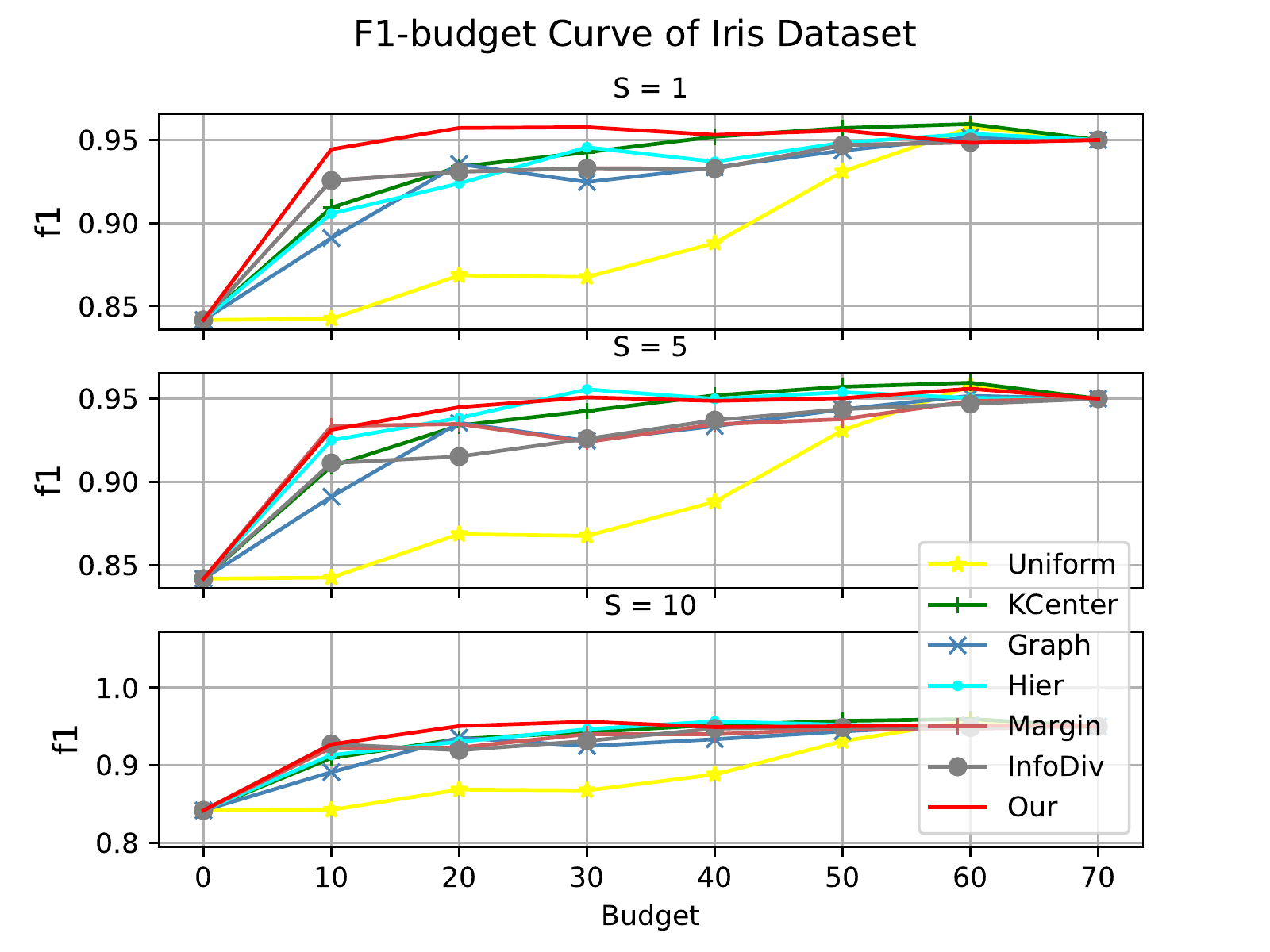}
\footnotesize(g) Iris
\end{minipage}
\begin{minipage}{4.2cm}
\centering
\includegraphics[width=4.2cm]{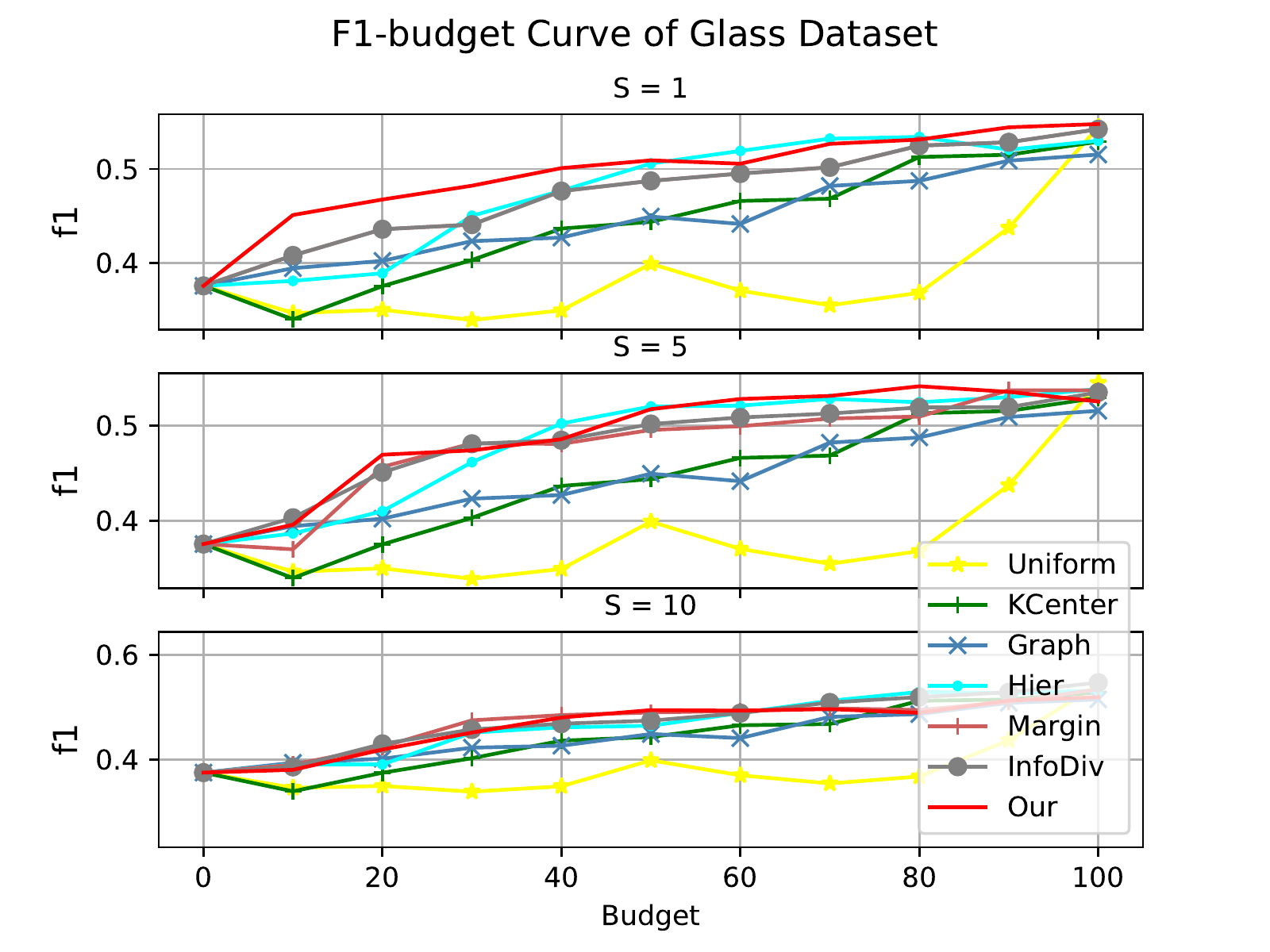}
\footnotesize(h) Glass
\end{minipage}
\begin{minipage}{4.2cm}
\centering
\includegraphics[width=4.2cm]{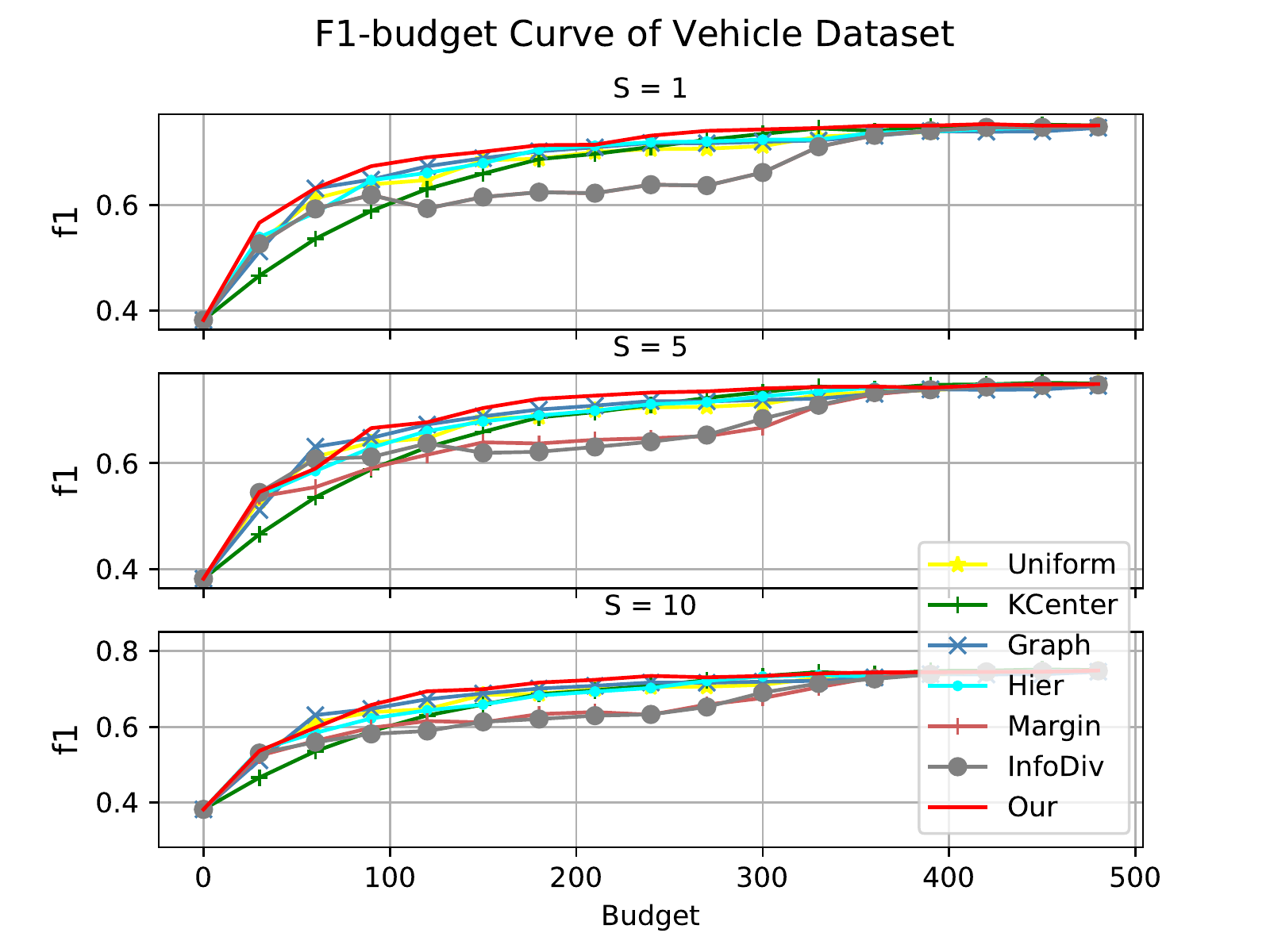}
\footnotesize(i) Vehicle
\end{minipage}
\begin{minipage}{4.2cm}
\centering
\includegraphics[width=4.2cm]{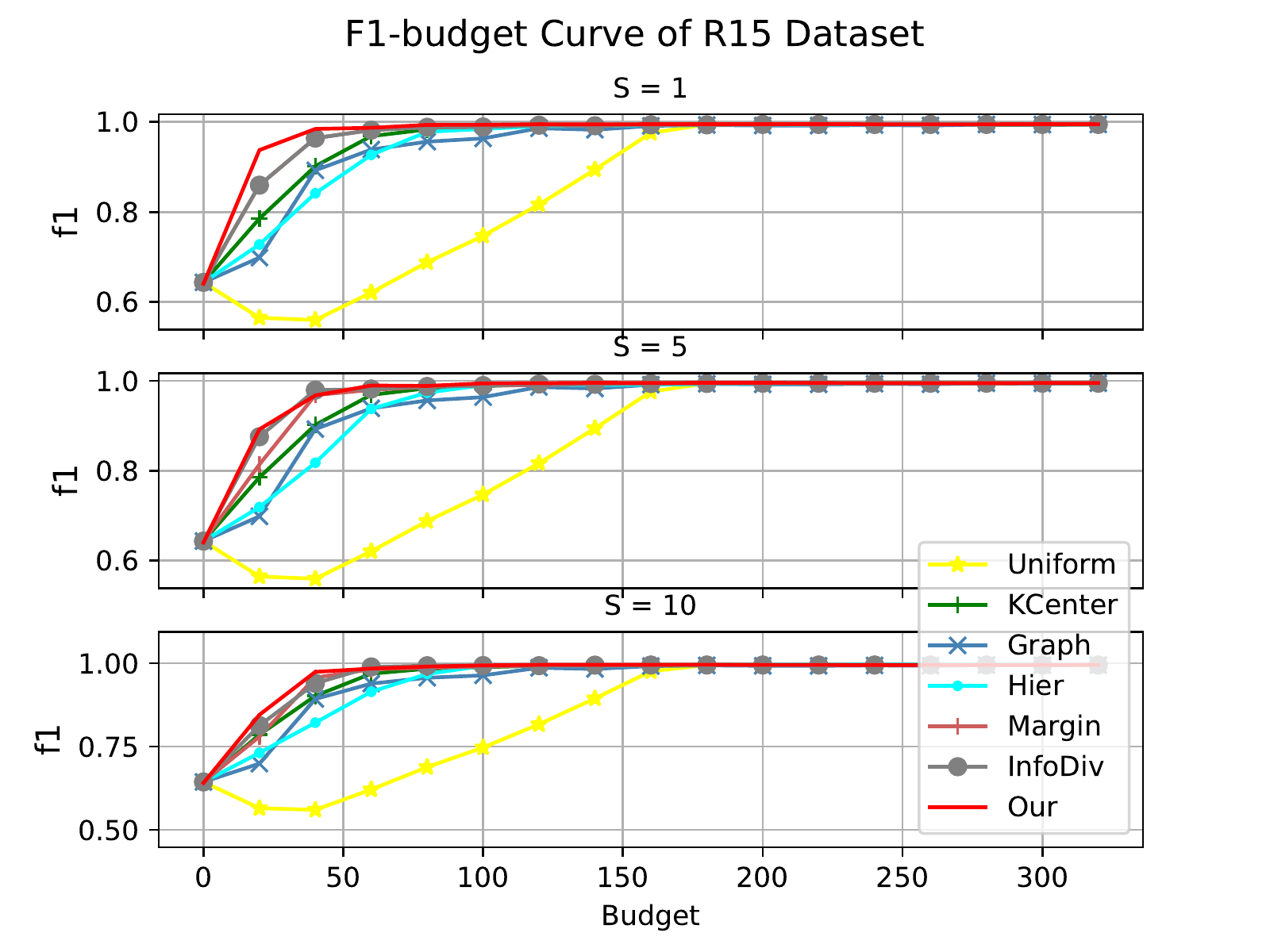}
\footnotesize(j) R15
\end{minipage}
\begin{minipage}{4.2cm}
\centering
\includegraphics[width=4.2cm]{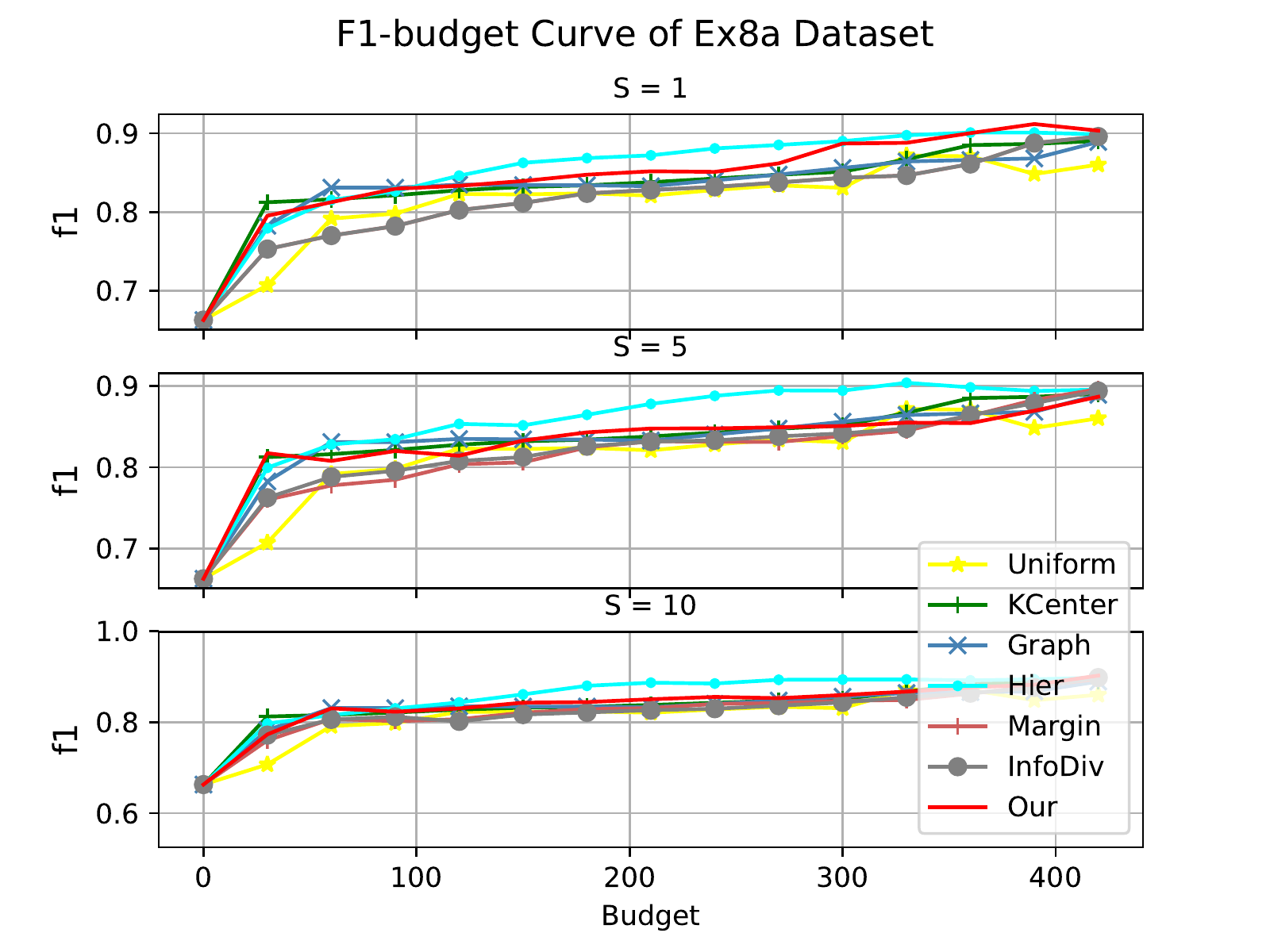}
\footnotesize(k) EX8a
\end{minipage}
\begin{minipage}{4.2cm}
\centering
\includegraphics[width=4.2cm]{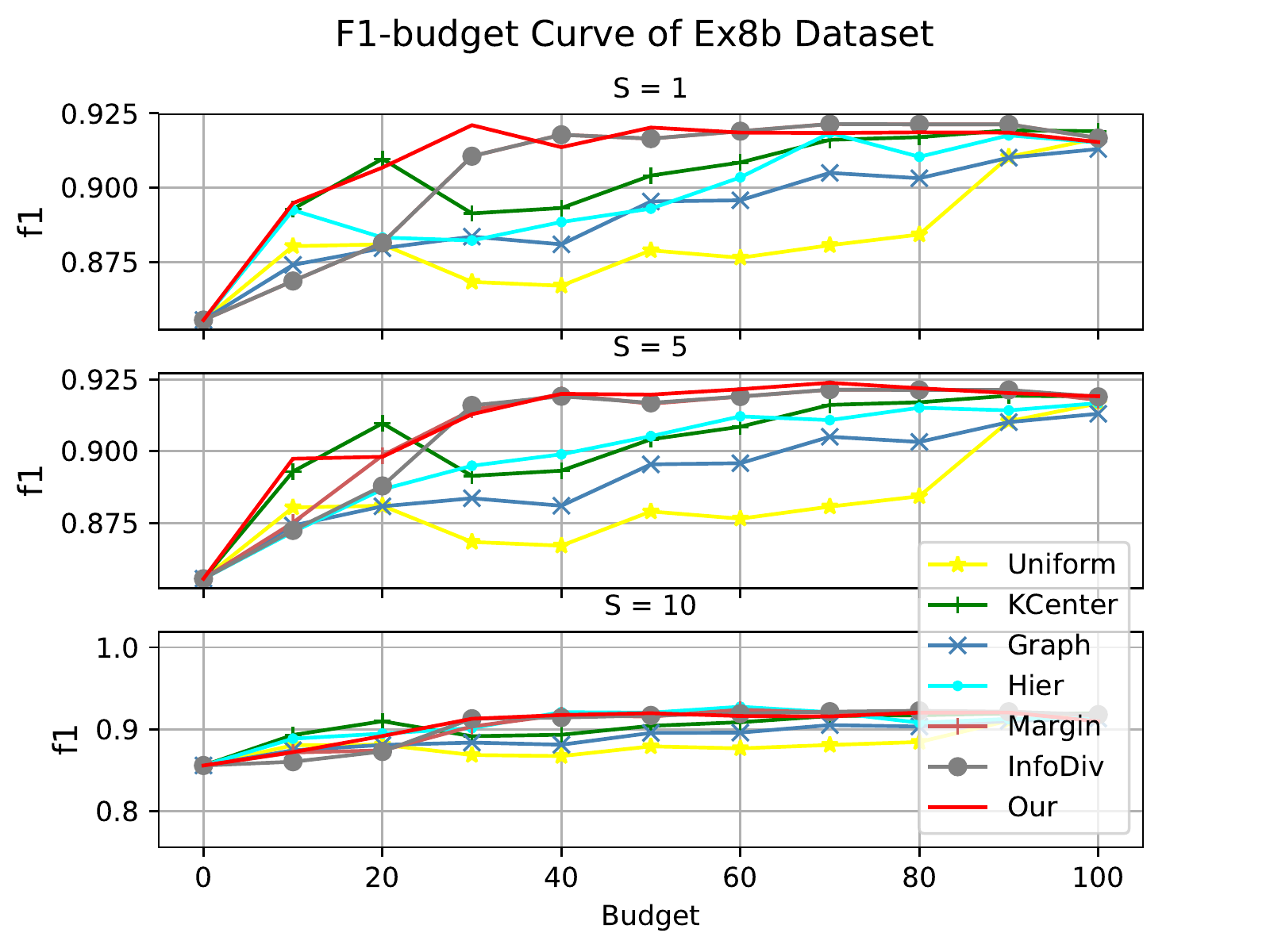}
\footnotesize(l) EX8b
\end{minipage}
\begin{minipage}{4.2cm}
\centering
\includegraphics[width=4.2cm]{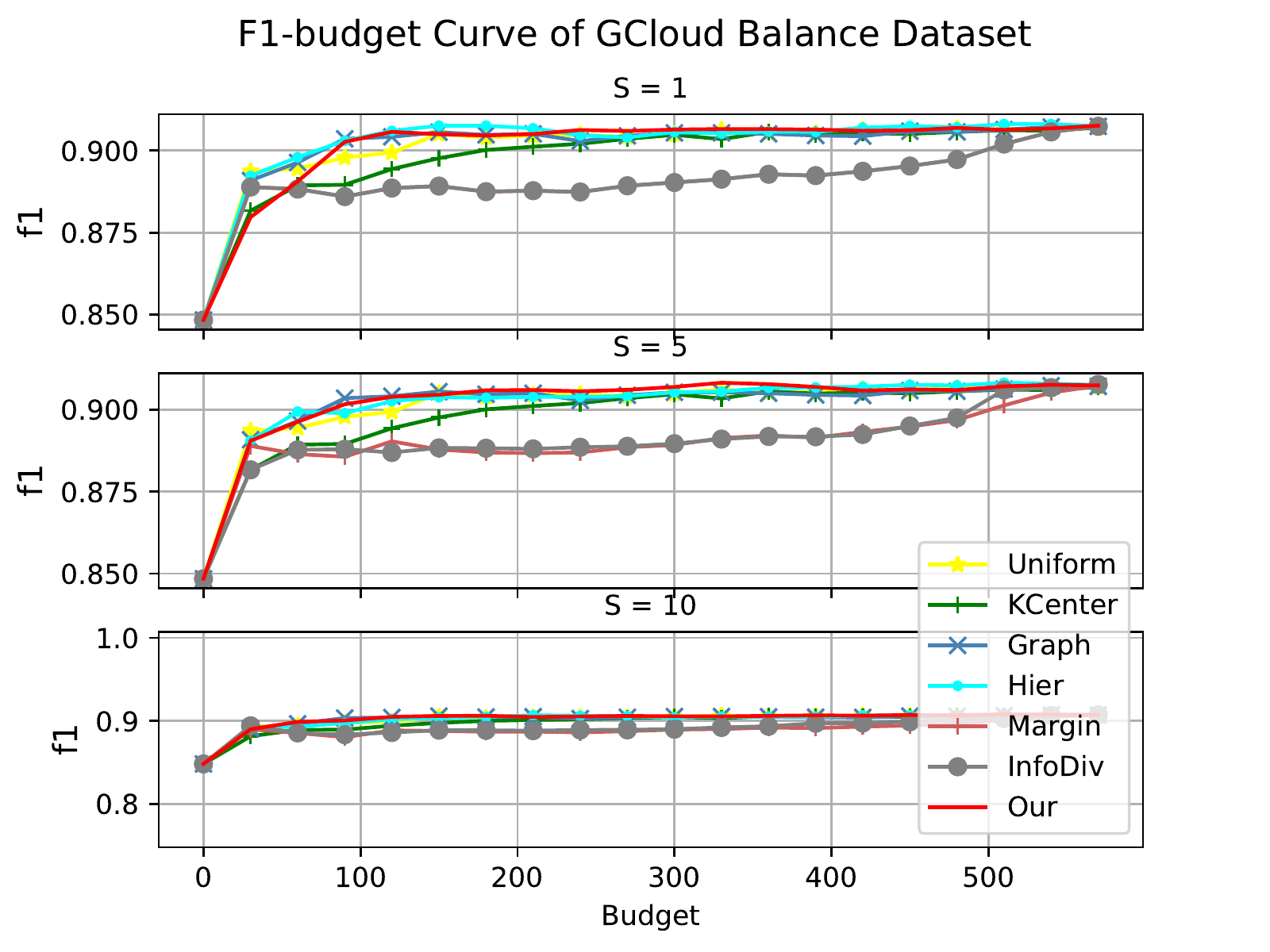}
\footnotesize(m) GCloud Balance
\end{minipage}
\begin{minipage}{4.2cm}
\centering
\includegraphics[width=4.2cm]{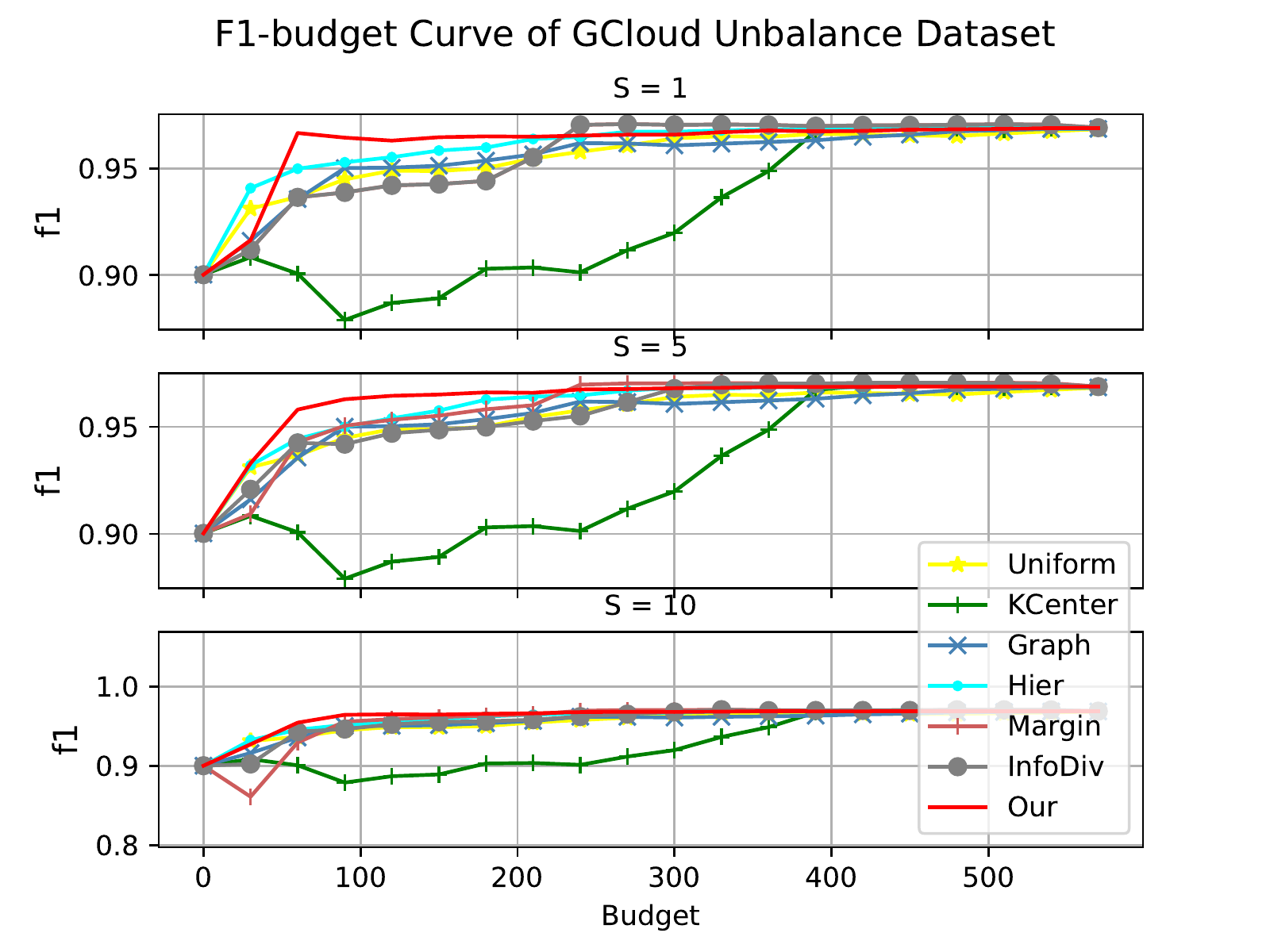}
\footnotesize(n) GCloud Unbalance
\end{minipage}
\caption{$f_1$ vs.~budget curves on $14$ real-world and synthetic datasets ($S=\{1,5,10\}$).}
\label{f1-figure}
\end{figure*}

Three typical evaluation metrics for AL classification performance include: accuracy (acc), area under the curve of ROC (auc) and F-measure ($f_1$). However, these metrics only present the classification performance under fixed budget, instead of the overall performance of active learning. Some algorithms may perform well when the budget is small, while others may perform well when the budget is large. Under these circumstances, it is difficult to judge which algorithm is superior. To evaluate the overall performance, we propose an evaluation metric based on the performance-budget curves, computed by evaluating the AL method for different fixed budgets (e.g., Accuracy vs.~Budget). The evaluation metric is called {\em area under the budget curve} (AUBC), which we denote as AUBC (acc), AUBC (auc), and AUBC ($f_1$) for the accuracy, auc, and $f_1$ budget curves, respectively. Examples of accuracy vs.~budget curve, auc vs.~budget and $f_1$ vs.~budget are shown in Figures~\ref{acc-figure},~\ref{auc-figure}, and \ref{f1-figure}. The area under the curve was calculated by the trapezoid method, and the higher values reflects better performance of the AL strategy under varying budgets.
To get an overall view of each AL algorithms performance, we rank each algorithm on each dataset and batch setting according to the AUBC metric.  We then compute the average rank over all the datasets and batch settings.

 Note that the performance of our proposed model and baselines using the AUBC evaluation metric sometimes exhibits small differences, since 1) the starting and ending points of the budget curves are fixed and identical to every algorithm; 2) most algorithms typically converge after using half of the total budget, which means that they will have similar AUBC due to considerable overlap in the remaining part of the curve.
  Therefore, in the following experimental results analysis, we use the AUBC, budget curve and paired t-test to form a comprehensive analysis.

\subsection{Overall Results on Real-world and Synthetic Data}

\begin{table}[tb]
\scriptsize
\centering
\caption{Average ranking performance across all datasets and batch settings, when using different AUBC metrics.}
\label{avgres}
{\begin{tabular}{c|ccccccc}
\hline AUBC & Uniform & KCenter & Graph & Hier & Margin & InfoDiv & Ours \\
\hline
acc & 6.21 & 4.40 & 4.00 & 3.33 & 3.83 & 3.86 & \textbf{1.21} \\
auc & 6.10 & 4.50 & 3.98 & 3.43 & 4.02 & 3.71 & \textbf{1.26} \\
$f_1$ & 6.38 & 4.38 & 3.84 & 3.66 & 3.69 & 3.63 & \textbf{1.19} \\
\hline
\end{tabular}}
\end{table}

\begin{table*}[!hbt]
\scriptsize
\centering
\caption{The mean and standard deviation performance on $9$ real-world datasets of AUBC (acc) with $S = \{1, 5, 10\}$.}
\label{realdataacc}
{\begin{tabular}{c|cccccccccc}
\toprule
 \multicolumn{1}{c}{}& & Ionosphere & Clean1 & Wdbc & Sonar & Tic-tac-toe & Iris & Wine & Glass & Vehicle \\
\toprule
\multirow{7}{*}{S = 1}
& Uniform & 0.917 (0.027) & 0.643 (0.026) & 0.953 (0.011) & 0.622 (0.051) & 0.692 (0.02) & 0.905 (0.021) & 0.944 (0.034) & 0.547 (0.034) & 0.704 (0.014) \\
& Kcenter & 0.917 (0.016) & 0.794 (0.021) & 0.965 (0.009) & 0.74 (0.036) & 0.76 (0.017) & 0.943 (0.021) & 0.969 (0.01) & 0.569 (0.045) & 0.688 (0.019) \\
& Graph & 0.934 (0.015) & 0.817 (0.017) & 0.961 (0.013) & 0.746 (0.031) & 0.793 (0.016) & 0.932 (0.028) & 0.969 (0.012) & 0.575 (0.04) & 0.705 (0.01) \\
& Hier & 0.937 (0.013) & 0.819 (0.028) & 0.963 (0.009) & 0.744 (0.034) & 0.786 (0.017) & 0.938 (0.024) & 0.956 (0.016) & 0.601 (0.045) & 0.706 (0.016) \\
& Margin & 0.929 (0.022) & 0.829 (0.017) & 0.963 (0.014) & 0.752 (0.026) & 0.754 (0.012) & 0.938 (0.034) & 0.967 (0.012) & 0.605 (0.04) & 0.657 (0.034) \\
& InfoDiv & 0.929 (0.022) & 0.829 (0.017) & 0.963 (0.014) & 0.752 (0.026) & 0.754 (0.012) & 0.938 (0.034) & 0.967 (0.012) & 0.605 (0.04) & 0.657 (0.034) \\
& Our & \textbf{0.941 (0.015)} & \textbf{0.845 (0.01)} & \textbf{0.97 (0.009)} & \textbf{0.764 (0.035)} & \textbf{0.823 (0.015)} & \textbf{0.952 (0.016)} & \textbf{0.977 (0.01)} & \textbf{0.622 (0.03)} & \textbf{0.715 (0.016)} \\
\hline

\multirow{7}{*}{S = 5}
& Uniform & 0.917 (0.026) & 0.643 (0.025) & 0.953 (0.011) & 0.622 (0.051) & 0.692 (0.02) & 0.904 (0.021) & 0.944 (0.034) & 0.546 (0.033) & 0.704 (0.014) \\
& KCenter & 0.917 (0.015) & 0.793 (0.021) & 0.965 (0.009) & 0.74 (0.037) & 0.76 (0.017) & 0.942 (0.022) & 0.969 (0.009) & 0.567 (0.044) & 0.688 (0.019) \\
& Graph & 0.934 (0.015) & 0.817 (0.017) & 0.961 (0.013) & 0.746 (0.03) & 0.793 (0.015) & 0.931 (0.028) & 0.969 (0.012) & 0.574 (0.04) & 0.705 (0.011) \\
& Hier & 0.935 (0.013) & 0.817 (0.026) & 0.962 (0.01) & 0.741 (0.032) & 0.785 (0.022) & 0.943 (0.017) & 0.958 (0.012) & 0.592 (0.034) & 0.701 (0.015) \\
& Margin & 0.935 (0.013) & 0.832 (0.015) & 0.962 (0.013) & 0.742 (0.027) & 0.757 (0.013) & 0.934 (0.038) & 0.967 (0.012) & \textbf{0.61 (0.045)} & 0.656 (0.029) \\
& InfoDiv & 0.93 (0.017) & 0.834 (0.015) & 0.964 (0.013) & 0.749 (0.03) & 0.756 (0.011) & 0.934 (0.028) & 0.967 (0.012) & 0.607 (0.033) & 0.66 (0.025) \\
& Our & \textbf{0.942 (0.012)} & \textbf{0.849 (0.013)} & \textbf{0.969 (0.009)} & \textbf{0.774 (0.035)} & \textbf{0.82 (0.012)} & \textbf{0.946 (0.022)} & \textbf{0.975 (0.01)} & 0.606 (0.034) & \textbf{0.71 (0.014)} \\
\hline

\multirow{7}{*}{S = 10}
& Uniform & 0.917 (0.027) & 0.643 (0.026) & 0.953 (0.011) & 0.623 (0.051) & 0.692 (0.02) & 0.904 (0.023) & 0.943 (0.034) & 0.545 (0.032) & 0.704 (0.014) \\
& KCenter & 0.916 (0.016) & 0.793 (0.021) & 0.965 (0.009) & 0.739 (0.035) & 0.76 (0.017) & 0.942 (0.022) & 0.969 (0.009) & 0.567 (0.043) & 0.689 (0.019) \\
& Graph & 0.935 (0.015) & 0.818 (0.016) & 0.961 (0.013) & 0.746 (0.031) & 0.793 (0.016) & 0.931 (0.029) & 0.968 (0.012) & 0.574 (0.039) & 0.705 (0.011) \\
& Hier & 0.933 (0.017) & 0.831 (0.015) & 0.962 (0.01) & 0.744 (0.03) & 0.784 (0.019) & 0.941 (0.019) & 0.956 (0.015) & 0.578 (0.047) & 0.7 (0.016) \\
& Margin & 0.936 (0.015) & 0.829 (0.022) & 0.961 (0.015) & 0.741 (0.032) & 0.752 (0.016) & 0.936 (0.029) & 0.965 (0.013) & \textbf{0.606 (0.037)} & 0.662 (0.03) \\
& InfoDiv & 0.93 (0.015) & 0.834 (0.017) & 0.964 (0.013) & 0.74 (0.034) & 0.751 (0.01) & 0.936 (0.028) & 0.966 (0.013) & 0.605 (0.037) & 0.653 (0.03) \\
& Our & \textbf{0.937 (0.014)} & \textbf{0.842 (0.01)} & \textbf{0.968 (0.009)} & \textbf{0.759 (0.034)} & \textbf{0.821 (0.011)} & \textbf{0.945 (0.023)} & \textbf{0.976 (0.011)} & 0.586 (0.05) & \textbf{0.708 (0.017)} \\
\hline

\end{tabular}}
\end{table*}

\begin{table*}[!hbt]
\scriptsize
\centering
\caption{The mean and standard deviation performance on $9$ real-world datasets of AUBC (auc) with $S = \{1, 5, 10\}$.}
\label{realdataauc}
{\begin{tabular}{c|cccccccccc}
\toprule
 \multicolumn{1}{c}{}& & Ionosphere & Clean1 & Wdbc & Sonar & Tic-tac-toe & Iris & Wine & Glass & Vehicle \\
\toprule
\multirow{7}{*}{S = 1}
& Uniform & 0.91 (0.023) & 0.679 (0.014) & 0.955 (0.01) & 0.647 (0.034) & 0.556 (0.008) & 0.929 (0.014) & 0.959 (0.024) & 0.65 (0.037) & 0.802 (0.006) \\
& KCenter & 0.906 (0.017) & 0.771 (0.019) & 0.959 (0.011) & 0.746 (0.034) & 0.665 (0.018) & 0.957 (0.017) & 0.978 (0.01) & 0.678 (0.022) & 0.792 (0.01) \\
& Graph & 0.912 (0.019) & 0.811 (0.017) & 0.954 (0.016) & 0.752 (0.03) & 0.724 (0.02) & 0.949 (0.022) & \textbf{0.978 (0.009)} & 0.679 (0.03) & 0.803 (0.005) \\
& Hier & 0.921 (0.017) & 0.811 (0.029) & 0.956 (0.013) & 0.748 (0.034) & 0.715 (0.025) & 0.954 (0.018) & 0.971 (0.011) & 0.695 (0.031) & 0.804 (0.009) \\
& Margin & 0.915 (0.024) & 0.823 (0.019) & 0.957 (0.017) & 0.756 (0.023) & 0.686 (0.024) & 0.953 (0.024) & 0.976 (0.01) & 0.694 (0.034) & 0.771 (0.022) \\
& InfoDiv & 0.915 (0.024) & 0.823(0.019) & 0.957 (0.017) & 0.756 (0.023) & 0.686 (0.024) & 0.953 (0.024) & 0.976 (0.01) & 0.694 (0.034) & 0.771 (0.022) \\
& Our & \textbf{0.925 (0.018)} & \textbf{0.84 (0.01)} & \textbf{0.965 (0.01)} & \textbf{0.769 (0.032)} & \textbf{0.773 (0.02)} & \textbf{0.963 (0.013)} & \textbf{0.978 (0.009)} & \textbf{0.708 (0.028)} & \textbf{0.81 (0.01)} \\
\hline

\multirow{7}{*}{S = 5}
& Uniform & 0.91 (0.023) & 0.679 (0.013) & 0.955 (0.01) & 0.647 (0.035) & 0.555 (0.008) & 0.928 (0.013) & 0.959 (0.023) & 0.65 (0.038) & 0.802 (0.006) \\
& KCenter & 0.906 (0.017) & 0.771 (0.019) & 0.959 (0.011) & 0.745 (0.035) & 0.665 (0.018) & 0.956 (0.018) & 0.978 (0.009) & 0.678 (0.021) & 0.792 (0.01) \\
& Graph & 0.913 (0.019) & 0.811 (0.017) & 0.954 (0.017) & 0.751 (0.029) & 0.723 (0.02) & 0.949 (0.021) & 0.978 (0.009) & 0.679 (0.03) & 0.803 (0.006) \\
& Hier & 0.916 (0.018) & 0.808 (0.025) & 0.955 (0.013) & 0.743 (0.033) & 0.712 (0.03) & 0.956 (0.015) & 0.972 (0.008) & 0.696 (0.02) & 0.8 (0.007) \\
& Margin & 0.919 (0.017) & 0.828 (0.013) & 0.957 (0.016) & 0.745 (0.021) & 0.695 (0.025) & 0.951 (0.026) & 0.976 (0.01) & 0.695 (0.032) & 0.771 (0.019) \\
& InfoDiv & 0.917 (0.017) & 0.83 (0.014) & 0.959 (0.015) & 0.751 (0.022) & 0.696 (0.016) & 0.951 (0.021) & 0.976 (0.01) & 0.696 (0.038) & 0.773 (0.018) \\
& Our & \textbf{0.927 (0.016)} & \textbf{0.843 (0.012)} & \textbf{0.965 (0.011)} & \textbf{0.776 (0.034)} & \textbf{0.772 (0.017)} & \textbf{0.958 (0.018)} & \textbf{0.978 (0.008)} & \textbf{0.702 (0.027)} & \textbf{0.806 (0.008)} \\
\hline

\multirow{7}{*}{S = 10}
& Uniform & 0.91 (0.023) & 0.679 (0.013) & 0.955 (0.01) & 0.648 (0.034) & 0.556 (0.008) & 0.928 (0.014) & 0.958 (0.024) & 0.649 (0.038) & 0.802 (0.006) \\
& KCenter & 0.906 (0.017) & 0.77 (0.018) & 0.958 (0.012) & 0.744 (0.033) & 0.665 (0.018) & 0.956 (0.017) & \textbf{0.978 (0.007)} & 0.677 (0.022) & 0.793 (0.01) \\
& Graph & 0.913 (0.02) & 0.812 (0.017) & 0.954 (0.016) & 0.751 (0.03) & 0.723 (0.021) & 0.948 (0.022) & 0.978 (0.009) & 0.679 (0.03) & 0.803 (0.006) \\
& Hier & 0.916 (0.02) & 0.824 (0.014) & 0.955 (0.013) & 0.745 (0.029) & 0.705 (0.024) & 0.955 (0.015) & 0.971 (0.01) & 0.687 (0.027) & 0.8 (0.011) \\
& Margin & 0.921 (0.019) & 0.825 (0.021) & 0.955 (0.019) & 0.743 (0.028) & 0.688 (0.034) & 0.952 (0.022) & 0.974 (0.01) & \textbf{0.694 (0.039)} & 0.774 (0.02) \\
& InfoDiv & 0.916 (0.02) & 0.829 (0.017) & 0.959 (0.015) & 0.744 (0.027) & 0.691 (0.023) & 0.953 (0.021) & 0.975 (0.01) & 0.69 (0.025) & 0.769 (0.021) \\
& Our & \textbf{0.921 (0.016)} & \textbf{0.835 (0.01)} & \textbf{0.963 (0.01)} & \textbf{0.762 (0.036)} & \textbf{0.774 (0.015)} & \textbf{0.958 (0.018)} & 0.976 (0.01) & 0.691 (0.032) & \textbf{0.805 (0.009)} \\

\hline

\end{tabular}}
\end{table*}

\begin{table*}[!hbt]
\scriptsize
\centering
\caption{The mean and standard deviation performance on $9$ real-world datasets of AUBC ($f_1$) with $S = \{1, 5, 10\}$.}
\label{realdataf1}
{\begin{tabular}{c|cccccccccc}
\toprule
 \multicolumn{1}{c}{}& & Ionosphere & Clean1 & Wdbc & Sonar & Tic-tac-toe & Iris & Wine & Glass & Vehicle \\
\toprule
\multirow{7}{*}{S = 1}
& Uniform & 0.936 (0.023) & 0.712 (0.025) & 0.938 (0.016) & 0.693 (0.033) & 0.808 (0.017) & 0.894 (0.024) & 0.942 (0.036) & 0.393 (0.062) & 0.684 (0.011) \\
& KCenter & 0.936 (0.013) & 0.687 (0.041) & 0.952 (0.013) & 0.703 (0.06) & 0.841 (0.014) & 0.937 (0.027) & 0.968 (0.01) & 0.449 (0.034) & 0.67 (0.018) \\
& Graph & 0.952 (0.011) & 0.783 (0.021) & 0.945 (0.019) & 0.719 (0.054) & 0.858 (0.011) & 0.926 (0.031) & 0.968 (0.013) & 0.451 (0.053) & 0.686 (0.009) \\
& Hier & 0.953 (0.01) & 0.774 (0.044) & 0.948 (0.013) & 0.705 (0.059) & 0.852 (0.013) & 0.933 (0.026) & 0.953 (0.019) & 0.481 (0.054) & 0.687 (0.015) \\
& Margin & 0.947 (0.018) & 0.793 (0.031) & 0.948 (0.019) & 0.726 (0.031) & 0.826 (0.008) & 0.932 (0.036) & 0.966 (0.012) & 0.481 (0.06) & 0.65 (0.037) \\
& InfoDiv & 0.947 (0.018) & 0.793 (0.031) & 0.948 (0.019) & 0.726 (0.031) & 0.826 (0.008) & 0.932 (0.036) & 0.966 (0.012) & 0.481 (0.06) & 0.65 (0.037) \\
& Our & \textbf{0.956 (0.012)} & \textbf{0.809 (0.013)} & \textbf{0.959 (0.013)} & \textbf{0.742 (0.042)} & \textbf{0.873 (0.012)} & \textbf{0.947 (0.021)} & \textbf{0.969 (0.009)} & \textbf{0.503 (0.045)} & \textbf{0.704 (0.017)} \\
\hline

\multirow{7}{*}{S = 5}
& Uniform & 0.936 (0.022) & 0.711 (0.025) & 0.938 (0.016) & 0.692 (0.033) & 0.807 (0.017) & 0.893 (0.024) & 0.942 (0.035) & 0.392 (0.062) & 0.683 (0.011) \\
& KCenter & 0.937 (0.012) & 0.687 (0.041) & 0.952 (0.013) & 0.702 (0.061) & 0.841 (0.014) & 0.936 (0.028) & 0.968 (0.009) & 0.448 (0.032) & 0.671 (0.018) \\
& Graph & 0.952 (0.011) & 0.783 (0.02) & 0.945 (0.019) & 0.718 (0.053) & 0.858 (0.011) & 0.926 (0.031) & 0.967 (0.012) & 0.451 (0.052) & 0.686 (0.009) \\
& Hier & 0.952 (0.01) & 0.765 (0.05) & 0.947 (0.015) & 0.692 (0.058) & 0.853 (0.014) & 0.937 (0.022) & 0.955 (0.013) & 0.486 (0.041) & 0.683 (0.012) \\
& Margin & 0.951 (0.011) & 0.798 (0.018) & 0.948 (0.017) & 0.713 (0.026) & 0.825 (0.011) & 0.928 (0.038) & 0.966 (0.012) & 0.485 (0.056) & 0.651 (0.033) \\
& InfoDiv & 0.948 (0.014) & 0.801 (0.019) & 0.95 (0.017) & 0.713 (0.036) & 0.825 (0.011) & 0.929 (0.031) & 0.966 (0.012) & 0.485 (0.064) & 0.657 (0.029) \\
& Our & \textbf{0.957 (0.009)} & \textbf{0.815 (0.013)} & \textbf{0.958 (0.014}) & \textbf{0.745 (0.051)} & \textbf{0.87 (0.01)} & \textbf{0.94 (0.027)} & \textbf{0.968 (0.008)} & \textbf{0.494 (0.04)} & \textbf{0.697 (0.013)} \\
\hline

\multirow{7}{*}{S = 10}
& Uniform & 0.936 (0.022) & 0.71 (0.025) & 0.938 (0.016) & 0.691 (0.032) & 0.807 (0.018) & 0.893 (0.025) & 0.941 (0.035) & 0.391 (0.064) & 0.683 (0.011) \\
& KCenter & 0.936 (0.013) & 0.687 (0.039) & 0.951 (0.013) & 0.7 (0.059) & 0.84 (0.013) & 0.936 (0.027) & \textbf{0.968 (0.009)} & 0.449 (0.032) & 0.672 (0.019) \\
& Graph & 0.952 (0.011) & 0.784 (0.019) & 0.945 (0.019) & 0.715 (0.055) & 0.858 (0.012) & 0.925 (0.032) & 0.967 (0.013) & 0.453 (0.053) & 0.687 (0.009) \\
& Hier & 0.95 (0.012) & 0.792 (0.023) & 0.946 (0.014) & 0.706 (0.045) & 0.853 (0.014) & 0.935 (0.023) & 0.953 (0.016) & 0.473 (0.048) & 0.68 (0.022) \\
& Margin & 0.952 (0.011) & 0.793 (0.03) & 0.946 (0.021) & 0.704 (0.038) & 0.822 (0.013) & 0.931 (0.034) & 0.964 (0.012) & \textbf{0.477 (0.072)} & 0.652 (0.036) \\
& InfoDiv & 0.947 (0.012) & 0.8 (0.023) & 0.95 (0.017) & 0.708 (0.037) & 0.82 (0.012) & 0.931 (0.03) & 0.965 (0.012) & 0.478 (0.043) & 0.649 (0.032) \\
& Our & \textbf{0.952 (0.01)} & \textbf{0.803 (0.017)} & \textbf{0.956 (0.013)} & \textbf{0.722 (0.056)} & \textbf{0.87 (0.009)} & \textbf{0.94 (0.026)} & 0.965 (0.014) & 0.472 (0.047) & \textbf{0.696 (0.017)} \\
\hline

\end{tabular}}
\end{table*}

\begin{table*}[!hbt]
\scriptsize
\centering
\caption{The mean and standard deviation performance on $5$ synthetic datasets of AUBC (acc) with $S = \{1, 5, 10\}$.}
\label{syncdataacc}
{\begin{tabular}{c|ccccccccc}
\toprule
 \multicolumn{1}{c}{}& & R15 & EX8a & EX8b & GCloud Balance & GCloud Unbalance \\
\toprule
\multirow{7}{*}{S = 1}
& Uniform & 0.881 (0.013) & 0.834 (0.017) & 0.89 (0.035) & 0.894 (0.011) & 0.943 (0.006) \\
& KCenter & 0.968 (0.006) & 0.858 (0.013) & 0.908 (0.027) & 0.89 (0.011) & 0.897 (0.027) \\
& Graph & 0.956 (0.015) & 0.855 (0.014) & 0.897 (0.027) & 0.895 (0.012) & 0.943 (0.006) \\
& Hier & 0.96 (0.006) & \textbf{0.867 (0.015)} & 0.904 (0.025) & 0.896 (0.012) & 0.95 (0.007) \\
& Margin & 0.976 (0.003) & 0.84 (0.013) & 0.915 (0.023) & 0.883 (0.016) & 0.943 (0.005) \\
& InfoDiv & 0.976 (0.003) & 0.84 (0.013) & 0.915 (0.023) & 0.883 (0.016) & 0.943 (0.005) \\
& Our & \textbf{0.986 (0.003)} & 0.863 (0.011) & \textbf{0.917 (0.02)} & \textbf{0.897 (0.014)} & \textbf{0.95 (0.006)} \\
\hline

\multirow{7}{*}{S = 5}
& Uniform & 0.881 (0.013) & 0.833 (0.017) & 0.89 (0.035) & 0.894 (0.011) & 0.943 (0.006) \\
& KCenter & 0.968 (0.006) & 0.858 (0.013) & 0.909 (0.027) & 0.89 (0.011) & 0.898 (0.027) \\
& Graph & 0.956 (0.015) & 0.855 (0.014) & 0.897 (0.027) & 0.895 (0.012) & 0.943 (0.006) \\
& Hier & 0.959 (0.005) & \textbf{0.872 (0.013)} & 0.907 (0.027) & 0.896 (0.013) & 0.949 (0.006) \\
& Margin & 0.974 (0.004) & 0.838 (0.012) & 0.915 (0.024) & 0.883 (0.015) & 0.945 (0.007) \\
& InfoDiv & 0.978 (0.003) & 0.842 (0.013) & 0.914 (0.024) & 0.883 (0.014) & 0.942 (0.004) \\
& Our & \textbf{0.981 (0.005)} & 0.858 (0.015) & \textbf{0.919 (0.021)} & \textbf{0.898 (0.012)} & \textbf{0.95 (0.005)} \\
\hline

\multirow{7}{*}{S = 10}
& Uniform & 0.881 (0.012) & 0.834 (0.017) & 0.89 (0.035) & 0.894 (0.011) & 0.943 (0.006) \\
& KCenter & 0.969 (0.006) & 0.858 (0.013) & 0.91 (0.027) & 0.89 (0.011) & 0.898 (0.027) \\
& Graph & 0.956 (0.016) & 0.855 (0.013) & 0.896 (0.027) & 0.895 (0.012) & 0.943 (0.006) \\
& Hier & 0.96 (0.01) & \textbf{0.87 (0.015)} & 0.912 (0.023) & 0.895 (0.013) & 0.949 (0.008) \\
& Margin & 0.973 (0.004) & 0.842 (0.015) & 0.911 (0.024) & 0.882 (0.015) & 0.943 (0.007) \\
& InfoDiv & 0.973 (0.003) & 0.844 (0.016) & 0.91 (0.023) & 0.884 (0.013) & 0.944 (0.004) \\
& Our & \textbf{0.978 (0.005)} & 0.86 (0.012) & \textbf{0.912 (0.019)} & \textbf{0.898 (0.013)} & \textbf{0.95 (0.004)} \\
\hline

\end{tabular}}
\end{table*}

\begin{table*}[!hbt]
\scriptsize
\centering
\caption{The mean and standard deviation performance on $5$ synthetic datasets of AUBC (auc) with $S = \{1, 5, 10\}$.}
\label{syncdataauc}
{\begin{tabular}{c|ccccccccc}
\toprule
 \multicolumn{1}{c}{}& & R15 & EX8a & EX8b & GCloud Balance & GCloud Unbalance \\
\toprule
\multirow{7}{*}{S = 1}
& Uniform & 0.935 (0.006) & 0.833 (0.018) & 0.892 (0.032) & 0.892 (0.011) & 0.953 (0.009) \\
& KCenter & 0.983 (0.002) & 0.856 (0.015) & 0.911 (0.025) & 0.889 (0.01) & 0.867 (0.046) \\
& Graph & 0.978 (0.006) & 0.853 (0.016) & 0.901 (0.024) & 0.894 (0.011) & 0.957 (0.004) \\
& Hier & 0.979 (0.003) & \textbf{0.866 (0.015)} & 0.907 (0.021) & 0.895 (0.013) & \textbf{0.961 (0.005)} \\
& Margin & 0.987 (0.001) & 0.839 (0.016) & 0.918 (0.022) & 0.881 (0.017) & 0.952 (0.005) \\
& InfoDiv & 0.987 (0.001) & 0.839 (0.016) & 0.918 (0.022) & 0.881 (0.017) & 0.952 (0.005) \\
& Our &\textbf{ 0.992 (0.002)} & 0.862 (0.014) & \textbf{0.92 (0.019)} & \textbf{0.896 (0.013)} & 0.958 (0.005) \\
\hline

\multirow{7}{*}{S = 5}
& Uniform & 0.935 (0.006) & 0.832 (0.018) & 0.892 (0.032) & 0.892 (0.011) & 0.954 (0.009) \\
& KCenter & 0.983 (0.003) & 0.856 (0.015) & 0.911 (0.024) & 0.889 (0.01) & 0.867 (0.046) \\
& Graph & 0.978 (0.006) & 0.853 (0.016) & 0.9 (0.024) & 0.894 (0.011) & \textbf{0.957 (0.004)} \\
& Hier & 0.978 (0.002) & \textbf{0.871 (0.013)} & 0.909 (0.024) & 0.894 (0.013) & 0.96 (0.005) \\
& Margin & 0.986 (0.002) & 0.836 (0.015) & 0.917 (0.022) & 0.881 (0.016) & 0.952 (0.008) \\
& InfoDiv & 0.988 (0.001) & 0.841 (0.016) & 0.917 (0.022) & 0.882 (0.015) & 0.95 (0.004) \\
& Our & \textbf{0.99 (0.003)} & 0.856 (0.018) & \textbf{0.921 (0.023)} & \textbf{0.897 (0.012)} & 0.956 (0.008) \\
\hline

\multirow{7}{*}{S = 10}
& Uniform & 0.935 (0.005) & 0.832 (0.019) & 0.892 (0.032) & 0.892 (0.011) & 0.953 (0.009) \\
& KCenter & 0.983 (0.003) & 0.856 (0.015) & 0.912 (0.024) & 0.889 (0.01) & 0.867 (0.046) \\
& Graph & 0.978 (0.006) & 0.853 (0.016) & 0.9 (0.023) & 0.895 (0.011) & 0.957 (0.004) \\
& Hier & 0.979 (0.004) & \textbf{0.869 (0.015)} & 0.914 (0.024) & 0.894 (0.013) & \textbf{0.96 (0.006)} \\
& Margin & 0.986 (0.002) & 0.841 (0.018) & 0.915 (0.022) & 0.881 (0.016) & 0.945 (0.01) \\
& InfoDiv & 0.985 (0.002) & 0.843 (0.019) & 0.914 (0.021) & 0.883 (0.014) & 0.949 (0.006) \\
& Our & \textbf{0.988 (0.003)} & 0.858 (0.016) & \textbf{0.915 (0.019)} & \textbf{0.897 (0.013)} & 0.956 (0.005) \\
\hline

\end{tabular}}
\end{table*}

\begin{table*}[!hbt]
\scriptsize
\centering
\caption{The mean and standard deviation performance on $5$ synthetic datasets of AUBC ($f_1$) with $S = \{1, 5, 10\}$.}
\label{syncdataf1}
{\begin{tabular}{c|ccccccccc}
\toprule
 \multicolumn{1}{c}{}& & R15 & EX8a & EX8b & GCloud Balance & GCloud Unbalance \\
\toprule
\multirow{7}{*}{S = 1}
& Uniform & 0.853 (0.012) & 0.82 (0.023) & 0.883 (0.039) & 0.902 (0.011) & 0.956 (0.004) \\
& KCenter & 0.961 (0.006) & 0.844 (0.02) & 0.903 (0.031) & 0.9 (0.011) & 0.929 (0.017) \\
& Graph & 0.95 (0.014) & 0.839 (0.021) & 0.894 (0.031) & 0.903 (0.012) & 0.956 (0.004) \\
& Hier & 0.951 (0.006) & \textbf{0.86 (0.017)} & 0.899 (0.032) & 0.904 (0.012) & 0.962 (0.006) \\
& Margin & 0.972 (0.003) & 0.824 (0.024) & 0.909 (0.027) & 0.892 (0.013) & 0.956 (0.005) \\
& InfoDiv & 0.972 (0.003) & 0.824 (0.024) & 0.909 (0.027) & 0.892 (0.013) & 0.956 (0.005) \\
& Our & \textbf{0.983 (0.004)} & 0.852 (0.022) & \textbf{0.912 (0.024)} & \textbf{0.905 (0.014)} & \textbf{0.963 (0.005)} \\
\hline

\multirow{7}{*}{S = 5}
& Uniform & 0.853 (0.012) & 0.82 (0.023) & 0.883 (0.039) & 0.902 (0.011) & 0.956 (0.004) \\
& KCenter & 0.962 (0.006) & 0.844 (0.02) & 0.904 (0.031) & 0.9 (0.011) & 0.929 (0.017) \\
& Graph & 0.95 (0.014) & 0.839 (0.021) & 0.893 (0.031) & 0.903 (0.012) & 0.956 (0.004) \\
& Hier & 0.95 (0.005) & \textbf{0.865 (0.015)} & 0.901 (0.032) & 0.903 (0.012) & 0.961 (0.005) \\
& Margin & 0.969 (0.004) & 0.823 (0.025) & 0.909 (0.028) & 0.891 (0.012) & 0.959 (0.005) \\
& InfoDiv & 0.974 (0.004) & 0.826 (0.026) & 0.909 (0.028) & 0.892 (0.012) & 0.956 (0.004) \\
& Our & \textbf{0.978 (0.006)} & 0.843 (0.027) & \textbf{0.914 (0.024)} & \textbf{0.904 (0.011)} & \textbf{0.962 (0.004)} \\
\hline

\multirow{7}{*}{S = 10}
& Uniform & 0.853 (0.011) & 0.82 (0.023) & 0.883 (0.039) & 0.902 (0.011) & 0.956 (0.004) \\
& KCenter & 0.962 (0.007) & 0.843 (0.02) & 0.905 (0.031) & 0.9 (0.011) & 0.929 (0.017) \\
& Graph & 0.95 (0.015) & 0.839 (0.02) & 0.892 (0.031) & 0.903 (0.012) & 0.956 (0.004) \\
& Hier & 0.95 (0.01) & \textbf{0.863 (0.016)} & 0.908 (0.026) & 0.903 (0.012) & 0.961 (0.006) \\
& Margin & 0.968 (0.004) & 0.829 (0.028) & 0.906 (0.029) & 0.891 (0.012) & 0.958 (0.005) \\
& InfoDiv & 0.967 (0.004) & 0.829 (0.028) & 0.905 (0.027) & 0.893 (0.013) & 0.958 (0.003) \\
& Our & \textbf{0.973 (0.006)} & 0.846 (0.024) & \textbf{0.907 (0.023)} & \textbf{0.904 (0.013)} & \textbf{0.962 (0.003)} \\
\hline

\end{tabular}}
\end{table*}


Table~\ref{avgres} presents the average ranking of each method over all datasets when evaluating with respect to each AUBC metric.
The detailed performance results (mean and standard deviations) of the AL methods on the real-world datasets using the AUBC (acc), AUBC(auc) and AUBC($f_1$) metrics are presented in Tables \ref{realdataacc},~\ref{realdataauc}, and \ref{realdataf1}.
Likewise, the performance results on the synthetic datasets are presented in Tables \ref{syncdataacc}, \ref{syncdataauc}, and \ref{syncdataf1}.


According to Table~\ref{avgres}, the overall ranks of the proposed model and baselines are as follows: ours $>$ \textbf{Hier} $>$ \textbf{InfoDiv}, \textbf{Margin} $>$ \textbf{Graph} $>$ \textbf{KCenter} $>$ \textbf{Uniform}. On a large majority of datasets in the experiments, our model achieves optimal performance (ranking 1). The performance of the AL baselines (i.e., \textbf{Hier}, \textbf{InfoDiv}, \textbf{Margin}, \textbf{Graph}, \textbf{KCenter}) are typically inconsistent across tasks, scoring well on some tasks but not on others.
 For example, based on AUBC (acc) metric with $S=1$, \textbf{Hier} has high rankings on \emph{ionosphere}(2), \emph{ex8a}(1), \emph{vehicle}(2), \emph{gcloud unbalance}(1), but has low rankings on \emph{sonar}(5), \emph{wine}(6), \emph{r15}(5), \emph{ex8b}(5) datasets.
 Similar trends are observed with the other baselines.
The better performance of our method shows the advantage of DPP to dynamically combine the three AL criterion for effective selection of samples.
Our DPP outperforms \textbf{Graph}, the latter also uses three criteria using a weighted sum optimization that highly depends on the trade-off hyperparameter setting (set to $0.5$ in \cite{ebert2012ralf}). Other methods, such as \textbf{Margin}, \textbf{InfoDiv}, \textbf{KCenter}, and \textbf{Hier}, also cannot handle all datasets well, due to the reliance on only 1 or 2 criteria.

\begin{figure*} [tb]
\centering
\begin{minipage}{5.5cm}
\centering
\includegraphics[width=5.5cm]{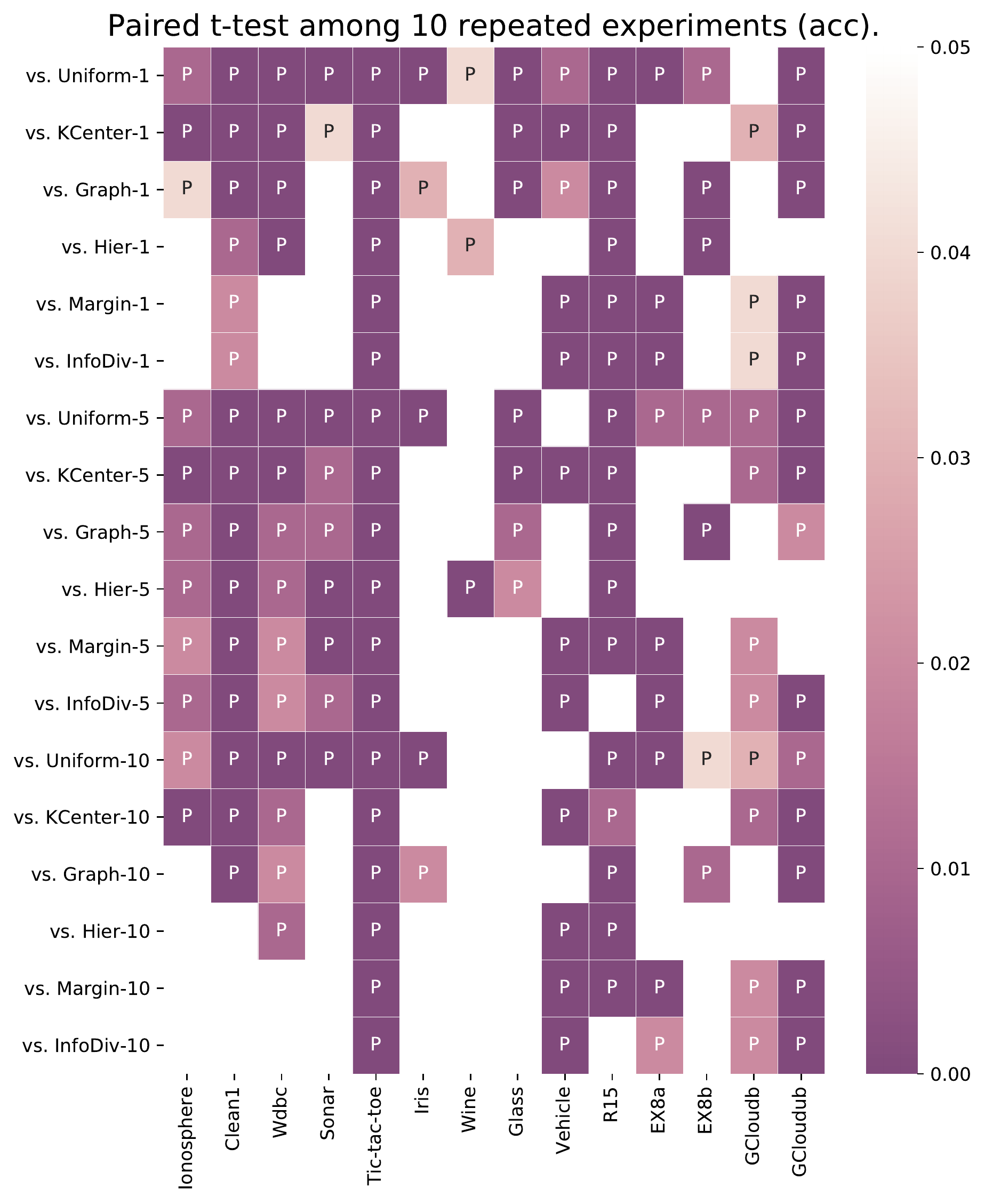}
\footnotesize(a) accuracy
\end{minipage}
\begin{minipage}{5.5cm}
\centering
\includegraphics[width=5.5cm]{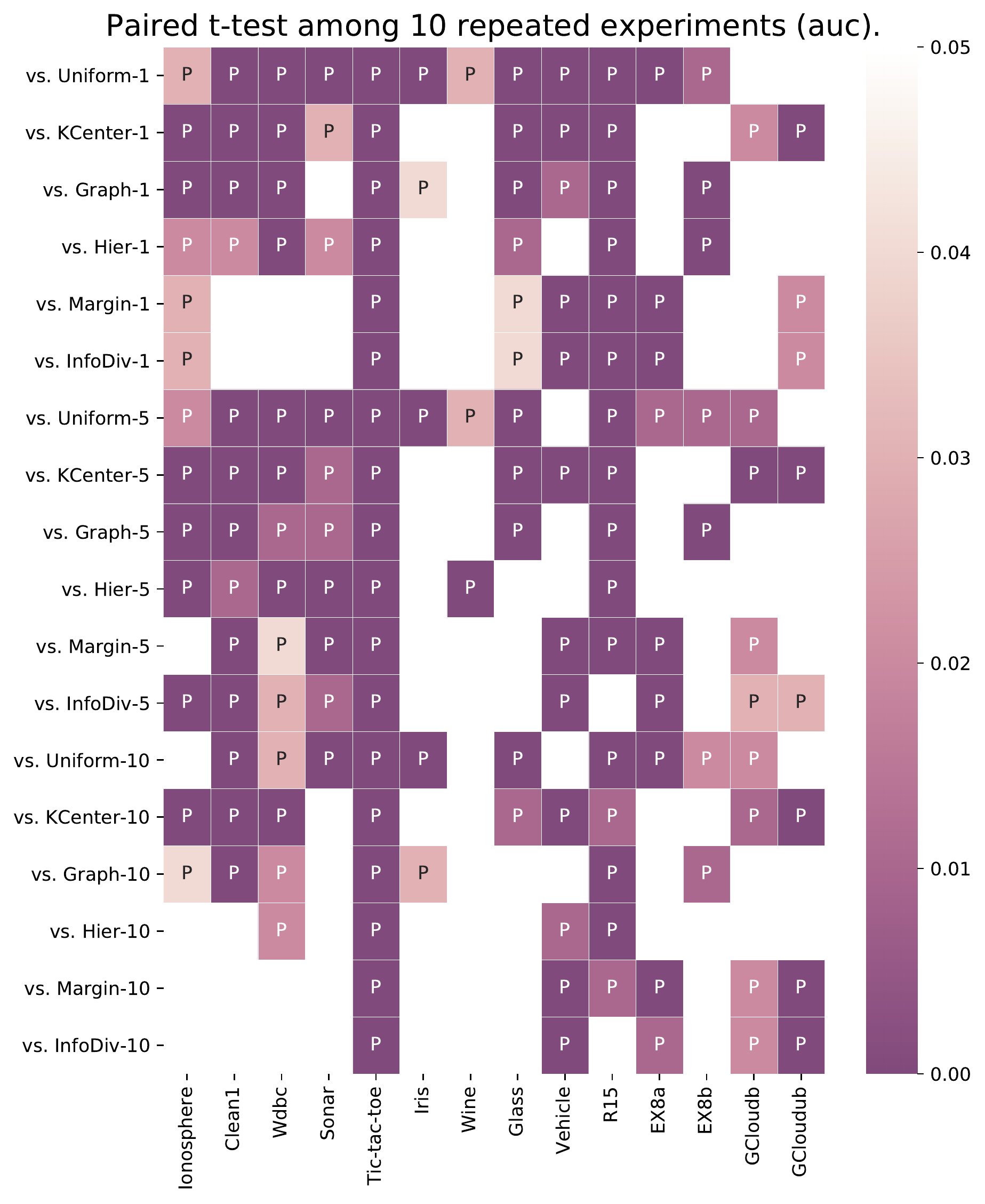}
\footnotesize(b) auc
\end{minipage}
\begin{minipage}{5cm}
\centering
\includegraphics[width=5.5cm]{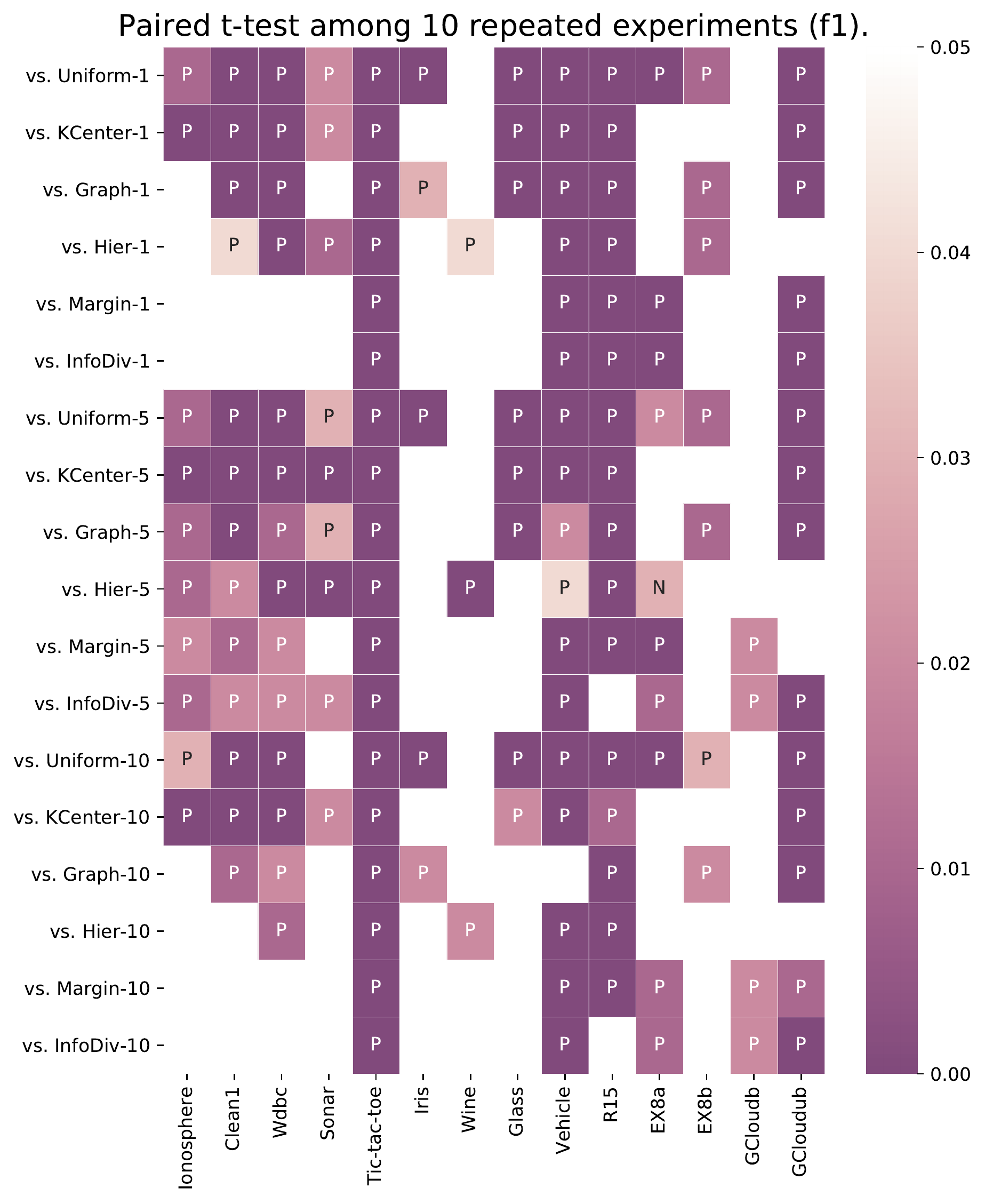}
\footnotesize(c) $f_1$
\end{minipage}
\caption{Paired t-test comparing our model and baselines on 14 real-world and synthetic datasets ($S=\{1,5,10\}$) using 10 trials using (a) accuracy, (b) auc, and (c) $f_1$.
The color bar are centered at $p = 0.01$, with dark purple indicating $p < 0.01$, and light purple indicating $0.01 <p < 0.05$. Comparisons with no significant difference ($p>0.05$) are marked as white.
For significant differences, the letter ``P'' indicates that our proposed method has average higher metric than the baseline on that dataset, while ``N'' indicates our method has lower metric.
Our method performs better than or the same as the baselines on all datasets, except for one case (Hier-5 on Ex8a using $f_1$).
}
\label{ttest}
\end{figure*}

We further examine the differences in performance between our method and baselines by using a paired t-test on the 10 repeated trials on each dataset.
The test results are shown in Figure~\ref{ttest}.
Our method outperforms the baselines at a statistically significant level ($p<0.05$) on $467$ out of $756$ ($62\%$) of the experiments\footnote{Totally we have $6$ baselines $\times$ $3$ batch settings $\times$ $14$ datasets $\times$ $3$ evaluation metrics = $756$ experimental results to analyze.}, while performing similarly to baselines ($p>0.05$) on $289$ out of $756$ ($38\%$).
This is meaningful
since on some datasets, e.g., \emph{vehicle}, \emph{ex8a}, \emph{gcloud balance} and \emph{gcloud unbalance}, even random sampling (\textbf{Uniform}) achieves good performance, suggesting that there
%
is not much room for improvement on some datasets.
Our method only underperforms on one experiment comparing with \textbf{Hier} on \emph{ex8a} dataset using AUBC ($f_1$) metric.
In summary, the t-test results indicate that, compared with the baseline models, our model can achieve better or similar results under different task scenarios.

Another advantage of our method is that there are no hyperparameter to fine-tune (except for the batch size settings) in the $k$DPPs. In contrast, for the baseline models,  \textbf{Graph} needs to fine-tune the trade-off parameters between exploration and exploitation, and \textbf{Hier} needs to find optimal trade-off parameters for each task. Compared with these multiple-criteria based AL strategies that adopt linear combination (or other combination strategies that need to set trade-off parameters between various criteria), our method can ``just run'' for any upcoming AL task.

\subsection{Budget Curves}

Figures~\ref{acc-figure},~\ref{auc-figure}, and \ref{f1-figure} show the budget curves when using 
the accuracy, auc, $f_1$ metrics, respectively.
For the AUBC metric, a method that first reaches the peak value and then converges will typically have better AUBC performance. Our model often reaches the peak value faster than baselines. For example, based on AUBC(acc) curves in Figure~\ref{acc-figure}, on \emph{wdbc} our model reaches peak value at budget 50, compared to 200 for the baselines (with similar trends on \emph{ionosphere}, \emph{tic-tac-toe}, \emph{r15}, and \emph{gcloud unbalance}).
 %
In our DPP, the informativeness criterion enables the peak value to be reached faster, while the representativeness criterial helps capture the data topology to achieve faster convergence.
%

A good AL algorithm should also produce 
budget curves that are stable and monotonically increasing, so as to handle scenarios with different budget requirements. Our method is more stable than other baselines, whereas the performance of some baselines drop significantly after reaching the peak values, e.g., based on AUBC (acc) Curves (see Figure~\ref{acc-figure}),
\textbf{KCenter} exhibits performance degradation on \emph{ionosphere}, \emph{tic-tac-toe}, \emph{ex8b} and \emph{gcloud unbalance}, while \textbf{Margin} and \textbf{InfoDiv} show degradation on  \emph{iris}  and \emph{gcloud balance}.
More details can be found in Figures~\ref{acc-figure},~\ref{auc-figure}, and \ref{f1-figure}.

\subsection{Different Batch Sizes and Diversity}
To evaluate the effectiveness of the diversity measure, we compare the performance with batch-sizes $S = 1$ (no diversity used since only one sample is selected) and $S > 1$ (diversity used).
If the performance does not drop with increasing batch size, then this suggests that the diversity strategy works since the large batch yields equivalent performance to selecting one sample at a time.
Then batch-mode AL can be used to reduce computational cost (re-training the classifiers) without degradation in performance.
Without using a diversity measure, the same (or similar) points would be selected in a batch, which reduces the performance when compared with selecting 1 sample at a time.

We summarize the average performance drop between $S = 1$ and $S > 1$ of our method and baselines across different datasets in Table~\ref{div}. For our algorithm and the baselines that use diversity, the AUBC performance does not decrease much with the increase of batch size (e.g., the maximum drop is $0.006$ on our method with $S = 10$ based on AUBC ($f_1$).
%
This demonstrates the effectiveness of our DPP for batch sampling for AL.


\begin{table*}[!hbt]
\scriptsize
\centering
\caption{Performance drop of batch-mode AL: the average difference between $S = 1$ with $S = 5$ and $S = 10$ across $14$ datasets.}
\label{div}
{\begin{tabular}{c|cccccccc}
\toprule
\multicolumn{1}{c}{}& & Uniform & KCenter & Graph & Hier & Margin & InfoDiv & Our  \\
\toprule

\multirow{3}{*}{S =1 vs. S = 5}
& AUBC (acc) & 0.000 & 0.000 & 0.000 & 0.003 & 0.003 & 0.002 & 0.004 \\
& AUBC (auc) & 0.000 & 0.000 & 0.000 & 0.003 & 0.003 & 0.003 & 0.003 \\
& AUBC ($f_1$) & 0.000 & 0.001 & 0.000 & 0.004 & 0.003 & 0.003 & 0.004 \\
\hline

\multirow{3}{*}{S =1 vs. S = 10}
& AUBC (acc) & 0.000 & 0.001 & 0.000 & 0.005 & 0.003 & 0.003 & 0.005 \\
& AUBC (auc) & 0.000 & 0.001 & 0.001 & 0.004 & 0.003 & 0.003 & 0.005 \\
& AUBC ($f_1$) & 0.001 & 0.001 & 0.001 & 0.004 & 0.004 & 0.004 & 0.005 \\
\hline

\end{tabular}}
\end{table*}

\subsection{Impact of Different Kernels}
\label{kernels}
In Section~\ref{model}, we have introduced various kernel functions that can be used to construct similarity matrix (also L matrix). To explore the impact of different kernels to our batch AL approach, we compare the performances of the heat, Gaussian, polynomial, sigmoid, and Laplacian kernels, and the results are in Table~\ref{kernel}. 
We compute the average rank of the various kernels across the 9 datasets and various batch size settings, resulting in the following ranking:
Laplacian (1.5), Poly (2.3), Sigmoid (2.8), Heat (3.3), Gaussian (3.7).
Thus, we choose the Laplacian kernel in our experiments since it performs better on average.

\begin{table*}[!hbt]
\tiny
\centering
\caption{The mean and standard deviation performance of the proposed model among $5$ different kernels on various datasets with AUBC (acc) ($S = \{1, 5, 10\}$).}
\label{kernel}
{\begin{tabular}{c|ccccccccccc}
\toprule
 \multicolumn{1}{c}{}& & Ionosphere & Iris & Wine & Glass & Vehicle & EX8a & EX8b & GCloud Balance & GCloud Unbalance \\
\toprule
\multirow{5}{*}{S = 1}
& Heat & 0.941 (0.011) & 0.948 (0.016) & \textbf{0.977 (0.01)} & 0.609 (0.037) & 0.709 (0.014)  & 0.857 (0.014) & 0.913 (0.025) & 0.896 (0.013) & 0.948 (0.004) \\
& Gaussian & \textbf{0.943 (0.013)} & 0.95 (0.019) & 0.975 (0.011) & 0.604 (0.036) & 0.71 (0.016)  & 0.855 (0.015) & 0.912 (0.022) & \textbf{0.898 (0.013)} & 0.946 (0.005) \\
& Poly & 0.941 (0.013) & 0.948 (0.023) & 0.976 (0.012) & 0.61 (0.03) & 0.711 (0.015) & \textbf{0.864 (0.018)} & 0.914 (0.023) & 0.896 (0.012) & 0.948 (0.007) \\
& Sigmoid & 0.943 (0.014) & 0.947 (0.017) & 0.974 (0.012) & 0.609 (0.032) & \textbf{0.716 (0.015)} & 0.858 (0.014) & 0.91 (0.017) & 0.897 (0.013) & 0.949 (0.003) \\
& Laplacian & 0.941 (0.015) & \textbf{0.952 (0.016)} & 0.977 (0.011) & \textbf{0.622 (0.03)} & 0.715 (0.016) & 0.863 (0.011) & \textbf{0.917 (0.02)} & 0.897 (0.014) & \textbf{0.95 (0.006)} \\
\hline

\multirow{5}{*}{S = 5}
& Heat & 0.93 (0.017) & 0.94 (0.024) & \textbf{0.978 (0.012)} & 0.587 (0.042) & 0.698 (0.013) & \textbf{0.861 (0.014)} & 0.906 (0.033) & 0.898 (0.013) & 0.946 (0.006) \\
& Gaussian & 0.935 (0.019) & 0.936 (0.019) & 0.974 (0.012) & 0.571 (0.039) & 0.695 (0.013) & 0.858 (0.011) & 0.907 (0.018) & 0.897 (0.012) & 0.946 (0.007) \\
& Poly & 0.937 (0.012) & \textbf{0.947 (0.02)} & 0.975 (0.012) & 0.596 (0.026) & 0.706 (0.015) & 0.851 (0.008) & 0.912 (0.019) & 0.897 (0.016) & 0.945 (0.004) \\
& Sigmoid & 0.939 (0.012) & 0.946 (0.018) & 0.974 (0.009) & 0.598 (0.041) & 0.705 (0.017) & 0.855 (0.01) & 0.912 (0.019) & \textbf{0.898 (0.012)} & 0.943 (0.006) \\
& Laplacian & \textbf{0.942 (0.012)} & 0.946 (0.022) & 0.975 (0.01) & \textbf{0.606 (0.034)} & \textbf{0.71 (0.014)} & 0.858 (0.015) & \textbf{0.919 (0.021)} & \textbf{0.898 (0.012)} & \textbf{0.95 (0.005)} \\
\hline

\multirow{5}{*}{S = 10}
& Heat & 0.926 (0.019) & 0.932 (0.021) & 0.976 (0.014) & 0.576 (0.037) & 0.681 (0.016) & 0.855 (0.015) & 0.908 (0.03) & 0.897 (0.013) & 0.946 (0.009) \\
& Gaussian & 0.928 (0.019) & 0.928 (0.025) & 0.977 (0.014) & 0.58 (0.045) & 0.692 (0.016) & 0.854 (0.018) & 0.904 (0.035) & 0.897 (0.013) & 0.946 (0.006) \\
& Poly & 0.935 (0.014) & \textbf{0.946 (0.022)} & \textbf{0.978 (0.011)} & 0.575 (0.046) & 0.71 (0.016) & 0.852 (0.013) & 0.91 (0.013) & 0.897 (0.012) & 0.947 (0.007) \\
& Sigmoid &\textbf{ 0.939 (0.013)} & 0.941 (0.018) & 0.976 (0.009) & \textbf{0.596 (0.044)} & \textbf{0.712 (0.018)} & 0.848 (0.019) & 0.908 (0.02) & 0.896 (0.013) & 0.943 (0.01) \\
& Laplacian & 0.937 (0.014) & 0.945 (0.023) & 0.976 (0.011) & 0.586 (0.05) & 0.708 (0.017) & \textbf{0.86 (0.012)} & \textbf{0.912 (0.019)} & \textbf{0.898 (0.013)} & \textbf{0.95 (0.004)} \\
\hline

\end{tabular}}
\end{table*}

\subsection{Analysis of the Expertise Weight of the Committee}

\begin{figure} [tb]
\centering
\begin{minipage}{4.2cm}
\centering
\includegraphics[width=4.2cm]{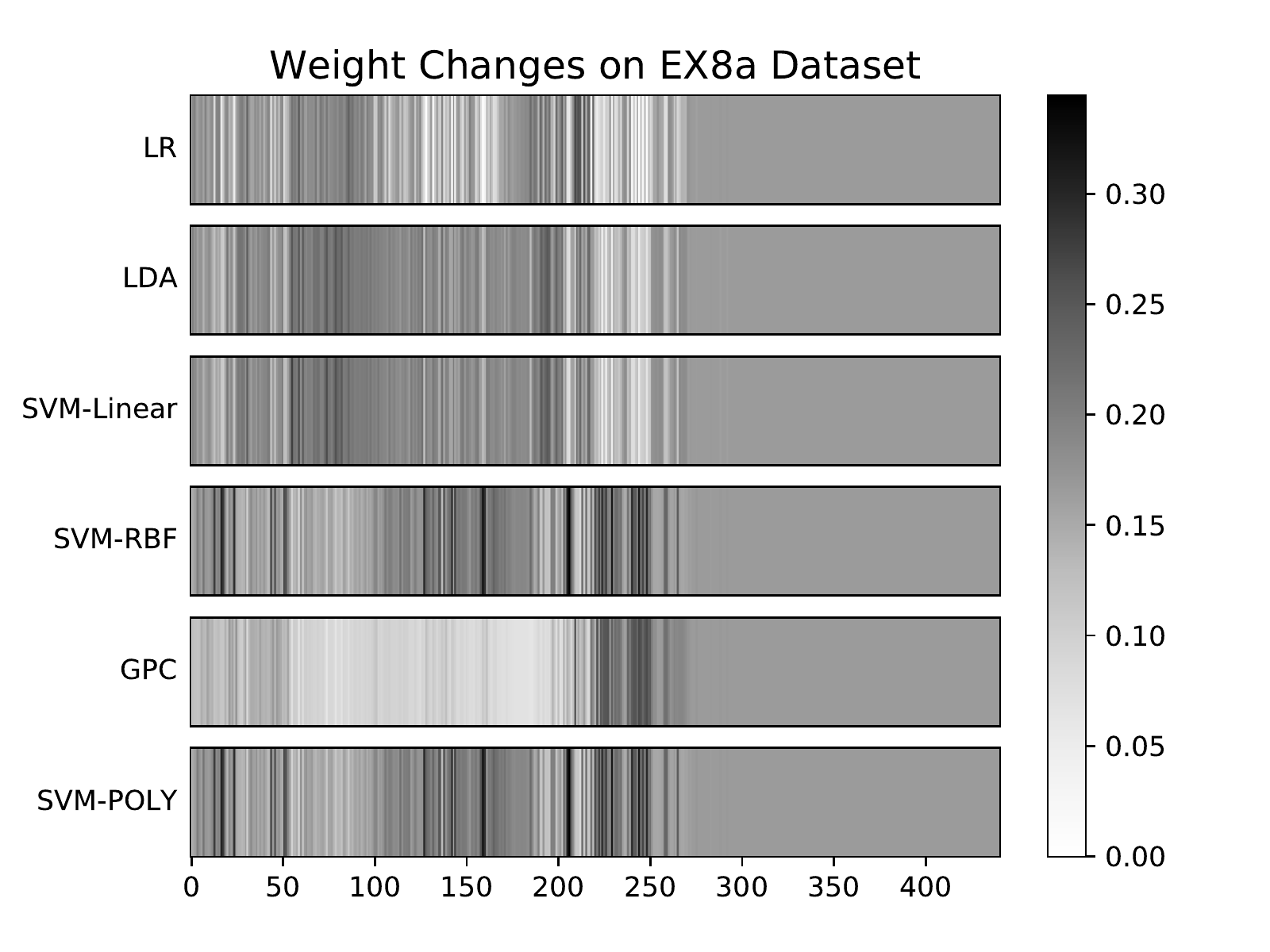}
\footnotesize(a) EX8a
\end{minipage}
\begin{minipage}{4.2cm}
\centering
\includegraphics[width=4.2cm]{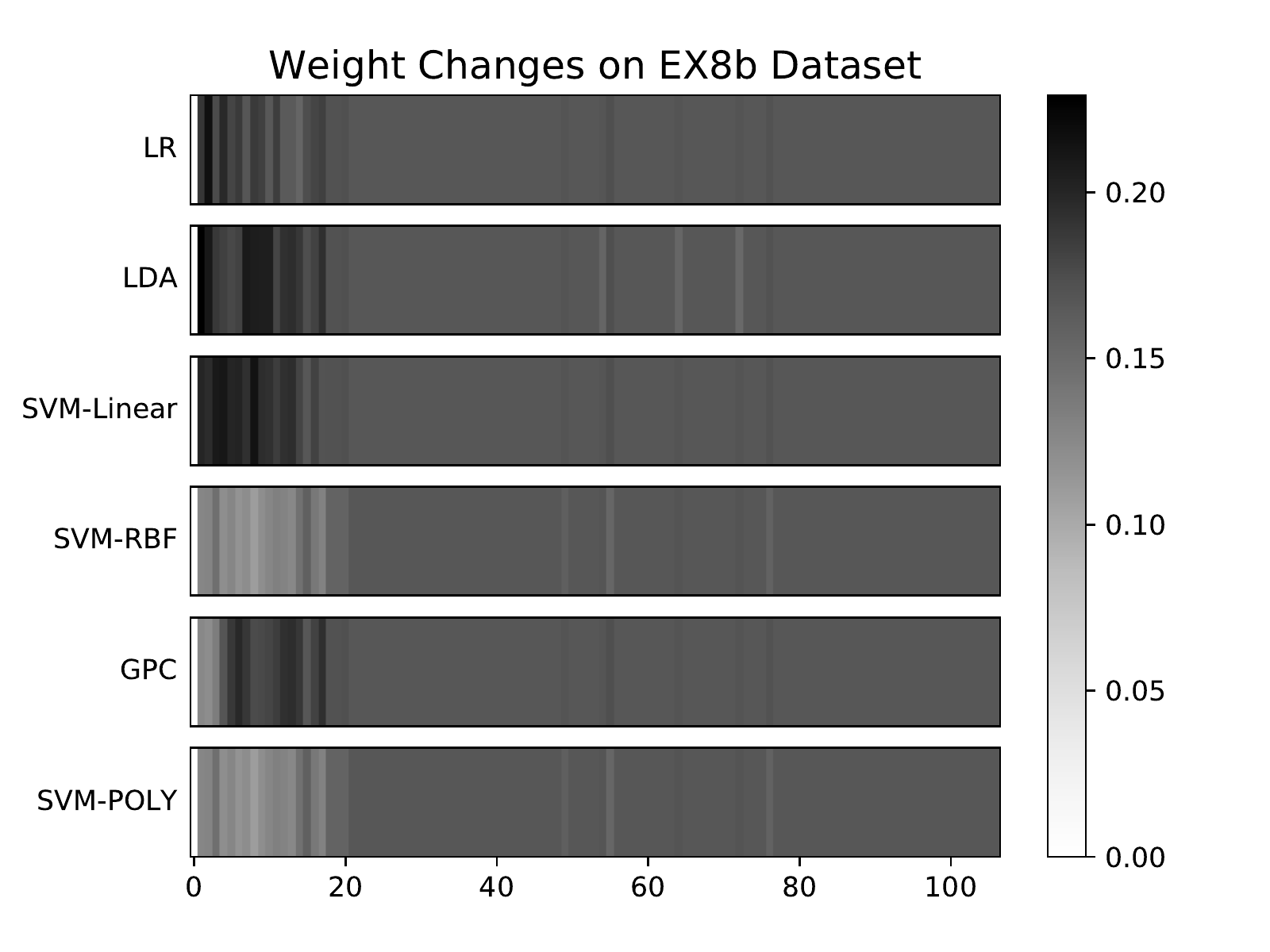}
\footnotesize(b) EX8b
\end{minipage}
\begin{minipage}{4.2cm}
\centering
\includegraphics[width=4.2cm]{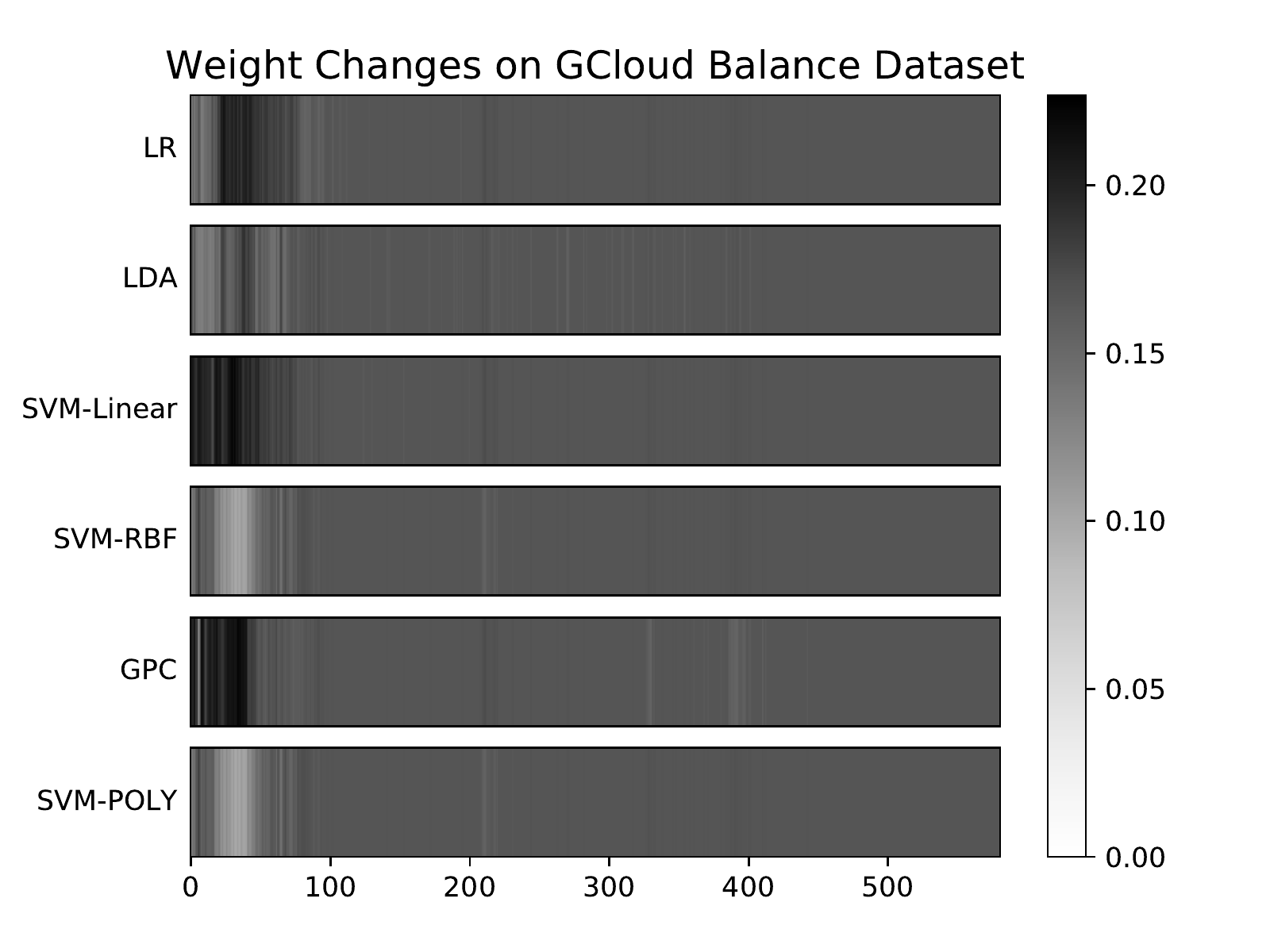}
\footnotesize(c) GCloud Balance
\end{minipage}
\begin{minipage}{4.2cm}
\centering
\includegraphics[width=4.2cm]{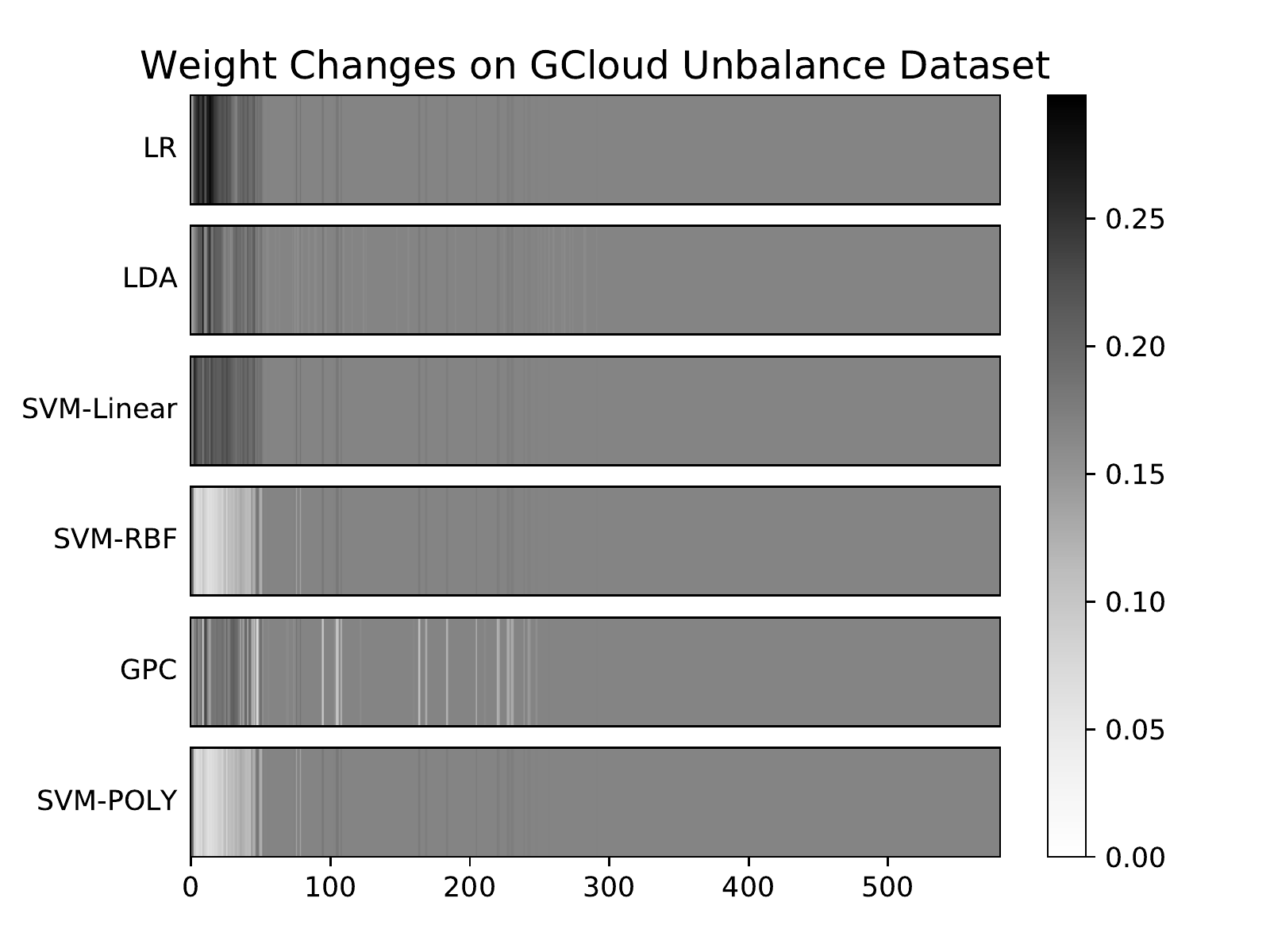}
\footnotesize(d) GCloud Unbalance
\end{minipage}
\caption{The expertise weights ($\bbeta$) of classifiers during the AL iterations on the synthetic binary-classification datasets.}
\label{weight-synt-figure}
\end{figure}

It is interesting to explore how the expertise weights $\bbeta$ of the committee classifiers vary during the active learning process. Figure~\ref{weight-synt-figure} shows the computed weights of the committee classifiers during AL iterations, as the amount of labeled data increases.
The weights converge after collecting enough data points to estimate the classification boundary (i.e., $260$ samples for \emph{ex8a}, $20$ for \emph{ex8b} and $100$ for \emph{gcloud}). Our approach also assigns higher weights to the classifiers that are more suitable for the current data. For example, SVM-RBF and SVM-POLY are assigned higher weights than linear classifiers for non-linear data (e.g., \emph{ex8a}), while LDA, LR and SVM-Linear have higher weights than non-linear classifiers for linear separable data (e.g., \emph{ex8b}). These examples show how the informativeness criterion adapts the committee of classifiers to best fit the current data distribution.

\subsection{Analysis of Data Imbalance Situations}
\begin{figure} [tb]
\begin{minipage}{9.5cm}
\centering
\includegraphics[width=9.5cm]{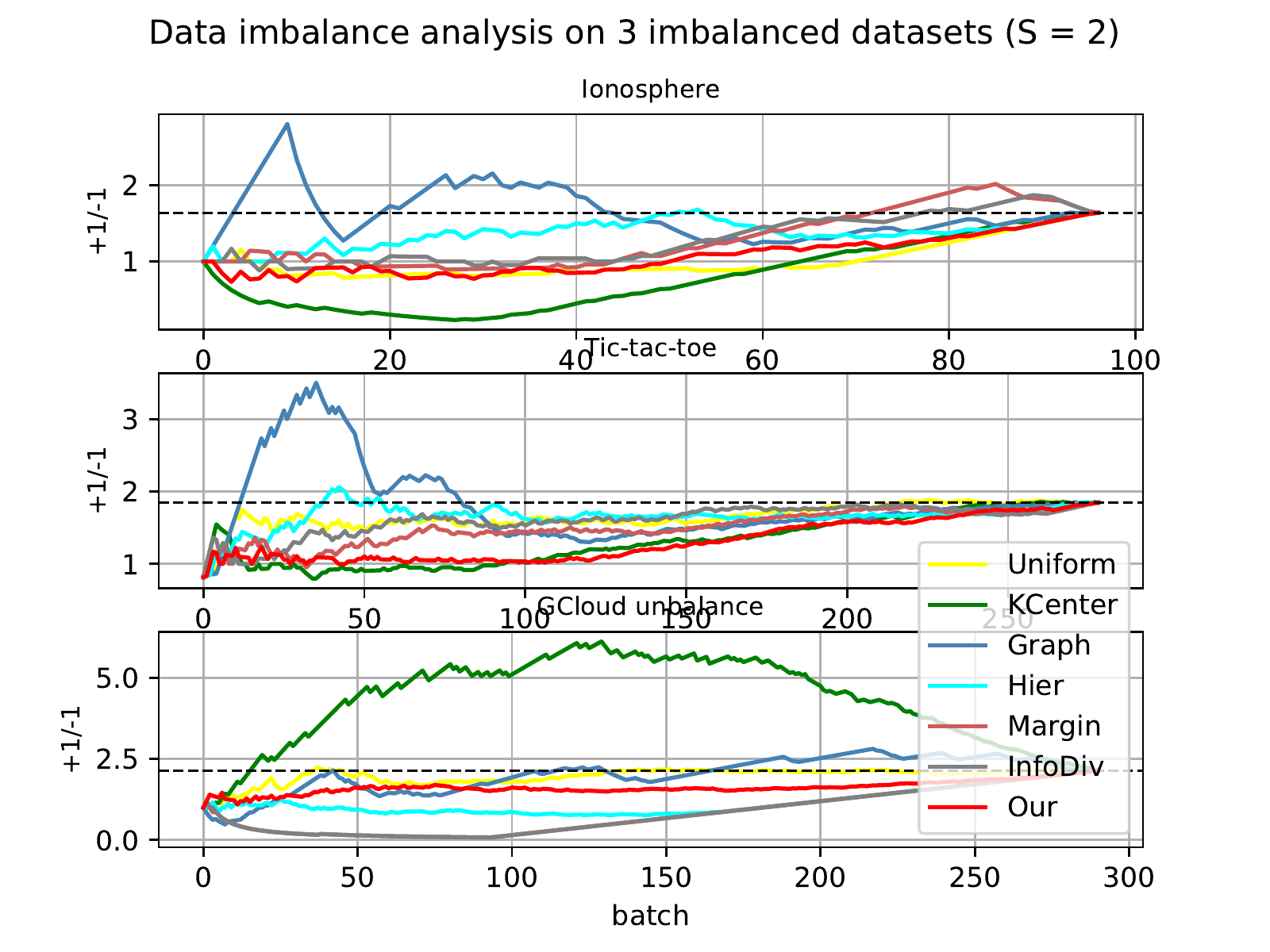}
\end{minipage}
\caption{Analysis of imbalanced datasets (\emph{ionosphere}, \emph{tic-tac-toe} and \emph{gcloud unbalance}) with PN Ratio ($+1/-1$) of selected samples during AL processes ($S = 2$). The dashed line represents for the PN ratio of the whole unlabeled data pool.}

\label{imbalance}
\end{figure}

We also examine the behavior of AL models on data with unbalanced classes (e.g., tic-tac-toe dataset), in particular
whether they can select subsets from the unlabeled data to make the labeled set balanced.
Figure~\ref{imbalance} shows the ratio of positive/negative (PN) samples during the AL process on \emph{ionosphere} (PN of 9:5 in the whole dataset), \emph{tic-tac-toe} (PN of 7:1) and \emph{GCloud Unbalance} (PN of 2:1). \textbf{Graph} prefers selecting positive samples first, which makes the PN ratio very large in the early AL rounds. \textbf{KCenter} initially selects too many
samples that belong to the same class, 
which causes a very serious data imbalance problem in its AL process. Because \textbf{KCenter} does not use the labels of the labeled set, it fails to detect and fix the data imbalance problem.
In contrast, our method reacts to the imbalance problem -- it keeps the PN ratio around 1:1 at the
beginning
 of the AL process on \emph{ionosphere} (first 50 rounds) and tic-tac-toe datasets (first 125 rounds), and reduces the PN ratio to $1.5:1$ on the \emph{gcloud unbalance} dataset. The negative samples among the three class imbalanced datasets are more difficult to classify due to their rarity. In particular, our model assigns higher weights in $\balpha$ to the negative samples, causing the probability of selecting negative samples to increase.

\subsection{Analysis of the Importance Level of Various Criteria Under Varying Budgets}

\begin{figure} [tb]
\centering
\begin{minipage}{10cm}
\centering
\includegraphics[width=10cm]{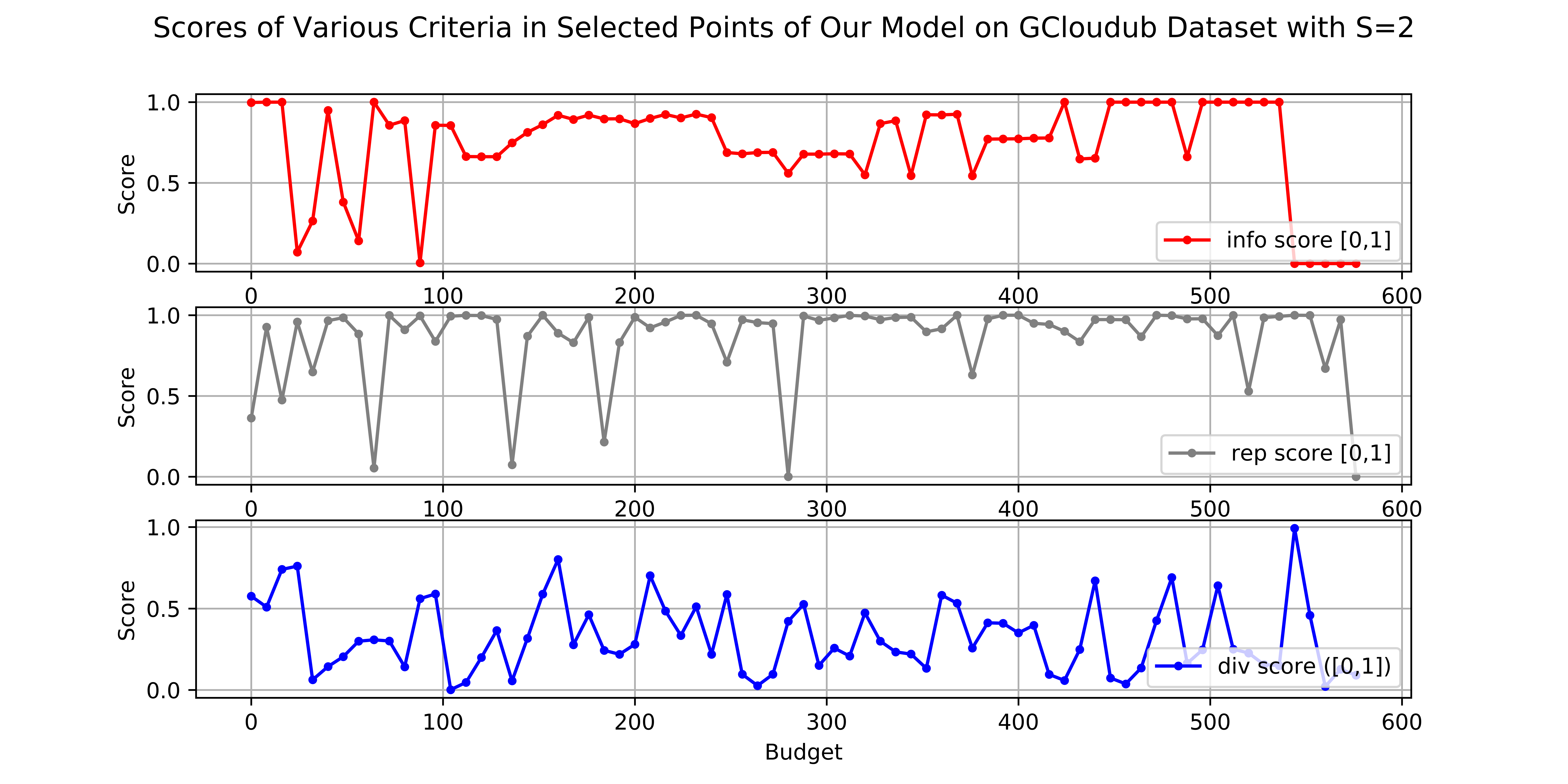}
\end{minipage}
\caption{Changes of the importance level (score) of the three criteria (\emph{GCloud Unbalance} dataset), higher scores mean more informative, representative, or diverse ($S=2$).}
\label{gcloudub_seq}
\end{figure}

In Figure~\ref{gcloudub_seq}, we record the scores of informativeness, representativeness and diversity criteria for each selected point during the AL process. All the informativeness and representativeness scores are normalized by the total informativeness and representativeness values of unlabeled data pool, respectively, while the diversity score is measured by the pair-wise distance with the batch (we set $S=2$ to facilitate computations). In each iteration, the selection criteria from DPP change dynamically, with the selected points having varying levels of informativeness, representativeness, and diversity. From the aspect of the whole process, the representativeness measure is more important, since most of its scores are above $0.5$.
Based on the consideration of high representativeness and diversity score, our approach also tends to select data points with higher informativeness score, which is consistent with our analysis in Section~\ref{model}. In particular, we note that the data pairs that have higher inner dissimilarity are more probable because their feature vectors are more orthogonal to span larger volumes. Furthermore, the data points with larger magnitude feature representations are more likely to be selected, because they multiply the spanned volumes for sets containing them. 

\section{Conclusion}
\label{summary}
In this paper, we introduce a multiple-criteria based AL algorithm that incorporates the informativeness, representativeness and diversity criteria together with $k$DPPs, which can be used for both binary and multi-class classification tasks. Firstly, we refine the query process by selecting the data samples that are the hardest for labeling, while adjusting the weights to adapt the classifiers to the given data distribution simultaneously. Then, the representative information mined by the $k$-center based clustering algorithm also ensures the adaptability of our proposed model under various data topologies. Finally, DPP makes the selected batch in each iteration to have enough quality and diversity. We propose the AUBC for evaluating the overall performance of each AL algorithm under ever-increasing budgets. The evaluations on synthetic and real-world datasets show that our proposed method performs better and is more stable than baseline methods. In the future, we will explore more combination methods of multiple-criteria based AL which could integrate more diverse goals for optimization. 
We will also study approximation methods for $k$DPP to reduce its computational complexity.


\bibliographystyle{IEEEtran}
\bibliography{IEEEexample}


%

\ifCLASSOPTIONcaptionsoff
  \newpage
\fi

\end{document}